\documentclass{article}
\usepackage{arxiv}
\usepackage{authblk}
\usepackage[utf8]{inputenc} % allow utf-8 input
\usepackage[T1]{fontenc}    % use 8-bit T1 fonts
\usepackage{hyperref}       % hyperlinks
\usepackage{url}            % simple URL typesetting
\usepackage{booktabs}       % professional-quality tables
\usepackage{amsmath}
\usepackage{amsfonts}       % blackboard math symbols
\usepackage{nicefrac}       % compact symbols for 1/2, etc.
\usepackage{microtype}      % microtypography
\usepackage{amssymb}
\usepackage{systeme}
\usepackage{bbm}
\usepackage{stmaryrd}
\newcommand{\ssymbol}[1]{^{\@fnsymbol{#1}}}
\usepackage{bm}
\usepackage{graphicx}
\usepackage{natbib}
\usepackage{doi}
\usepackage{subfig}
\usepackage{float}
\usepackage{multirow}

\makeatletter
\renewcommand\AB@authnote[1]{\textsuperscript{\normalfont\bfseries#1}}
\makeatother

\title{Physics-informed neural networks for non-Newtonian fluid thermo-mechanical problems:\\ an application to rubber calendering process}

\author[1,2,3]{Thi Nguyen Khoa Nguyen}
\author[2]{Thibault Dairay}
\author[2]{Raphaël Meunier}
\author[1,4]{Mathilde Mougeot}
\affil[1]{Universite Paris-Saclay, ENS Paris-Saclay, CNRS, Centre Borelli, Gif-sur-Yvette, 91190 France}
\affil[2]{Michelin, Centre de Recherche de Ladoux, Cébazat, 63118 France}
\affil[3]{CEA, DAM, DIF, F-91297 Arpajon, France}
\affil[4]{ENSIIE, Évry-Courcouronnes, 91000 France}
\begin{document}
\maketitle

\begin{abstract}
Physics-Informed Neural Networks (PINNs) have gained much attention in various fields of engineering thanks to their capability of incorporating physical laws into the models. However, the assessment of PINNs in industrial applications involving coupling between mechanical and thermal fields is still an active research topic. In this work, we present an application of PINNs to a non-Newtonian fluid thermo-mechanical problem which is often considered in the rubber calendering process. We demonstrate the effectiveness of PINNs when dealing with inverse and ill-posed problems, which are impractical to be solved by classical numerical discretization methods. We study the impact of the placement of the sensors and the distribution of unsupervised points on the performance of PINNs in a problem of inferring hidden physical fields from some partial data. We also investigate the capability of PINNs to identify unknown physical parameters from the measurements captured by sensors. The effect of noisy measurements is also considered throughout this work. The results of this paper demonstrate that in the problem of identification, PINNs can successfully estimate the unknown parameters using only the measurements on the sensors. In ill-posed problems where boundary conditions are not completely defined, even though the placement of the sensors and the distribution of unsupervised points have a great impact on PINNs performance, we show that the algorithm is able to infer the hidden physics from local measurements. 
\end{abstract}

% \end{frontmatter}

% \linenumbers

\section{Introduction}
In the last few years, advances in deep learning have gained much attention in the field of physical modeling and engineering. There has been an increasing amount of research using deep learning methods to study the physics of systems by enforcing physical constraints, which are usually expressed by partial differential equations (PDEs), into the models \citep{raissi2019physics, sirignano2018dgm, rackauckas2020universal, li2020fourier, gao2021phygeonet}. Among these studies, the series of work of Karniadakis’ group about Physics-Informed Neural Networks (PINNs) \citep{raissi2019physics} rise as an attractive and remarkable scheme of solving forward and inverse PDEs problems using only a moderate amount of supervised data \citep{zhu2019physics, sun2020physics, sun2020surrogate}. PINNs take advantage of both data-driven modeling and physics-based modeling to estimate the unknown physics of interest, which may be physical fields or physical parameters of a system. PINNs use neural networks as approximators and integrate the physical constraints of the systems into the cost function. The training process of PINNs tries to penalize (1) the loss on the initial/boundary condition if the problem is well-defined, (2) the loss on supervised data, which are observed solutions or measurements captured by sensors if they are available, (3) the loss on collocation data to embed the physical constraints expressed by the governing PDEs. Thanks to their simplicity and powerful capability when dealing with forward or inverse problems, PINNs are gaining more and more attention from researchers and they are gradually being improved and extended \citep{lu2021deepxde, wang2020understanding, jagtap2020adaptive, mcclenny2020self,mcclenny2021tensordiffeq,jagtap2020locally, jagtap2022deep1}. Besides that, there are continuing efforts to overcome the limitation of PINNs in terms of high computational costs \citep{jagtap2020conservative, jagtap2020extended, shukla2021parallel}. Thus far, the applicability of PINNs has been demonstrated in various fields of research. For example, \cite{kissas2020machine} employed PINNs for the modeling of cardiovascular flows, \cite{tartakovsky2020physics} and \cite{he2020physics} applied PINNs to subsurface flow and transport problems, \cite{shukla2021physics} used PINNs to identify polycrystalline nickel material coefficients. Recently, software editors (\textit{e.g.} NVIDIA) are also showing great interest for this kind of methods  \citep{hennigh2021nvidia, cai2021physics1}.

As far as computational fluid dynamics (CFD) is concerned, there has been a growing number of studies that focus on the capability of PINNs to infer the solution or estimate unknown physical parameters in problems of fluid mechanics \citep{raissi2020hidden, rao2020physics, reyes2021learning, zobeiry2021physics}. \cite{raissi2020hidden} employed a method that used PINNs to infer hidden physics states including the velocity and pressure fields from partial knowledge of a relevant variable (the concentration field in their work) by using the governing physical laws. This work demonstrated the effectiveness of PINNs to solve ill-posed problems and inspired many applications \citep{cai2021physics, cai2021flow, cai2021physics1, jin2021nsfnets, mao2020physics}. \cite{rao2020physics} proposed a new scheme for PINNs to model incompressible laminar fluid flows without using any supervised measurements. \cite{reyes2021learning} used PINNs to estimate the viscosity of non-Newtonian fluid flow by using only velocity measurements. Recently, \cite{mahmoudabadbozchelou2022nn} proposed the non-Newtonian PINNs for complex fluid flows with different non-Newtonian constitutive models. Through these previous studies, PINNs have shown a promising power to deal with ill-posed inverse problems with multi-physics features. 

In the industrial context, we often dispose of data that are captured from sensors. However, the impact of the location of these sensors on the performance of PINNs has not been paid much attention to. This is a crucial question in industrial use cases as the supervised measurements, which mostly depend on the placement of the sensors, are not randomly or uniformly distributed but only located at specific positions. Recently, increasing efforts have been made to study the optimal locations of the sensors \citep{jagtap2022deep, cai2021physics1}. Besides that, the distribution of unsupervised points (or collocation points) also plays an important role in the training process as the PDE residual is evaluated at these points. In most existing applications of PINNs, these collocation points are often chosen randomly inside the domain or generated using a Latin Hypercube Sampling strategy, which may cause large errors on the prediction at the points of discontinuity or at high gradient locations. This opens the opportunity to improve the accuracy of PINNs by taking the collocation points from some mesh that provides an \textit{a priori} knowledge on the location of discontinuity or high gradient location (\textit{e.g.} finite element mesh).

In this paper, we introduce a novel application of PINNs to infer the solution and identify unknown physical parameters in a non-Newtonian fluid thermo-mechanical problem. We consider an incompressible non-Newtonian fluid flow that is governed by a generalized Stokes equation coupled with a heat transfer equation. This set of PDEs is commonly used to model rubber manufacturing processes in the tire industry. First, we assess the performance of PINNs using virtual sensors coming from an industrial calendering problem simulation. To this end, we study the influence of the location of sensors and the distribution of unsupervised data on the accuracy of PINNs prediction in an ill-posed problem, where boundary conditions are not completely defined. More precisely, we suppose that only the measurements for the temperature are captured by the sensors, we aim to infer the velocity, the pressure, and the temperature at all points in the domain. Then, we examine the capability of PINNs to estimate unknown physical parameters from noisy measurements in an inverse problem. All our results are compared to a reference solution obtained by an in-house finite element solver. 

The following of this paper is organized as follows. In section 2, we briefly review the framework of PINNs, their scope of use and methods using adaptive activation functions and deep Kronecker networks to improve the accuracy of PINNs. We then study the influence of the position of supervised and unsupervised data on the accuracy of PINNs to solve a thermo-mechanical rubber calendering problem in section 3. The effect of noisy measurements is also investigated in this section. Finally, we summarize the conclusions in section 4.  

\section{Methodology}

In this section, the framework of Physics-Informed Neural Networks (PINNs) \citep{raissi2019physics} is briefly presented. Later in this section, the approaches using locally adaptive activation functions \citep{jagtap2020locally} and deep Kronecker networks with adaptive activation functions \citep{jagtap2022deep1} to improve the performance of PINNs are discussed.
\subsection{Physics-Informed Neural Networks}

PINNs aim at solving three main classes of problems: forward problems, inverse problems, and ill-posed problems for PDEs. For the forward problems the governing equations are known, \textit{i.e.} the physics laws (i.e. PDEs with boundary/initial conditions), and the physical parameters are well-defined. In this case, the goal is to infer the solution of the PDEs. For the inverse problems, some measurements for the PDEs solutions are available and some physical parameters are unknown. The goal is to identify the values of unknown parameters. For ill-posed problems, partial data associated with the PDEs solutions are available, but the initial and/or boundary conditions are not well-defined, and the purpose is to infer all the physical fields of interest.

To illustrate the methodology of PINNs, let us consider the following parameterized PDE defined on the domain $\Omega \subset \mathbb{R}^d$ with the boundary $\partial\Omega$:
\begin{align*}
    &\bm{u}_t + \mathcal{N}_{\bm{\mathrm{x}}}(\bm{u}, \bm{\lambda}) = 0, \text{ for } \bm{\mathrm{x}}\in \Omega, t\in [0, T]\\
    &\mathcal{B}(\bm{u},\bm{\mathrm{x}},t) = 0, \text{ for } \bm{\mathrm{x}} \in \partial \Omega\\
    &\bm{u}(\bm{\mathrm{x}}, 0) = g(\bm{\mathrm{x}}), \text{ for } \bm{\mathrm{x}}\in \Omega
\end{align*}
where $\bm{\mathrm{x}} \in \mathbb{R}^d$ and $t$ are the spatial and temporal coordinates, $\mathcal{N}_{\bm{\mathrm{x}}}$ is a differential operator, $\bm{\lambda}$ is the PDE parameter, $\bm{u}$ is the solution of the PDE with initial condition $g(\bm{\mathrm{x}})$ and boundary condition $\mathcal{B}$, which could be Dirichlet, Neumann, Robin, or periodic boundary conditions. The subscripts denote the partial differentiation in time or space.

In the conventional framework of PINNs \citep{raissi2019physics}, the solution $\bm{u}$ of the PDE is approximated by a fully-connected feedforward neural network $\mathcal{NN}$, which takes the spatial and temporal coordinates $(\bm{\mathrm{x}}, t)$ as inputs and the solution $\bm{u}$ as output, given a set of supervised data. This can be represented as follows:
\begin{align*}
    \bm{u} \approx \bm{\hat{u}} = \mathcal{NN}^{D}(\bm{\mathrm{x}}, t, \bm{\theta})
\end{align*}
where $\bm{\hat{u}}$ denotes the prediction value for the solution and $\bm{\theta}$ denotes the trainable parameters of the neural network. The neural network $\mathcal{NN}$ is composed of $D$ hidden layers. Let the hidden variable of the $l^{th}$ hidden layer be denoted by $\mathcal{NN}^l$ then the neural network can be expressed as:
\begin{align*}
    &\mathcal{NN}^0 = (\bm{\mathrm{x}},t)\\
    &\mathcal{NN}^l=\sigma(\bm{W^l}\mathcal{NN}^{l-1} + \bm{b^l}) \text{ for } 1\le l\le D-1\\
    &\mathcal{NN}^D = \bm{W^D}\mathcal{NN}^{D-1} + \bm{b^D}
\end{align*}

For the problems of inferring the PDE solutions, the trainable parameters $\bm{\theta}$ are defined by the weights $\{\bm{W^l}\}_{l=1}^D$ and the biases $\{\bm{b^l}\}_{l=1}^D$ of the neural networks. For the problems of identification, these parameters include, in addition, the PDEs parameters that need to be defined.  The parameters of the neural network are trained by minimizing the cost function $L$:
\begin{align}
    L = L_{pde} + L_{ic} + L_{bc} + L_{data} \label{loss}
\end{align}
where the terms $L_{pde},L_{ic},L_{bc}$ and $L_{data}$ penalize the loss in the residual of the PDE, the initial condition, the boundary condition, and the supervised data (measurements), respectively, which can be represented as follow:
\begin{align*}
    L_{pde} &= \dfrac{1}{N_{pde}}\sum_{i=1}^{N_{pde}}|\hat{\bm{u}}_{t_{i}} + \mathcal{N}_{\bm{\mathrm{x}}_{i}}(\hat{\bm{u}}_{i}, \bm{\lambda})|^2\\
    L_{ic} &= \dfrac{w_{ic}}{N_{ic}}\sum_{i=1}^{N_{ic}}|\bm{\hat{u}}(\bm{\mathrm{x}}_i,0) - g(\bm{\mathrm{x}}_i)|^2\\
    L_{bc} &= \dfrac{w_{bc}}{N_{bc}}\sum_{i=1}^{N_{bc}}|\mathcal{B}(\bm{\hat{u}}_{i}, \bm{\mathrm{x}}_{i}, t_{i})|^2\\
    L_{data} &= \dfrac{w_{data}}{N_{data}}\sum_{i=1}^{N_{data}}|\bm{\hat{u}}(\bm{\mathrm{x}}_i,t_i) - \bm{u}(\bm{\mathrm{x}}_i,t_i)|^2\\
\end{align*}
where $N_{ic}, N_{bc}, N_{data}$ denote the numbers of learning points for the initial condition, boundary condition, and measurements, respectively, and $N_{pde}$ denotes the number of residual points (or collocation points or unsupervised points) of the PDE. $w_{ic}, w_{bc}$ and $w_{data}$ are the weight coefficients for different loss terms, which are used to adjust the contribution of each term to the cost function. These coefficients can be either pre-specified before the training or tuned during the training. Recently, many studies aim at improving the efficiency of PINNs by using adaptive weights in the cost function \citep{wang2020understanding, wang2020and, mcclenny2020self}. We note that the number $N_{data}$ depends on the supervised data while the numbers $N_{ic}, N_{bc}$ (if the initial and boundary conditions are well-defined) and $N_{pde}$ are user-defined. To construct the residual loss $L_{pde}$, automatic differentiation (AD) \citep{baydin2018automatic} is used to compute the partial derivatives of the output $\bm{\hat{u}}$ of the network with respect to the inputs $(\bm{\mathrm{x}}, t)$. AD builds a computation graph to calculate the gradients numerically. The benefit of using AD to compute the derivatives is that we do not need to discretize the domain, which is essential in classical numerical methods.

The definition of the cost function (\ref{loss}) depends on the problems. For example, for forward and well-posed problems where the initial and boundary conditions of the PDE are well-defined, the loss term for the measurements $L_{data}$ can be eliminated. However, for ill-posed problems, i.e. the initial and boundary conditions are unknown and the terms $L_{ic}, L_{bc}$ dissolve, then the term $L_{data}$ becomes mandatory for the neural network to approximate the solutions. We also note that PINNs may provide different performances with different network initializations. In this work, when comparing results of the same configuration of PINNs (\textit{i.e.} the same networks architecture, activation function and optimizer), we train PINNs from random initialization five times and then choose the model with the smallest training loss as the final solution for both visualization and numerical comparison. When comparing results of different configuration of PINNs, we train PINNs from random initialization five times and then choose the model with the smallest training loss as the final solution for the visualization, and choose the mean and standard deviation values of the five models for numerical comparison.

\subsection{Locally adaptive activation functions}
\cite{jagtap2020locally} proposed two approaches using layer-wise (L-LAAFs) and neuron-wise (N-LAAFs) locally adaptive activation functions to improve the performance of deep and physics-informed neural networks. These methods have been proven to be a very efficient way to train the network and have been employed successfully in many applications \citep{markidis2021old, gnanasambandam2022self, bai2022application}. Here, we briefly present the formulation of L-LAAFs and N-LAAFs which closely follows the framework introduced by \cite{jagtap2020locally}. These methods introduce a scalable parameter for each layer (layer-wise) or for every neuron (neuron-wise) separately. A slope recovery term is also added in the cost function to accelerate the training process.

\begin{itemize}
    \item Layer-wise locally adaptive activation functions (L-LAAFs): in the $l^{th}$ layer of the neural network, we add an adaptive parameter $a^l$ scaled by a pre-specified factor $n$ as: \begin{align*}
        \mathcal{NN}^l=\sigma(na^l(\bm{W^l}\mathcal{NN}^{l-1} + \bm{b^l})) \quad \text{ for } l=1,\dots D-1
    \end{align*}
    \item Neuron-wise locally adaptive activation functions (N-LAAFs): in the $l^{th}$ layer with $N_l$ neurons, we add adaptive parameters $a^l_i$ scaled by a pre-specified factor $n$ for each neuron as: \begin{align*}
        \mathcal{NN}^l=\sigma(na^l_i(\bm{W^l}\mathcal{NN}^{l-1} + \bm{b^l})_i)  \quad \text{ for } l=1,\dots, D-1; i=1, \dots, N_l
    \end{align*}
\end{itemize}

The trainable parameters of the neural networks $\bm{\theta}$ is constituted of $\{\bm{W^l}, \bm{b^l}\}_{l=1}^D$ and $ \{\bm{a^l}\}_{l=1}^{D-1}$ for L-LAAFs or $ \{\bm{a^l_i}\}_{l=1}^{D-1} \text { } \forall i\in\{1,N_l\}$ for N-LAAFs. The scaling factor $n$ and the adaptive parameters $a^l$ (or $a^l_i$) are often initialized as $na^l=1$ (or $na^l_i=1$), so that the initialization of LAAFs is the same as the standard fixed activation function. A slope recovery term $\mathcal{S}_a$ is added to the cost function in order to force the neural network to increase the value of activation slope quickly and thus increase the training speed:
\begin{align*}
    L = L_{pde} + L_{ic} + L_{bc} + L_{data} +\mathcal{S}_a
\end{align*} where $S_a$ is defined as:
\begin{align*}
    \mathcal{S}_a = \begin{cases}
    &\dfrac{1}{\frac{1}{D-1}\sum_{l=1}^D\exp{(a^l)}} \text{ for L-LAAFs}\\
    &\dfrac{1}{\frac{1}{D-1}\sum_{l=1}^D\exp{(\frac{\sum_{i=1}^{N_l}a^l_i}{N_l})}} \text{ for N-LAAFs}
    \end{cases}
\end{align*}

We note that introducing additional trainable parameters for each layer or each neuron will lead to higher computational cost compared to vanilla PINNs or the method using the global adaptive activation function introduced in \citep{jagtap2020adaptive}.  However, the computational cost remains comparable between these methods as the depth and width of the networks increase. Furthermore, adding additional parameters in L-LAAFs and N-LAAFs gives the neural networks more flexibility (as higher degrees of freedom) and thus increases the capability of training.

\subsection{Deep Kronecker neural networks with adaptive activation functions}
Kronecker neural networks (KNNs) \citep{jagtap2022deep1} is a general framework for neural networks with adaptive activation functions. KNNs employ the Kronecker product to construct the weight matrices and bias vectors. A detailed description of the method can be found in the cited reference. Here we present the simplest formulation of KNNs, which can be viewed as a generalization of the standard feedforward networks. More precisely, in the $l^{th}$ layer of KNNs, we introduce trainable parameters $\{\alpha_k^l,\omega_k^l\}_{k=1}^K$ as:
\begin{align*}
    \mathcal{NN}^l = \sum_{k=1}^K \alpha_k^l\sigma_k(\omega_k^l(\bm{W^l}\mathcal{NN}^{l-1} + \bm{b^l})) \quad \text{ for } l=1,\dots, D-1
\end{align*}
where $\{\sigma_k\}_{k=1}^K$ are fixed activation functions. We note that the feedforward network is a special case of KNNs when $K=1$ and $\alpha_k^l,\omega_k^l=1 \text{ } \forall l,k$. When $K=1$ and $\alpha_k^l=1 \text{ } \forall l,k$, KNNs can be seen as a feedforward network with layer-wise locally adaptive activation functions (L-LAAFs). When $K$ increases, the number of trainable parameters also increases, which helps to improve the training capability but however leads to high computation cost in KNNs.

In the original work of \cite{jagtap2022deep1}, the authors proposed a particular network of KNNs named as Rowdy neural network (Rowdy Net), which uses sinusoidal functions as activation functions to capture high-frequency components in the target function:
\begin{align*}
    \sigma_k(x) = n\sin((k-1)nx) \quad \text{ for } k=2,\dots, K
\end{align*}
where $n$ is a scaling factor and $\sigma_1$ can be any standard activation function. In this work, we choose $\sigma_1(x)=\tanh(x)$. We will compare Rowdy Net with the KNNs using only \textit{tanh} activation functions $\sigma_k(x)=tanh(x)$ for $k=1,\dots, K$. We refer to this network as KNNs-tanh in the following.
\section{Application to rubber calendering modeling}

\subsection{Problem configuration}

\begin{figure}[H]
\centering
\includegraphics[width=7cm]{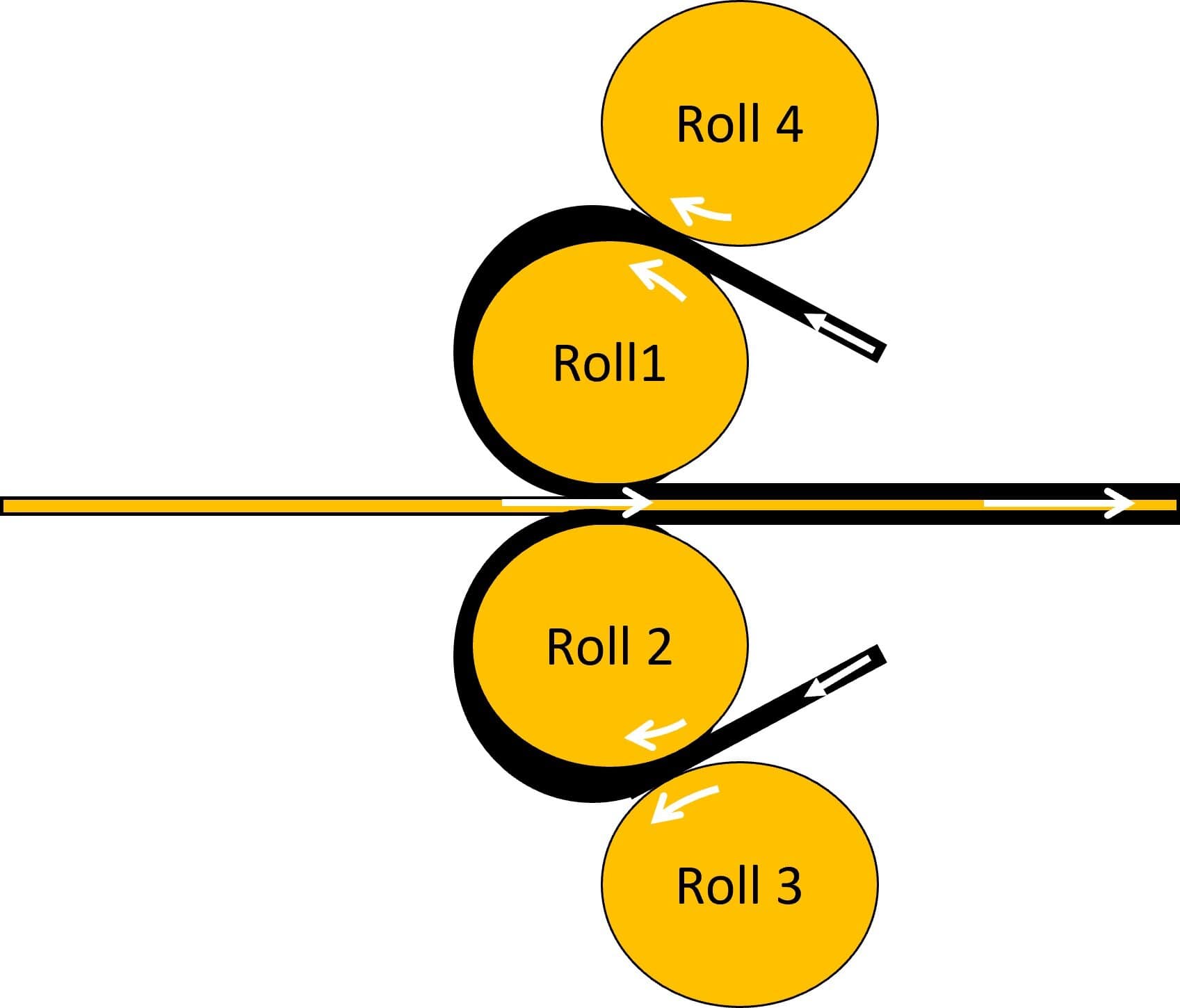}
\caption{Sketch of a calender used to create thin tissues of rubber for tire manufacturing. The rubber material is colored in black for illustration.}
\label{calender}
\end{figure}

\begin{figure}[H]
    \centering
    \includegraphics[width=10cm]{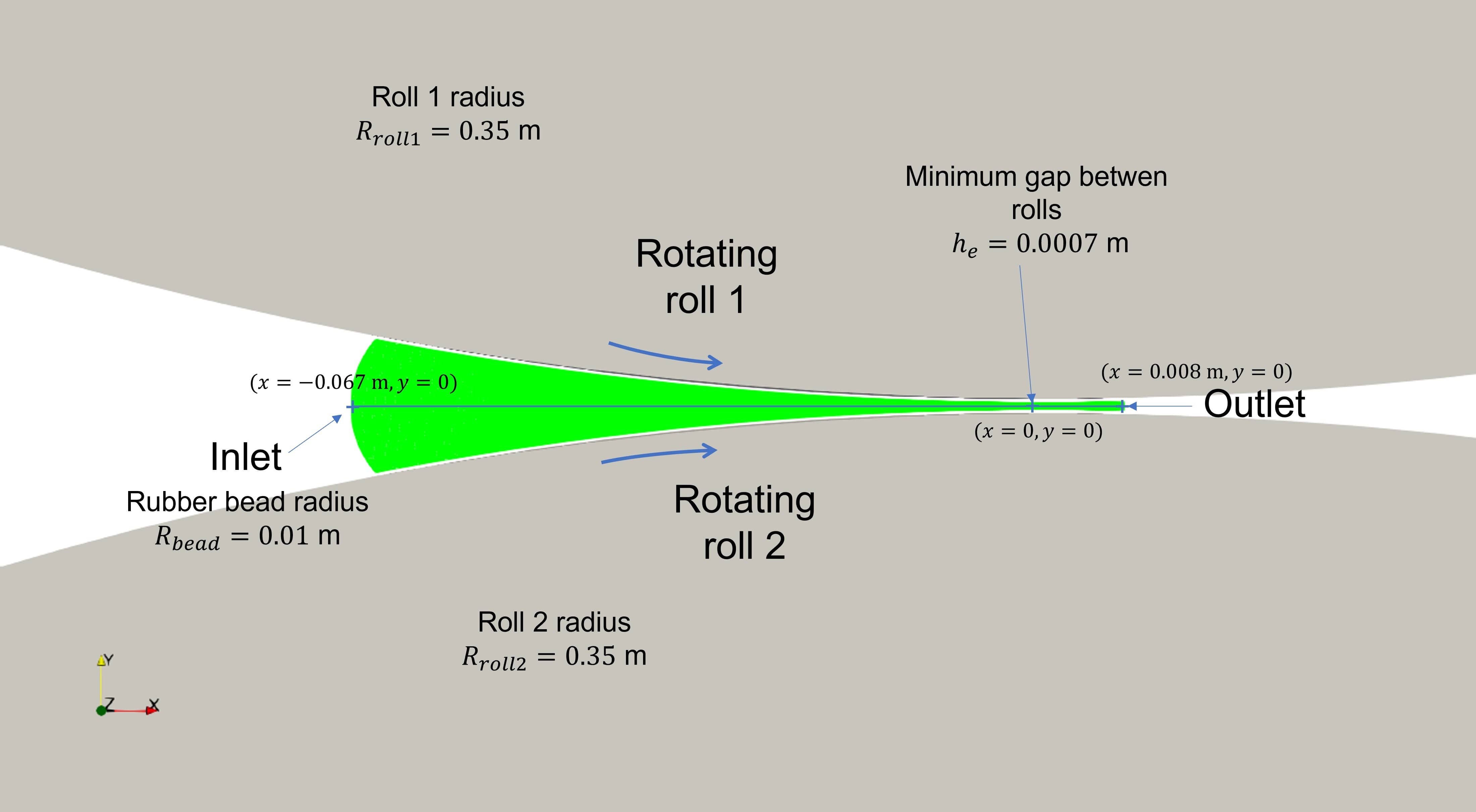}
    \caption{Sketch of the 2D configuration and geometrical setup considered in the present study. The physical domain of interest is colored in green.}%
    \label{calender2d}%
\end{figure}

In the industry of tire manufacturing, calendering is a mechanical process used to create and assemble thin tissues of rubber. As sketched in Figure \ref{calender}, during calendering, the rubber is smoothed out through several rotating cylindrical rolls. The modeling of the entire calendering process is a challenging task involving different length and time scales, different materials, and potentially 3D effects. However, for the sake of simplicity, the goal of the present study is only to model the 2D temperature, velocity, and pressure fields inside rubber materials going through two contra-rotating rolls of the calender as depicted in Figure \ref{calender2d}. 

The physical domain of interest together with the geometrical configuration of the study are presented in Figure \ref{calender2d}: the 2D domain is simply bounded by an inlet, two rotating boundaries corresponding to the edges of the rolls, and an outlet. From an industrial point of view, it is of prime interest to monitor the rubber temperature inside the domain. In particular, because rubber materials are subjected to viscous heating, an important temperature elevation can be observed in highly-sheared regions (specifically at the inlet where the material can be accumulated in a rubber bead and in the very thin gap between the rolls). Such high temperatures can lead to unexpected rubber curing that could deteriorate the final material quality. In the next section, the physical equations used to model the rubber mechanical and thermal fields will be introduced.

\subsection{Problem formulation}

From the physical point of view, the rubber is assimilated as an incompressible non-Newtonian fluid flow in the present study (in particular, the elastic part of the material is not considered here). Moreover, only the steady-state regime is considered. Consequently, the equations describing the behaviour of the rubber are a generalized Stokes equation for momentum conservation, a divergence-free velocity equation for mass conservation, and a convection-diffusion equation with a viscous heating term for energy conservation. The constitutive law considered in all the presented cases is a power-law combined with an Arrhenius law to take into account the dependence of viscosity on temperature. The governing system of PDEs can be written as follows:
\begin{align}
    &\nabla.(2\eta(\bm{\vec{u}}, T)\bar{\bar{\epsilon}}(\bm{\vec{u}})) - \vec{\nabla}p =\vec{0} \label{mouveconv_ori} \\
    &\nabla.\bm{\vec{u}} =0 \label{massconv_ori} \\
    &\bm{\vec{u}}.\vec{\nabla}T = \dfrac{\lambda}{\rho C_p}\nabla^2T + \dfrac{1}{\rho C_p}\eta(\bm{\vec{u}}, T)|\gamma(\bm{\vec{u}})|^2 \label{enerconv_ori}
\end{align}
where $\bm{\vec{u}}=(u_x, u_y)^T$ is the velocity vector, $p$ is the pressure, $T$ is the temperature, $\lambda=20 ~ W.m^{-1}.K^{-1}$ is the thermal conductivity, $\rho=1000 ~ kg.m^{-3}$ is the rubber density, $C_p=1000 ~ J.K^{-1}$ is the heat capacity at 25°C. The strain-rate tensor $\bar{\bar{\epsilon}}$ is defined as: 
\begin{align*}
    \bar{\bar{\epsilon}}(\bm{\vec{u}})=\dfrac{1}{2}\begin{pmatrix}
    2\dfrac{\partial u_x}{\partial x} & \dfrac{\partial u_x}{\partial y} +  \dfrac{\partial u_y}{\partial x} \\ \dfrac{\partial u_x}{\partial y} +  \dfrac{\partial u_y}{\partial x}& 2\dfrac{\partial u_y}{\partial y} \end{pmatrix}
\end{align*}
The dynamic viscosity $\eta$ is defined as:
\begin{align*}
    \eta(\bm{\vec{u}}, T)= K|\tau\gamma(\bm{\vec{u}})|^{n-1}\exp(\dfrac{E_{\alpha}}{R}(\dfrac{1}{T}-\dfrac{1}{T_{\alpha}}))
\end{align*}
where $K=2\times10^5 ~ Pa.s^n$ is the consistency, $\tau=1 ~ s$ is the characteristic time, $n=0.2$ is the power law exponent, $E_{\alpha}=7000 ~ J.mol^{-1}$ is the activation energy of the rubber, $R$ is the perfect gas constant, and $\gamma(\bm{\vec{u}})=\sqrt{2\sum_{i,j}\bar{\bar{\epsilon}}_{i,j}^2}$ is the generalized strain rate.

We consider the following quantities of reference: a length $L_0=0.01 ~m$, a velocity $u_0=0.5 ~m.s^{-1}$, two temperatures $T_1= 393.15 ~K, T_2= 476.75 ~K$, and for the pressure, we choose $p_0=K\dfrac{u_0}{L_0}=10^7 ~ Pa$. The corresponding dimensionless variables are: $\tilde{u}_x=\dfrac{u_x}{u_0}$, $\tilde{u}_y=\dfrac{u_y}{u_0}$, $\tilde{T}=\dfrac{T - T_1}{T_2-T_1}$, $\tilde{x}=\dfrac{x}{L_0}$, $\tilde{y}=\dfrac{y}{L_0}$. Since the dimension of the consistency $K$ is a function of $n$, the quantity $\tilde{p}=\dfrac{p}{p_0}$ is still a dimensional number.

The system of PDEs is rewritten as:
\begin{align}
     &\tilde{\nabla}.\Big(2\tilde{\eta}(\bm{\vec{\tilde{u}}}, \tilde{T})\tilde{\bar{\bar{\epsilon}}}(\bm{\vec{\tilde{u}}})\Big) - \vec{\tilde{\nabla}}\tilde{p} =\vec{0}\label{mouveconv}\\
     &\tilde{\nabla}.\bm{\vec{\tilde{u}}} = 0 \label{massconv}\\
     &\bm{\vec{\tilde{u}}}\vec{\tilde{\nabla}}\tilde{T} = \dfrac{1}{Pe}\tilde{\nabla}^2\tilde{T} + \dfrac{Br}{Pe}\tilde{\eta}(\bm{\vec{\tilde{u}}}, \tilde{T})|\tilde{\gamma}(\bm{\vec{\tilde{u}}})|^2\label{enerconv}
\end{align}
where the Péclet number $Pe=\dfrac{\rho C_pu_0L_0}{\lambda}$ and the Brinkman number $Br=\dfrac{Ku_0^2}{\lambda(T_2-T_1)}$ are dimensionless. The dimensionless dynamic viscosity $\tilde{\eta}$ is defined as follows:
\begin{align*}
    \tilde{\eta}(\bm{\vec{\tilde{u}}}, \tilde{T})= |\tilde{\tau}\tilde{\gamma}(\bm{\vec{\tilde{u}}})|^{n-1}\exp(\dfrac{E_{\alpha}}{R}(\dfrac{1}{(T_2-T_1)\tilde{T}+T_1}-\dfrac{1}{T_{\alpha}}))
\end{align*}
For the rest of this paper, we shall only consider the dimensionless PDE system in the PINNs training process. 

\subsection{Reference solutions and validation framework}

\begin{figure}[H]
    \centering
    \subfloat[Computational domain and boundary conditions]{\includegraphics[width=0.9 \textwidth]{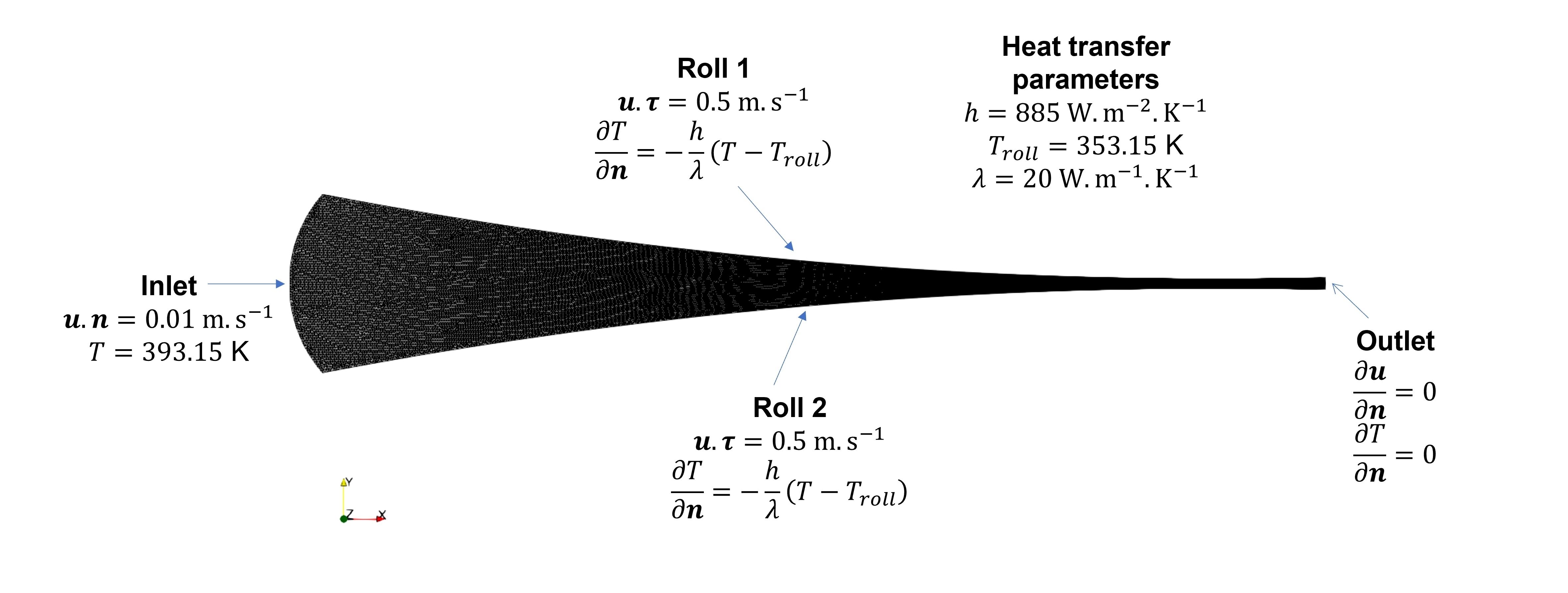}} \\
    \subfloat[Mesh close to the inlet]{\includegraphics[width=0.3 \textwidth]{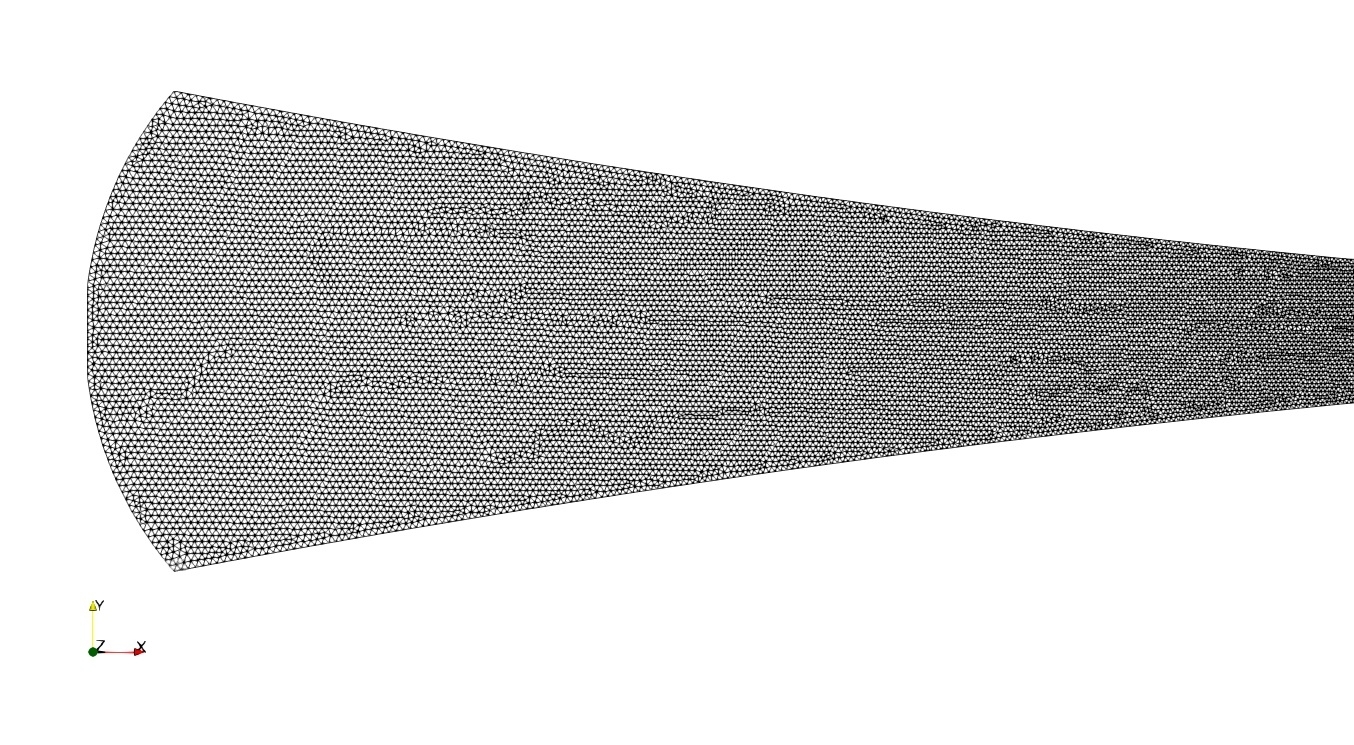}} ~~
    \subfloat[Mesh close to the outlet]{\includegraphics[width=0.3 \textwidth]{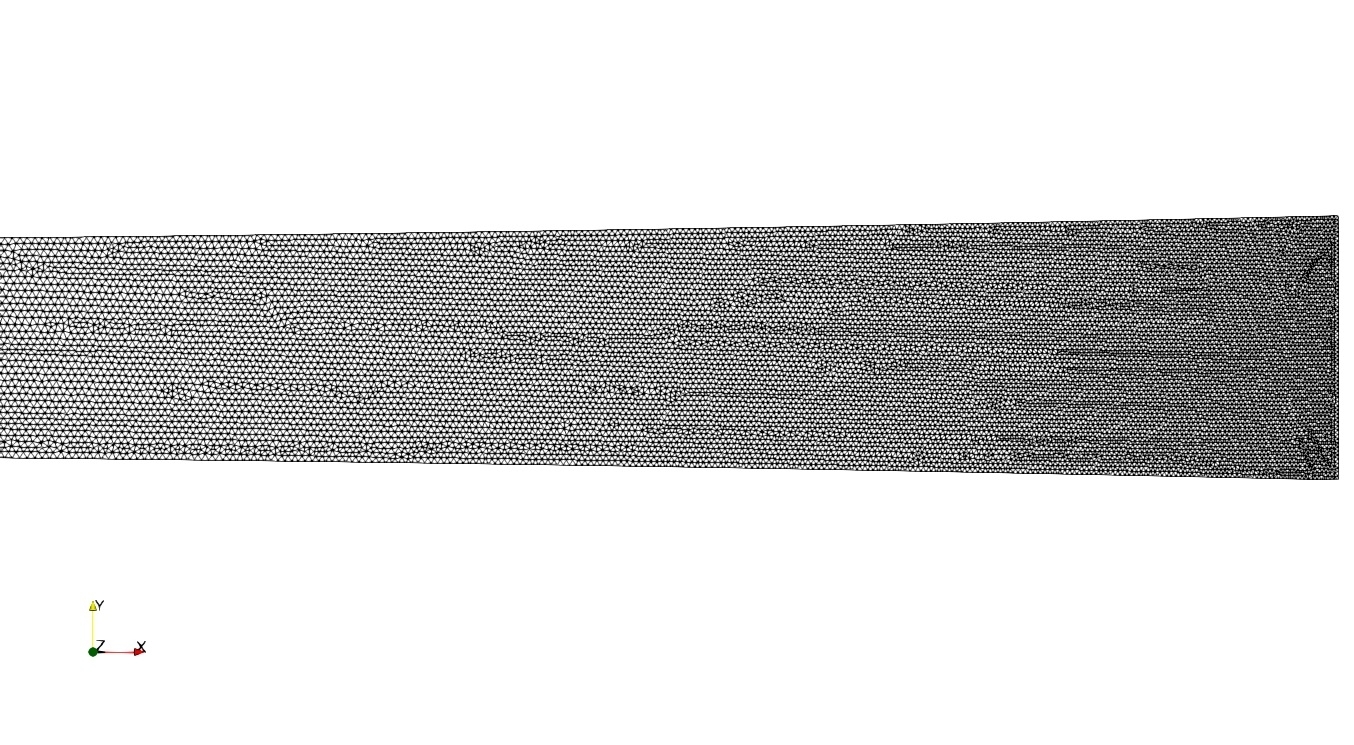}}
    \caption{Sketch of the computational domain (the mesh is colored in black) with the velocity and temperature boundary conditions used in the reference finite element simulation: $\mathbf{n}$ and $\mathbf{\bm{\tau}}$ denote respectively the normal and tangential vectors with respect to the corresponding boundary.}%
    \label{bc_calender}%
\end{figure}

To generate reference High-Fidelity (HF) solutions of velocity, pressure, and temperature fields, the equations (\ref{mouveconv}), (\ref{massconv}), (\ref{enerconv}) are discretized and solved using an in-house generic and multi-purpose finite element solver named MEF++ and co-developed by Laval University and Michelin \citep{mef, guenette2004}. The velocity and temperature boundary conditions prescribed in the finite element simulation are given in Figure \ref{bc_calender}. As the first aim of this study is to assess the impact of the location of supervised points on the performance of PINNs, data captured from real sensors would be very valuable. However, obtaining different scenarios of the placement of the sensors in a real calender is an expensive and impractical process for a research study. Therefore, in our work, we use the HF solutions provided by the finite element solver as virtual measurements obtained by sensors. To evaluate the performance of PINNs in predicting physical fields, we calculate the relative $\mathcal{L}^2$ error defined as follows:
\begin{align*}
    \epsilon_v = \dfrac{||v-\hat{v}||_2}{v_{max} - v_{min}}
\end{align*}
where $v$ denotes the reference simulated field of interest and $\hat{v}$ is the corresponding PINNs prediction. The choice is made to divide the absolute error by the reference field amplitude instead of the reference field L2 norm in order to avoid any artificial high values of the error for fields very close to zero (see for example $u_y$ in Figure \ref{infer_sol_ref}). In inverse problems where the goal is to identify unknown scalar parameters, we calculate the relative error as follows:
\begin{align*}
    \epsilon_\mu = \dfrac{(\mu-\hat{\mu})^2}{\mu^2}
\end{align*}
where $\mu$ denotes the reference value of the unknown parameter and $\hat{\mu}$ is the corresponding PINNs prediction.

\begin{figure}[H]
    \centering
    \subfloat[Case (1): random mesh]{{\includegraphics[width=6.5cm]{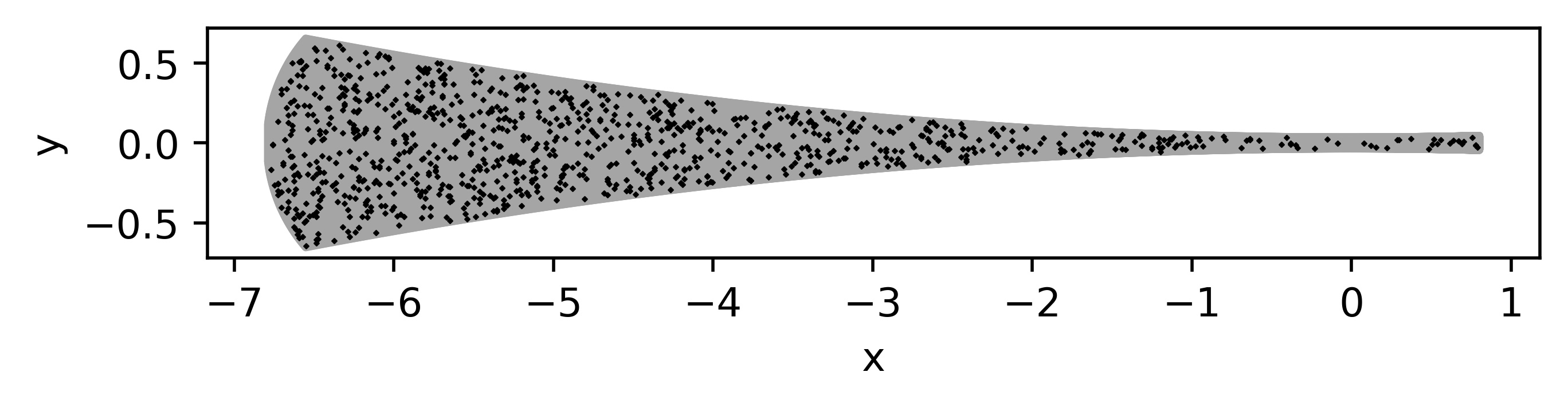} \label{calender_geo_colloc0}}}%%
    \subfloat[Case (2): finite element mesh]{{\includegraphics[width=6.5cm]{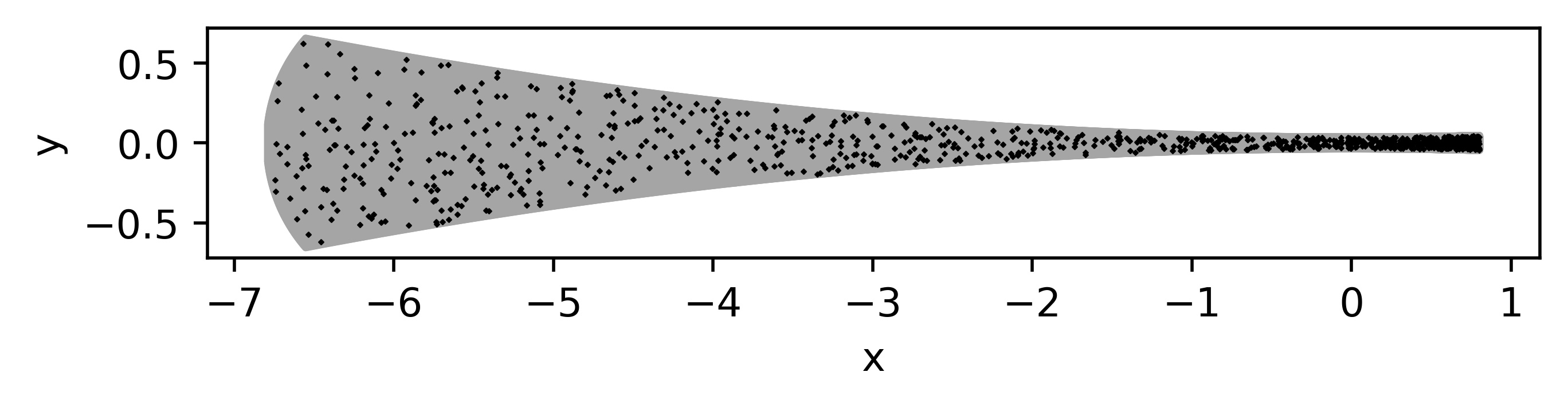} \label{calender_geo_colloc1}}}%%
    \caption{\textit{Visualization of $1,000$ random points on the meshes}}%
    \label{calender_geo_colloc}%
\end{figure}

\begin{figure}[H]
    \centering
    \subfloat[Case (1): random mesh]{{\includegraphics[width=5.5cm]{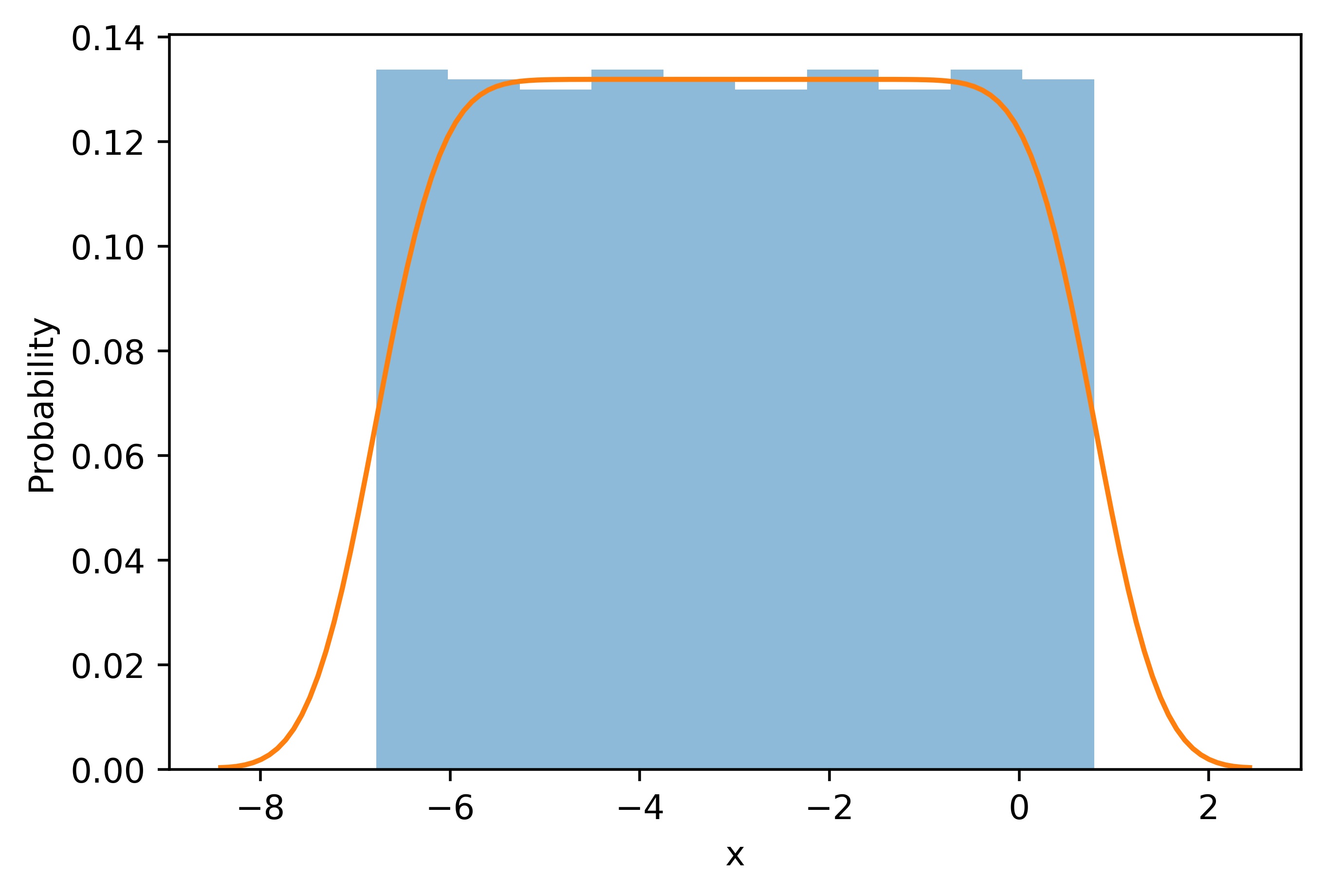} \label{calender_dens_rand}}}%%
    \subfloat[Case (2): finite element mesh]{{\includegraphics[width=5.4cm]{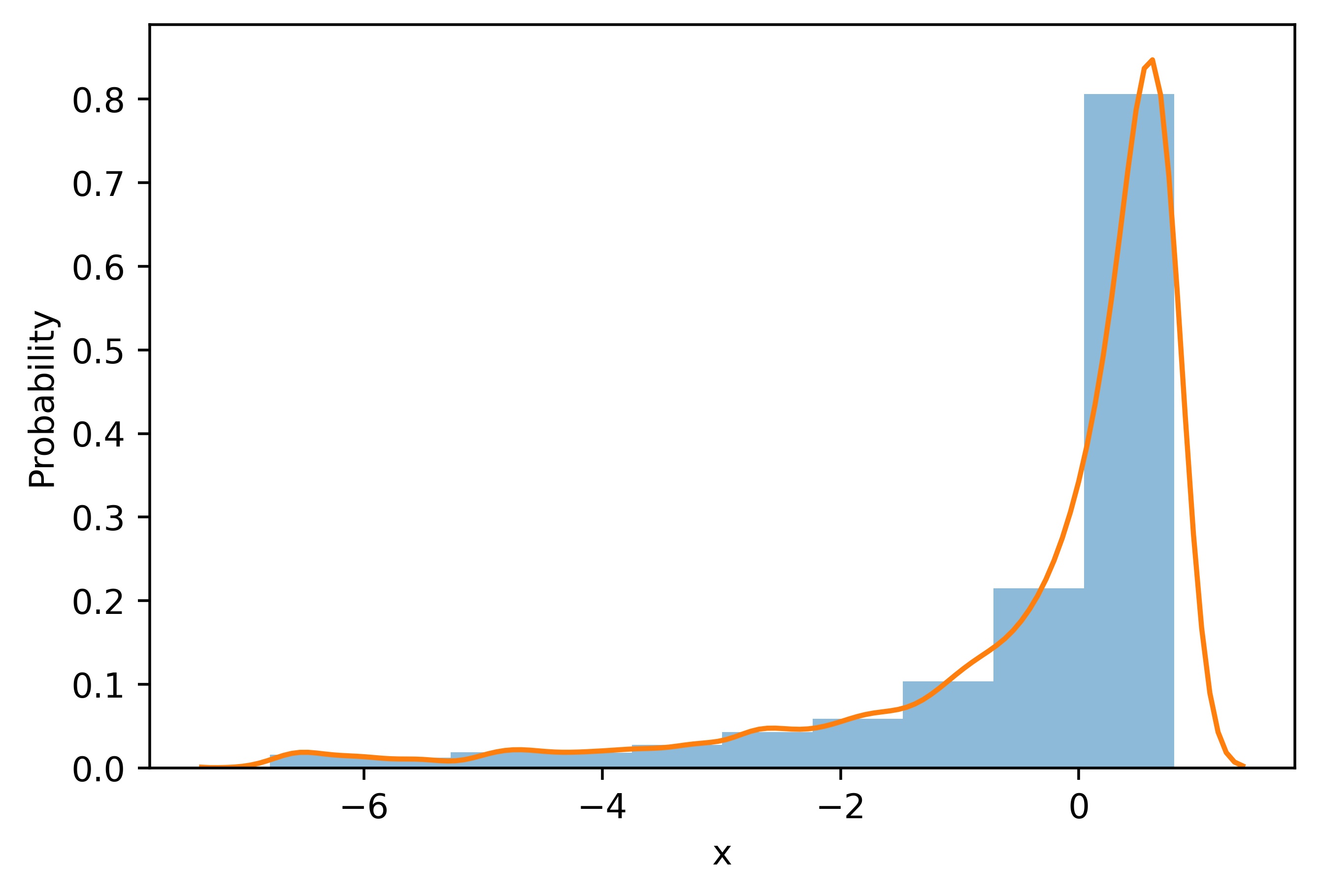} \label{calender_dens_fe}}}%%
    \caption{\textit{Density of points on the line $y=0$ for each mesh.}}%
    \label{calender_dens}%
\end{figure}

In our work, the errors are evaluated on the finite element mesh, which provides us an \textit{a priori} knowledge on the high gradient location. The number of points on the mesh is $N=163,596$. The in-house solver gives us the high-fidelity solution on the finite element mesh. Figure \ref{calender_geo_colloc} shows the visualization of $1,000$ random points on this mesh compared to $1,000$ randomly uniformly distributed points inside the domain. Figure \ref{calender_dens} illustrates the density of points on the line $y=0$ for each case. We point out that when the points are taken from the finite element mesh, the density of points is higher at the output than at the input of the calender. This is because the finite element mesh is refined to take into account the very thin gap between the rolls close to the domain outlet.

It is worth mentioning that before investigating the problem of calendering, PINNs implementation has been validated in a simple non-Newtonian laminar pipe flow configuration presented in \ref{sec_analytic_pipe}.

\subsection{Inferring physical fields of interest from sensor measurements}\label{sec_calender_infer}

In this section, we suppose that the boundary conditions of the problem are not completely defined, and we dispose of some measurements of the temperature from virtual sensors (in our case, coming from our finite element simulation). The goal is to infer the pressure, velocity, and temperature fields at all points in the domain. We note that here, no information on the pressure field is given, but only its gradient in the PDE residual is. Thus the pressure is only identifiable up to a constant.

\subsubsection{Impact of sensor location}
\begin{figure}[H]
    \centering
    \subfloat[Case 1]{{\includegraphics[width=5.4cm]{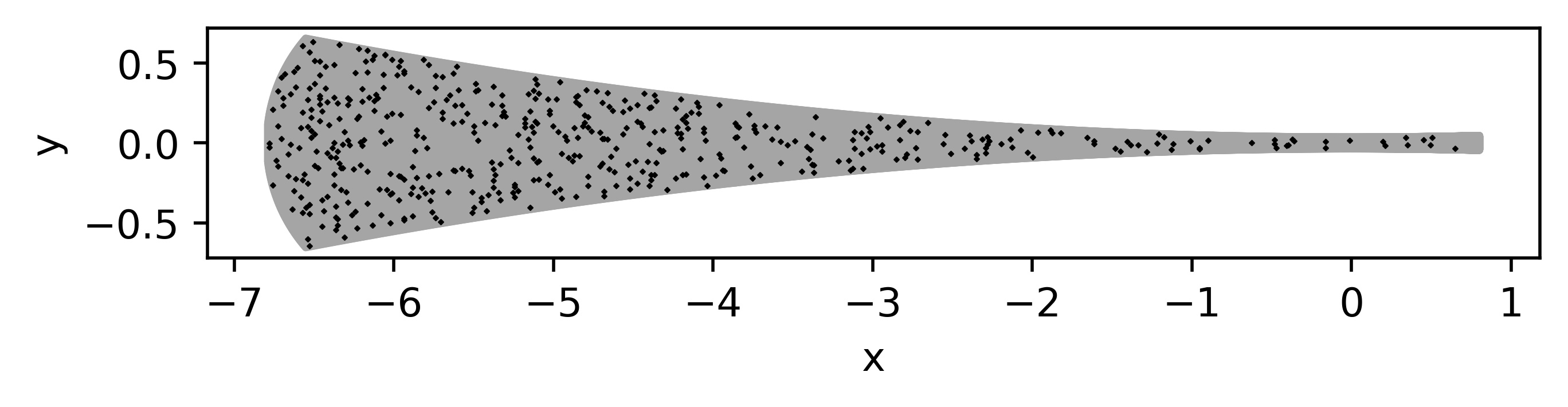} \label{calender_geo_sensor0}}}%%
    \subfloat[Case 2]{{\includegraphics[width=5.4cm]{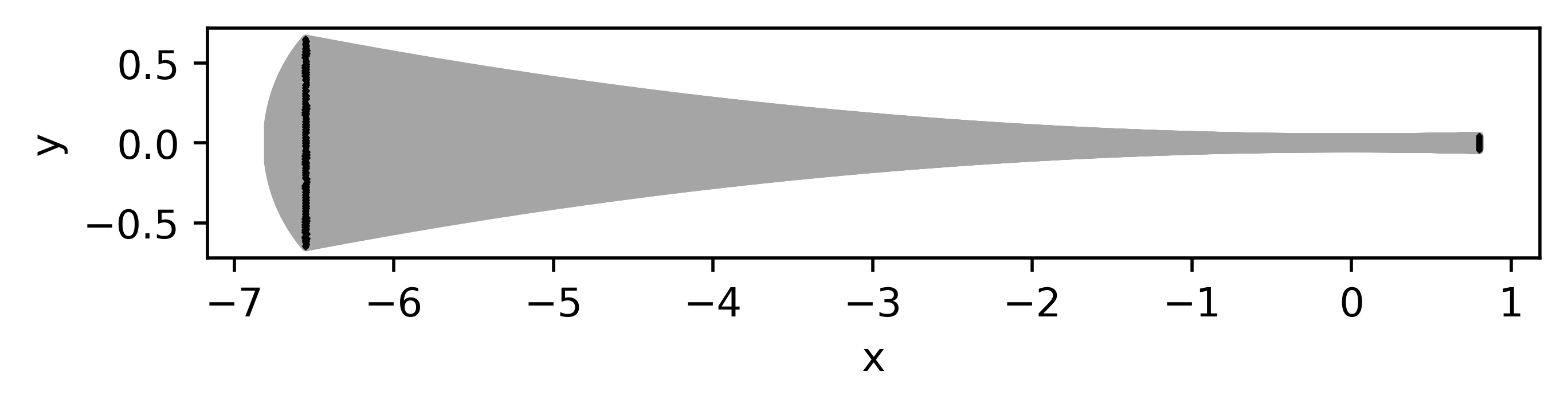} \label{calender_geo_sensor1}}}%%
    \subfloat[Case 3]{{\includegraphics[width=5.4cm]{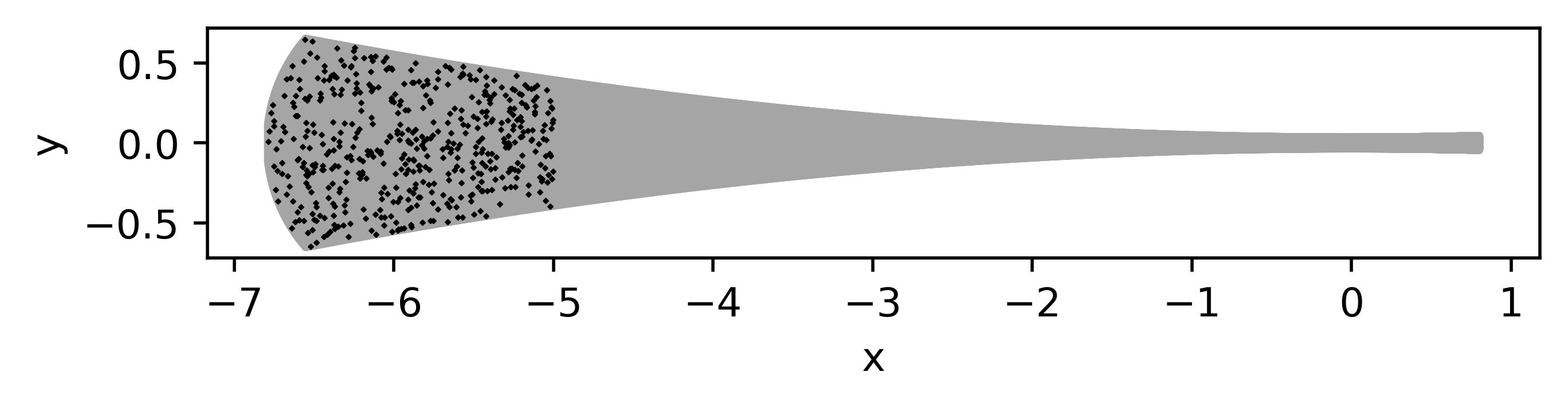} \label{calender_geo_sensor6}}}%%
    \caption{\textit{Geometry of the rubber calender and the location of sensors:} the black points represent the location of sensors (or supervised points for the temperature).}%
    \label{calender_geo_sensor}%
\end{figure}

\begin{figure}[H]
    \centering
    \subfloat[Case 1]{{\includegraphics[width=5cm]{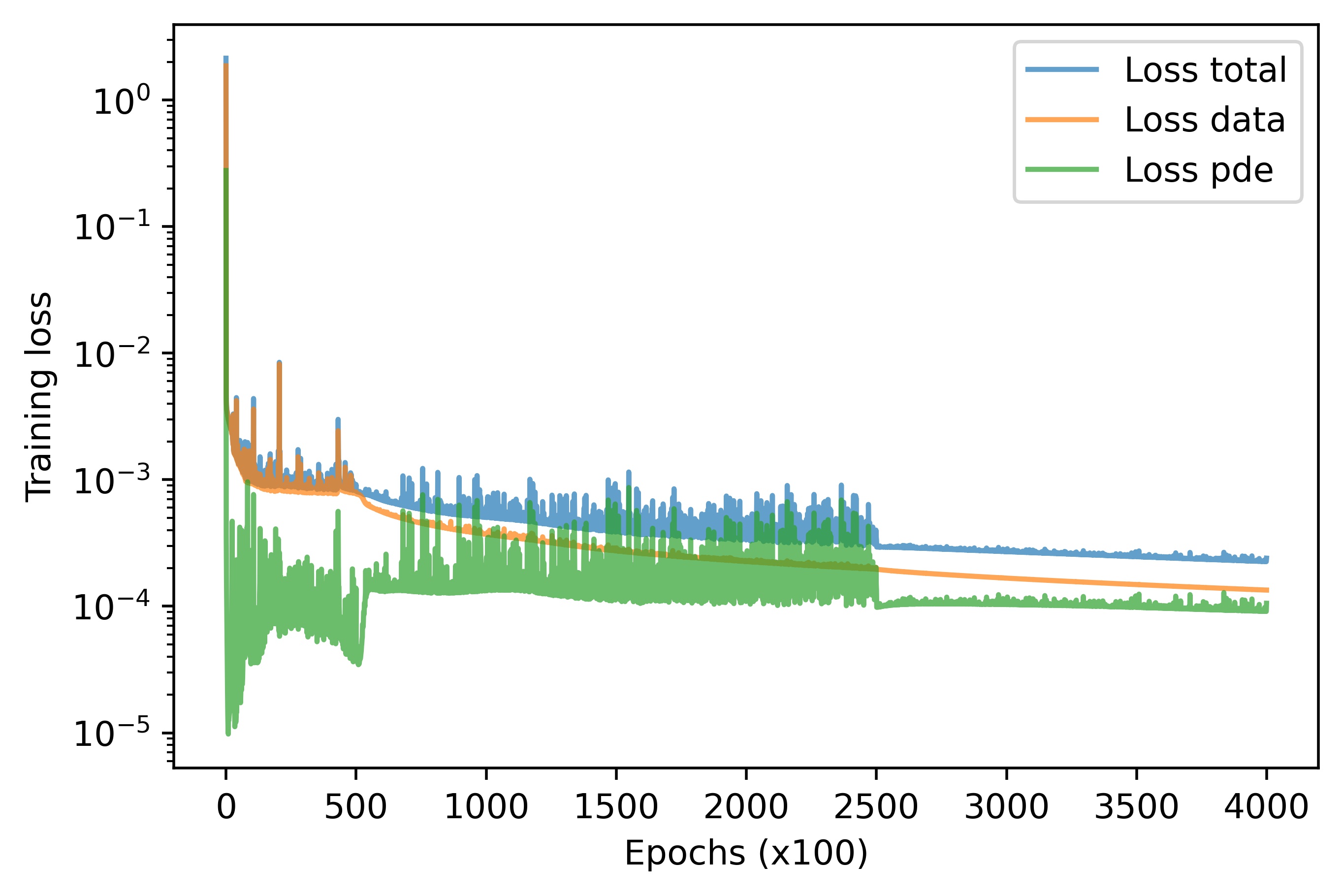} }}%%
    \subfloat[Case 2]{{\includegraphics[width=5cm]{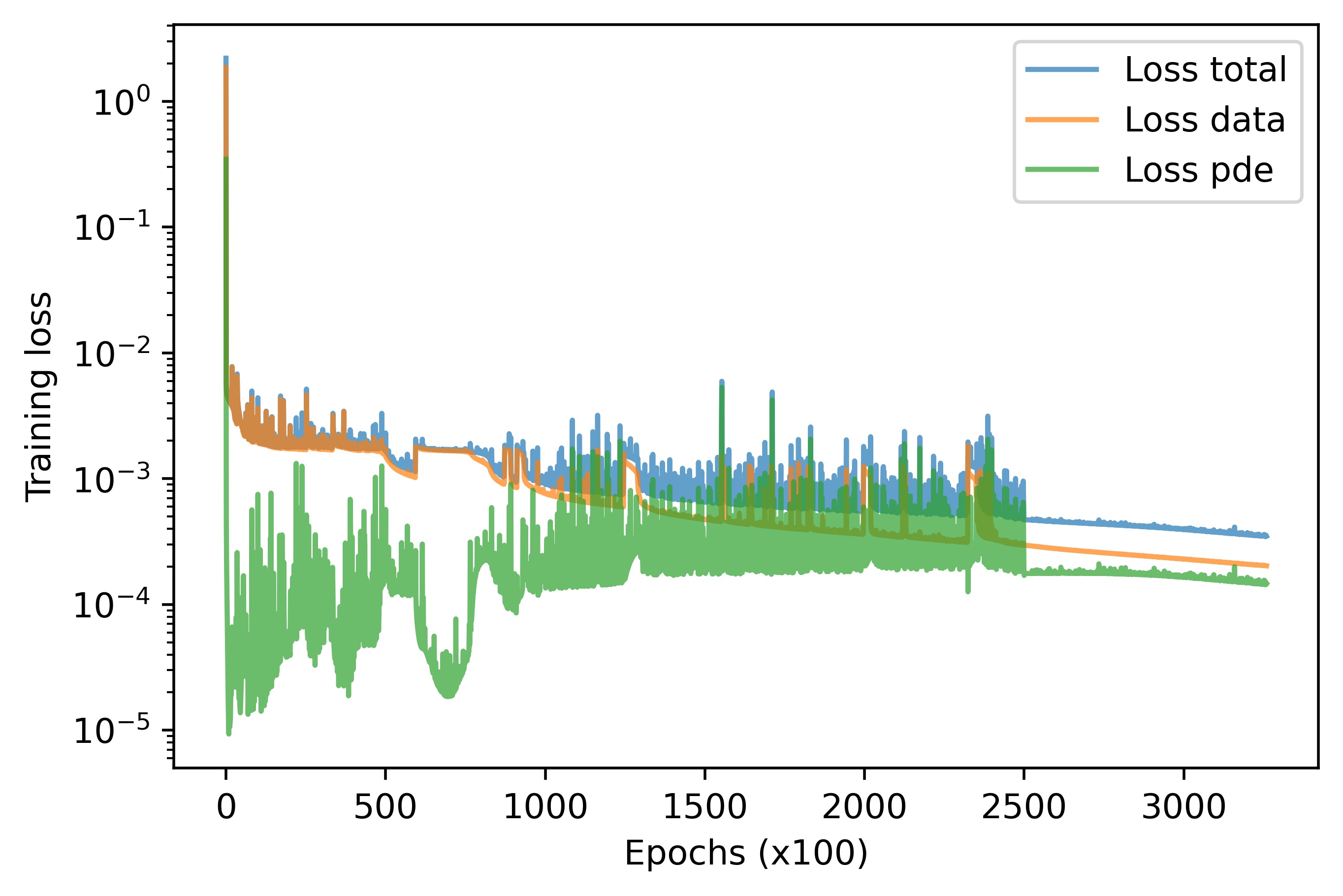} }}%
    \subfloat[Case 3]{{\includegraphics[width=5cm]{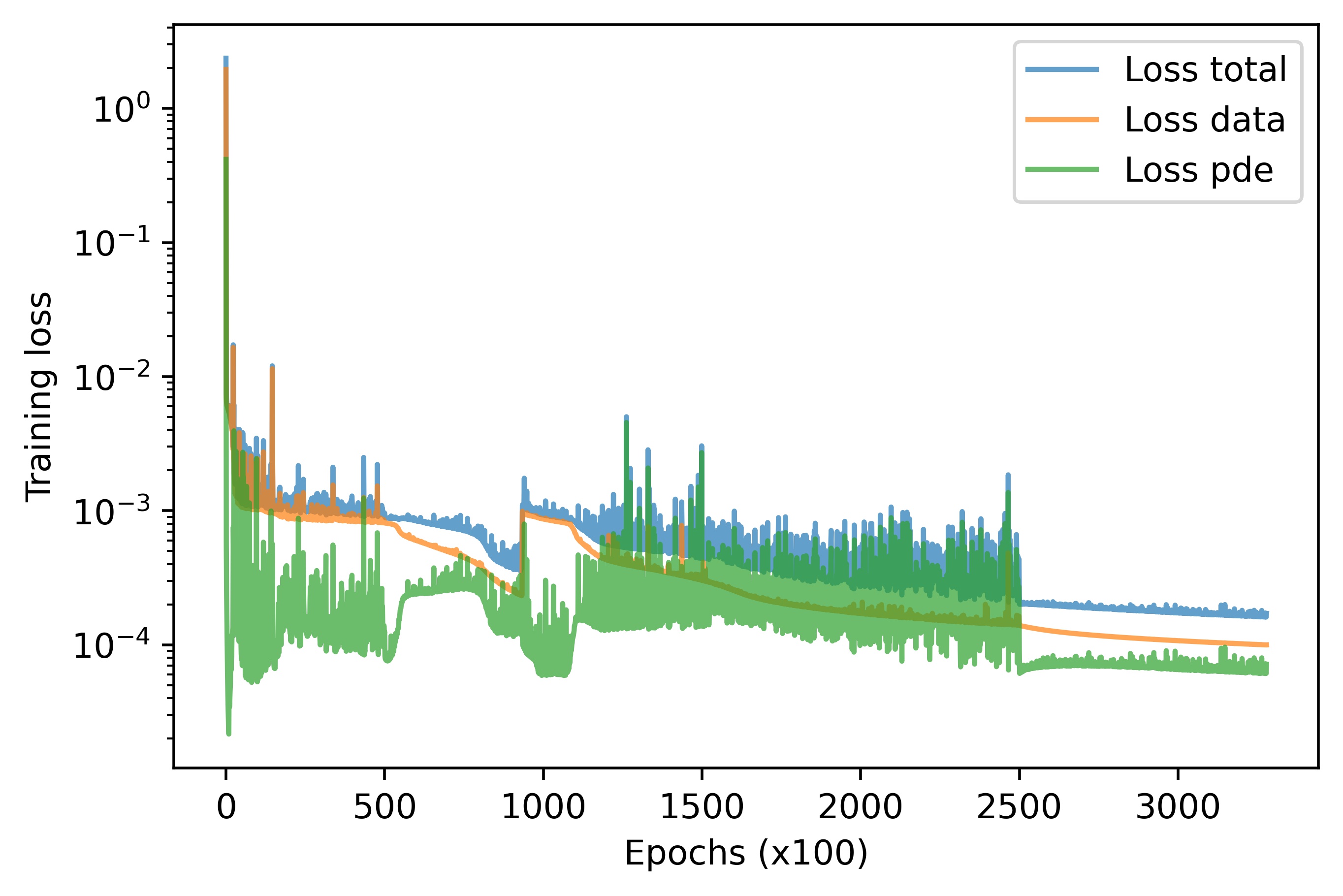} }}%
    \caption{\textit{Cost function during training process of PINNs.}}%
    \label{loss_detail}%
\end{figure}

We aim at studying the impact of the location of the sensors and the number of supervised points on the accuracy of PINNs prediction. For this purpose, we use PINNs with the spatial coordinates $x,y$ as inputs and $T, p, u_x, u_y$ as outputs and minimize the following cost function:
\begin{align*}
    L =& L_{data} + L_{pde}\\
    =&\dfrac{1}{N_T}\sum_{i=1}^{N_T}\omega_{T}(\hat{T}^i - T^{i*})^2+
    \dfrac{1}{N_f}\sum_{i=1}^{N_f}(\omega_1e_1^2+\omega_2e_2^2+\omega_3e_3^2+\omega_4e_4^2)
\end{align*}
where $\hat{T}$ are the predictions obtained by the neural network and $T^*$ are the supervised data (measurements captured from the sensors) for the temperature. We note that there is no data for the velocity and the pressure in the cost function except their relation with the temperature given by the system of PDEs. The quantities $e_1, e_2, e_3, e_4$ correspond to the dimensionless PDEs residuals defined in the equations (\ref{mouveconv}), (\ref{massconv}), (\ref{enerconv}). More precisely:
\begin{align*}
    e_1 &= \dfrac{\partial}{\partial x}\Big(2\eta(\bm{\vec{\hat{u}}},\hat{T})\dfrac{\partial \hat{u}_x}{\partial x}\Big) + \dfrac{\partial}{\partial y}\Big(\eta(\bm{\vec{\hat{u}}},\hat{T})(\dfrac{\partial \hat{u}_x}{\partial y} +  \dfrac{\partial \hat{u}_y}{\partial x})\Big) - \dfrac{\partial \hat{p}}{\partial x}\\
    e_2 &= \dfrac{\partial}{\partial x}\Big(\eta(\bm{\vec{\hat{u}}},\hat{T})(\dfrac{\partial \hat{u}_x}{\partial y} +  \dfrac{\partial \hat{u}_y}{\partial x})\Big) + \dfrac{\partial}{\partial y}\Big(2\eta(\bm{\vec{\hat{u}}},\hat{T})\dfrac{\partial \hat{u}_y}{\partial y}\Big) - \dfrac{\partial \hat{p}}{\partial y} \\
    e_3 &= \dfrac{\partial \hat{u}_x}{\partial x} + \dfrac{\partial \hat{u}_y}{\partial y}\\
    e_4 &= \hat{u}_x\dfrac{\partial \hat{T}}{\partial x} + \hat{u}_y\dfrac{\partial \hat{T}}{\partial y} - \dfrac{1}{Pe}(\dfrac{\partial^2 \hat{T}}{\partial x^2}+\dfrac{\partial^2 \hat{T}}{\partial y^2}) - \dfrac{Br}{Pe}\eta(\bm{\vec{\hat{u}}}, \hat{T})|\gamma(\bm{\vec{\hat{u}}})|^2
\end{align*}
The weight coefficients $\omega_i$ for $i\in\{T, 1,2,3,4\}$ are pre-specified in our case (see below for their specific values). We suppose that there are sensors in the domain so that we can measure the temperature of the flow at these points. The placement of these sensors may be important as it may affect the inferred results. We study three cases of the sensors' placements as shown in Figure \ref{calender_geo_sensor}. In the first case (Figure \ref{calender_geo_sensor0}), the sensors are put randomly inside the calender. This is often the test case used in academic applications, but not always realistic in an industrial context. In the second case (Figure \ref{calender_geo_sensor1}), the sensors are put on two vertical lines located at the input and the output of the calender. In the third case (Figure \ref{calender_geo_sensor6}), the sensors are put randomly in the input zone of the calender. The last two cases are tested as they are often the case in industrial tasks.
To study the impact of the sensor's location on PINNs performance, we fix the number of measurements (supervised points) for the temperature $N_T=500$. For the training of PINNs, the number of collocation points (unsupervised points) is $N_f=10,000$ and these points are generated randomly on the finite element mesh, which provides an \textit{a priori} knowledge on high gradient location (see Figure \ref{calender_geo_colloc1}). In all cases, we use a feedforward neural network with 5 hidden layers with 100 neurons per layer. This architecture is inspired by the work of \citep{raissi2019physics}. Our preliminary results suggest that this choice gives a good balance between the network representation capacity and the computational costs. First, we use PINNs with \textit{tanh} activation function without employing locally adaptive activate functions or deep Kronecker networks. To minimize the cost function, we adopt Adam optimizer with learning rate decay strategy, which is proven to be very efficient in training deep learning models \citep{you2019does}. The results are obtained with 50,000 epochs with the learning rate $lr=10^{-3},$ 200,000 epochs with the learning rate $lr=10^{-4}$ and 150,000 epochs with $lr=10^{-5}$. For the weight coefficients, we fix $\omega_1=\omega_2=\omega_3=\omega_4=1$ and $\omega_{T}=\dfrac{N_T}{\sum_{i=1}^{N_T}( T^{i*})^2}$. By putting these weights, we aim at minimizing the relative errors between the PINNs prediction and solution for the supervised points in the cost function, instead of minimizing the absolute errors as in the conventional PINNs \citep{raissi2019physics}. 
\begin{figure}[H]
    \centering
    \subfloat[Reference solution]{{\includegraphics[width=4cm]{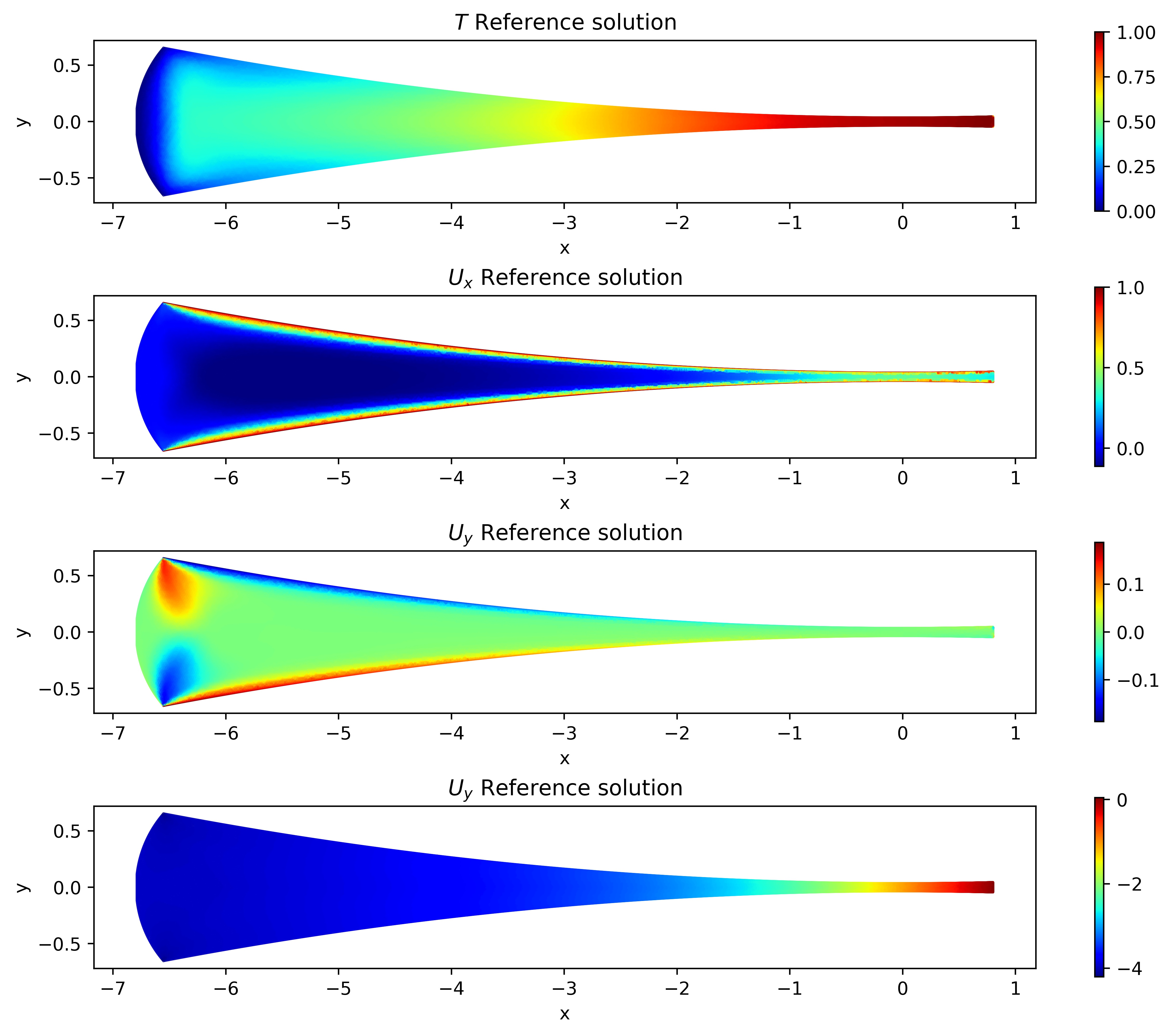}\label{infer_sol_ref} }}%%
    \subfloat[PINNs prediction: case 1]{{\includegraphics[width=4cm]{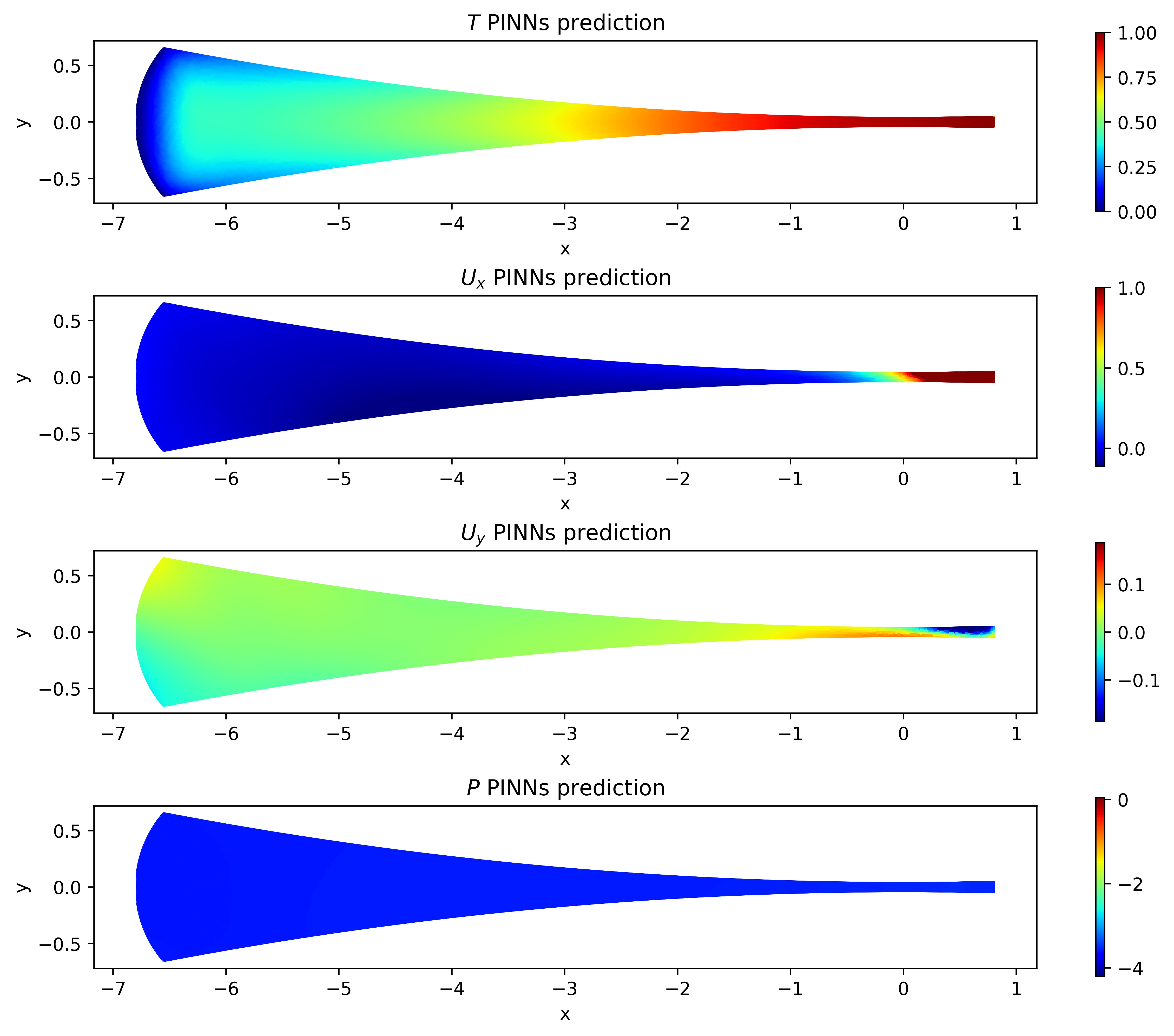} }}%
    \subfloat[PINNs prediction: case 2]{{\includegraphics[width=4cm]{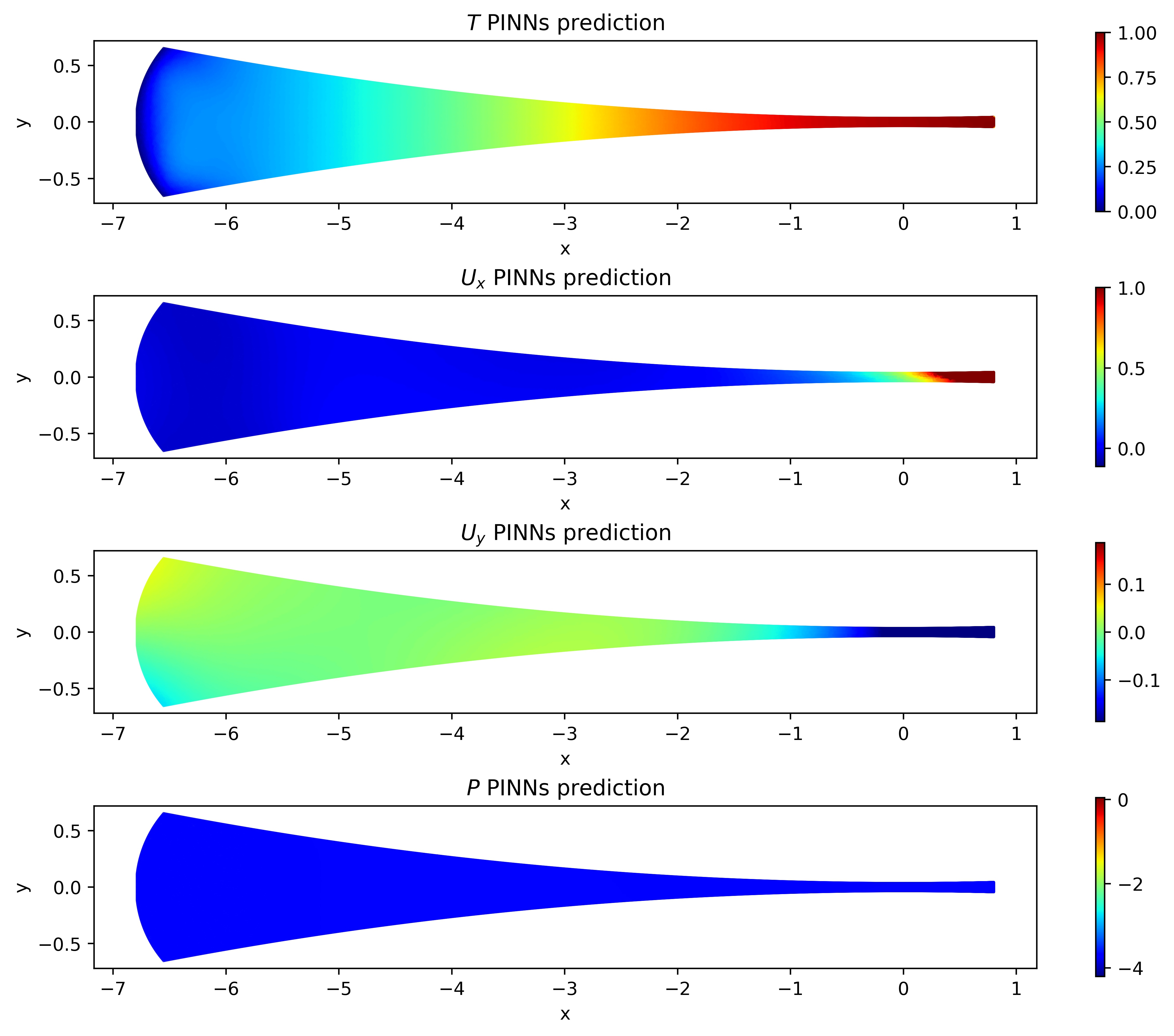} }}%
    \subfloat[PINNs prediction: case 3]{{\includegraphics[width=4cm]{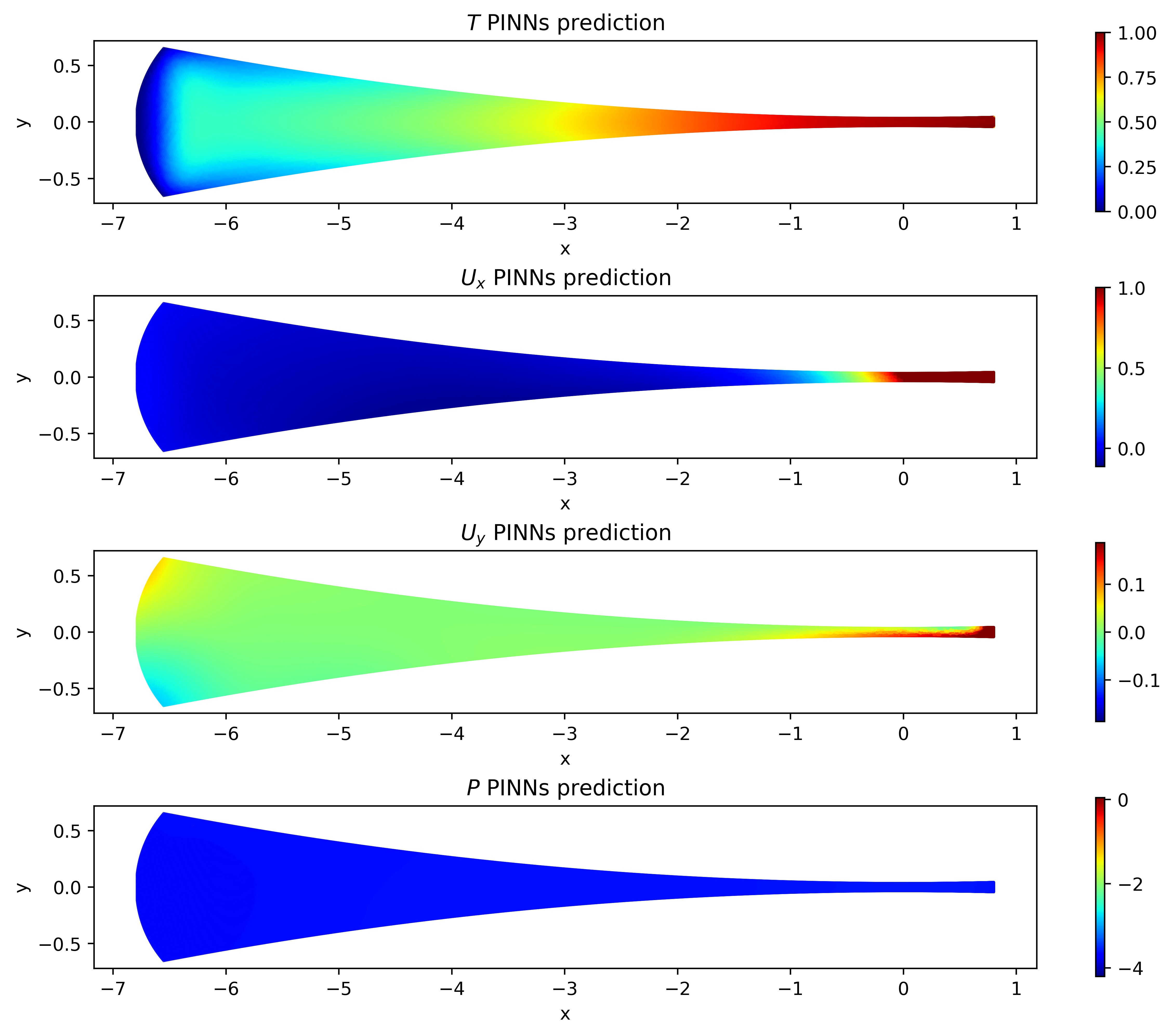} }}%
    \caption{\textit{Reference solution and PINNs predictions in different scenarios of the location of sensors:} from top to bottom: the solution for the temperature, the velocity components and the pressure, respectively.}%
    \label{infer_sol}%
\end{figure}
Figure \ref{infer_sol} illustrates the visual comparison between the reference solution and PINNs prediction in different scenarios of the location of sensors. We observe that in all three cases, PINNs are only able to approximate accurately the temperature in the zones where there are sensors and fail to predict the velocity and pressure. We note that since there is no supervised measurement for the velocity and pressure, these fields can be accurately approximated only after the temperature is.  Figure \ref{loss_detail} shows the detail of each loss term in the cost function. We see that after the training process, the loss for the data (temperature measurements) and the loss for PDEs residuals are both minimized to very small values. This performance demonstrates that even when the temperature is accurately approximated at all points inside the domain (as case 1) and the PDEs constraints are well respected, PINNs still fail to infer the velocity and the pressure, \textit{i.e.} the information of the temperature is not sufficient to guarantee a unique solution for the velocity and the pressure fields. The same conclusion is obtained when using PINNs with locally adaptive activation functions or deep Kronecker networks (which are not shown here for conciseness).

To make sure that the problem has a unique solution, we take in addition the knowledge of the velocity boundary conditions. Moreover, since the supervised points for the temperature are only available at some specific location in case 2 (Figure \ref{calender_geo_sensor1}) and case 3 (Figure \ref{calender_geo_sensor6}), PINNs, as other deep learning methods, have difficulties to extrapolating the values of the temperature at points far from the sensors. Thus we also provide PINNs with the information of the temperature on the roll boundaries only. We note that when the sensors are distributed as in case 1 (Figure \ref{calender_geo_sensor0}), the learning points for the temperature on the roll boundaries are not necessary. The cost function in PINNs now becomes:
\begin{align*}
    L =& L_{data} + L_{pde} + L_{bc}\\
    =&\dfrac{1}{N_T}\sum_{i=1}^{N_T}\omega_{T}(\hat{T}^i - T^{i*})^2+
    \dfrac{1}{N_f}\sum_{i=1}^{N_f}(\omega_1e_1^2+\omega_2e_2^2+\omega_3e_3^2+\omega_4e_4^2) \\+
    &\dfrac{1}{N_{rolls}}\sum_{i=1}^{N_{rolls}}(\omega_{T^{rolls}}(\hat{T}^{rolls,i} - T^{rolls,i*})) + \dfrac{1}{N_{bc}}\sum_{i=1}^{N_{bc}}\Big(\omega_{u_x^{bc}}(\hat{u}_x^{bc,i} - u_x^{bc,i*})^2+\omega_{u_y^{bc}}(\hat{u}_y^{bc,i} - u_y^{bc,i*})^2\Big)
\end{align*}
where $T^{rolls*}$ denotes the values of the temperature on the rolls and $u_x^{bc*}, u_y^{bc*}$ denotes the components of the velocity on the four boundaries of the calender. For the weight coefficients, we fix $\omega_1=\omega_2=\omega_3=\omega_4=1$, $\omega_{T}=\dfrac{N_T}{\sum_{i=1}^{N_T}( T^{i*})^2}$, $\omega_{T^{rolls}}=\dfrac{N_{rolls}}{\sum_{i=1}^{N_{rolls}}( T^{rolls,i*})^2}$, $\omega_{u_x^{bc}}=\dfrac{100N_{bc}}{\sum_{i=1}^{N_{bc}}( u_x^{bc,i*})^2}$ and $\omega_{u_y^{bc}}=\dfrac{10N_{bc}}{\sum_{i=1}^{N_{bc}}( u_y^{bc,i*})^2}$. Here we still minimize the relative errors between the prediction and the solution at the learning points, but we put more important weights on the boundary condition for the velocity. 
\begin{figure}[H]
    \centering
    \subfloat[Case 1]{{\includegraphics[width=5cm]{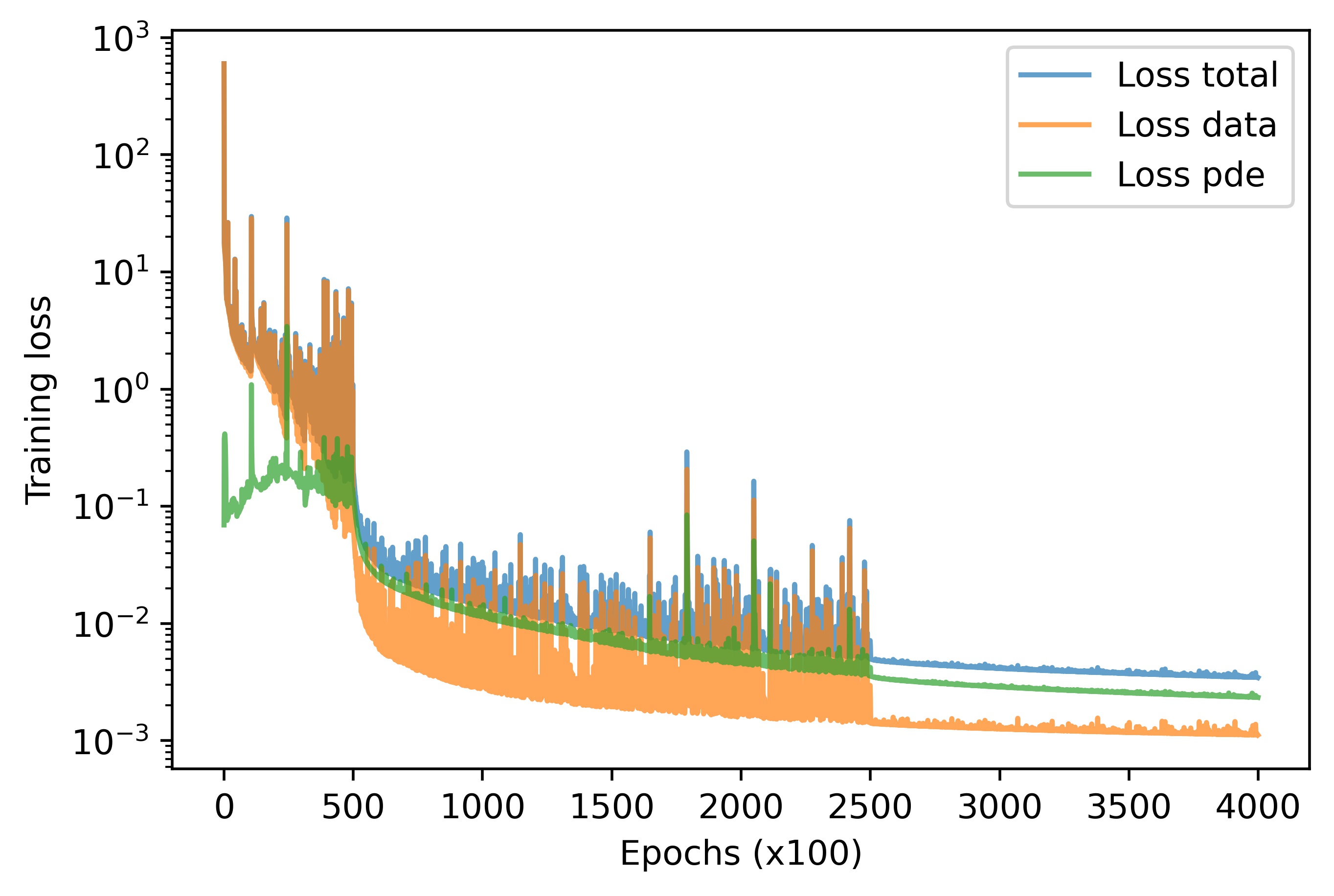} }}%%
    \subfloat[Case 2]{{\includegraphics[width=5cm]{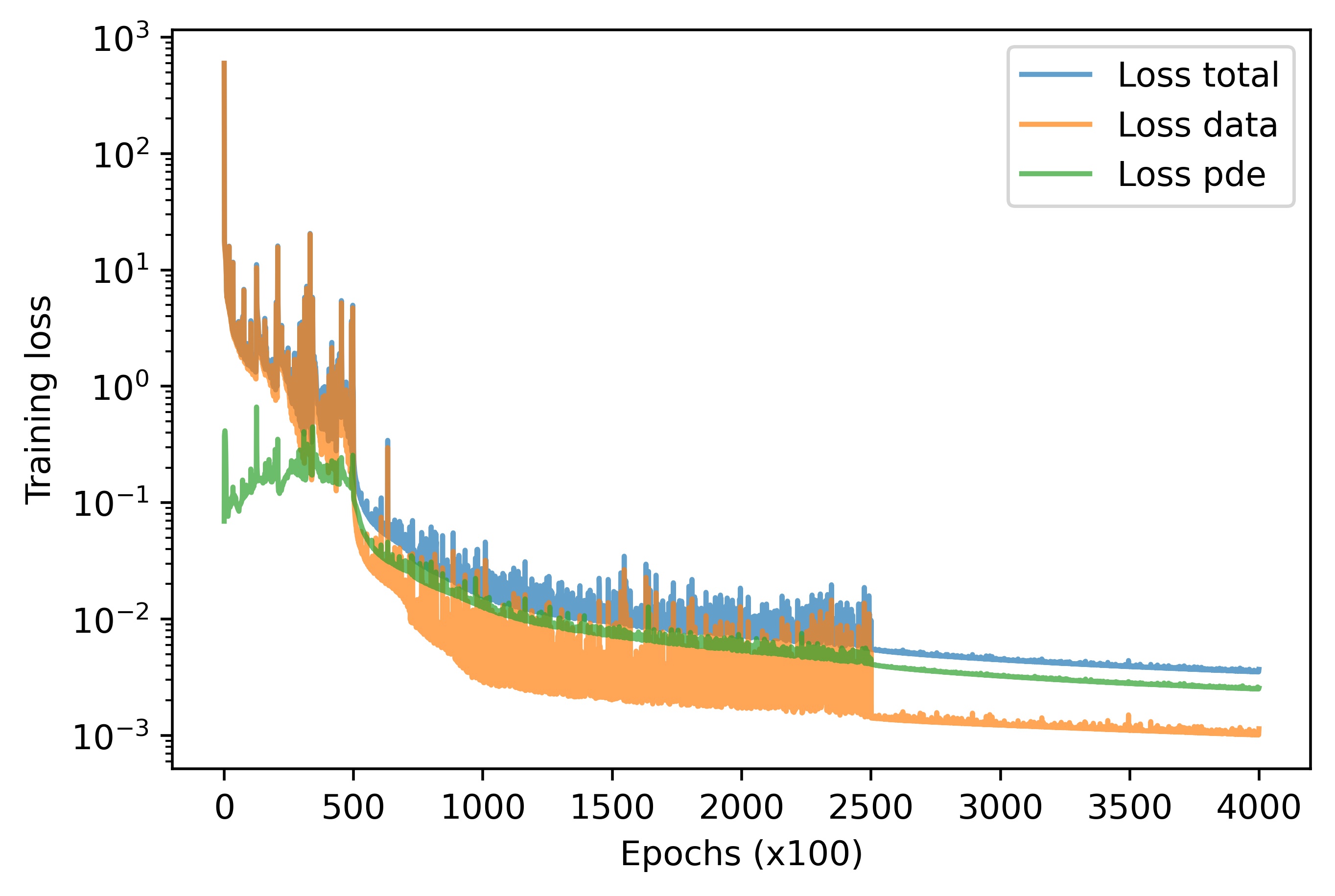} }}%
    \subfloat[Case 3]{{\includegraphics[width=5cm]{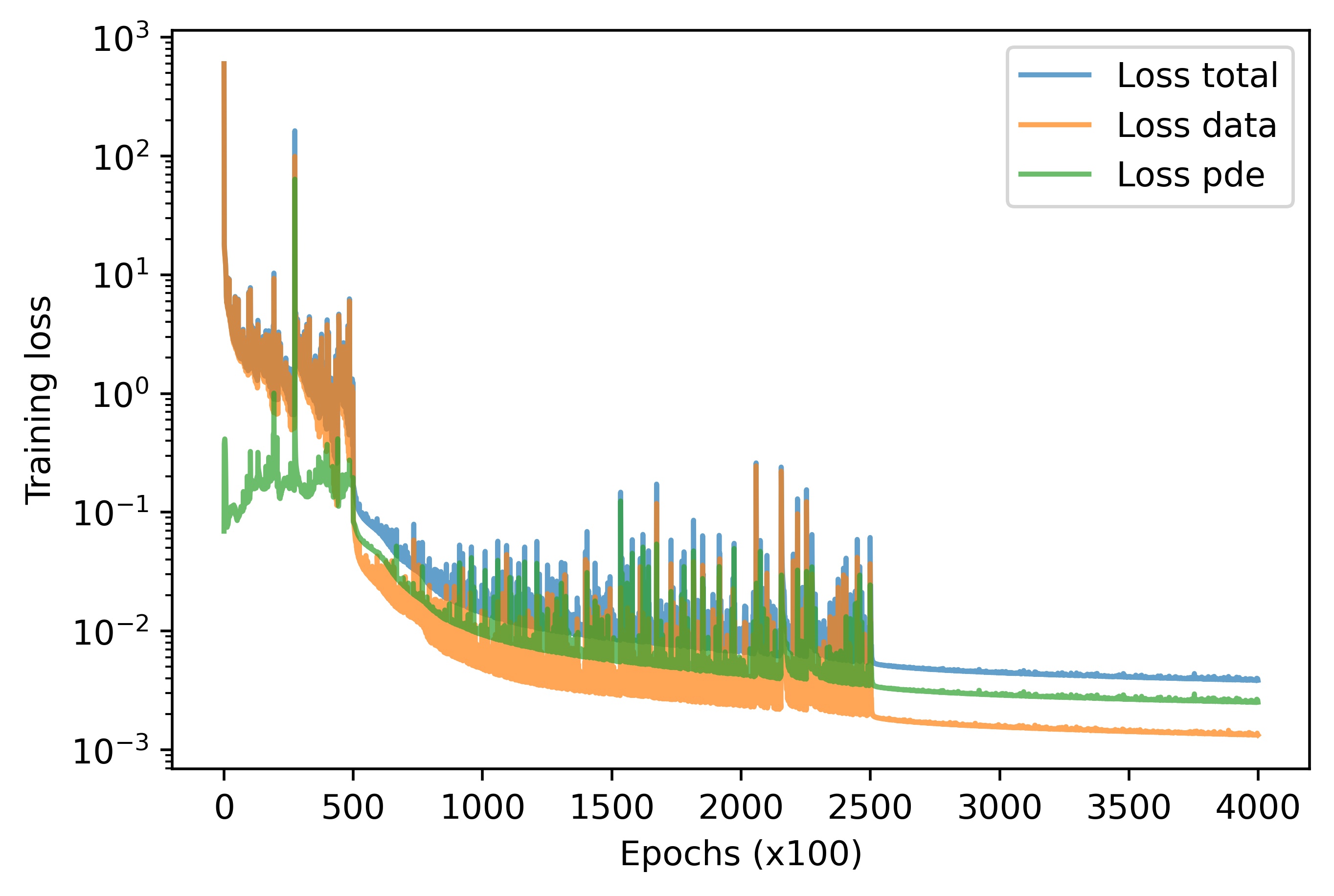} }}%
    \caption{\textit{Cost function during training process of PINNs after adding the knowledge for the velocity boundary conditions.}}%
    \label{loss_compare_bc}%
\end{figure}

\begin{figure}[H]
    \centering
    \subfloat[Reference solution]{{\includegraphics[width=4cm]{paper_hf.jpg} }}%%
    \subfloat[PINNs prediction: case 1]{{\includegraphics[width=4cm]{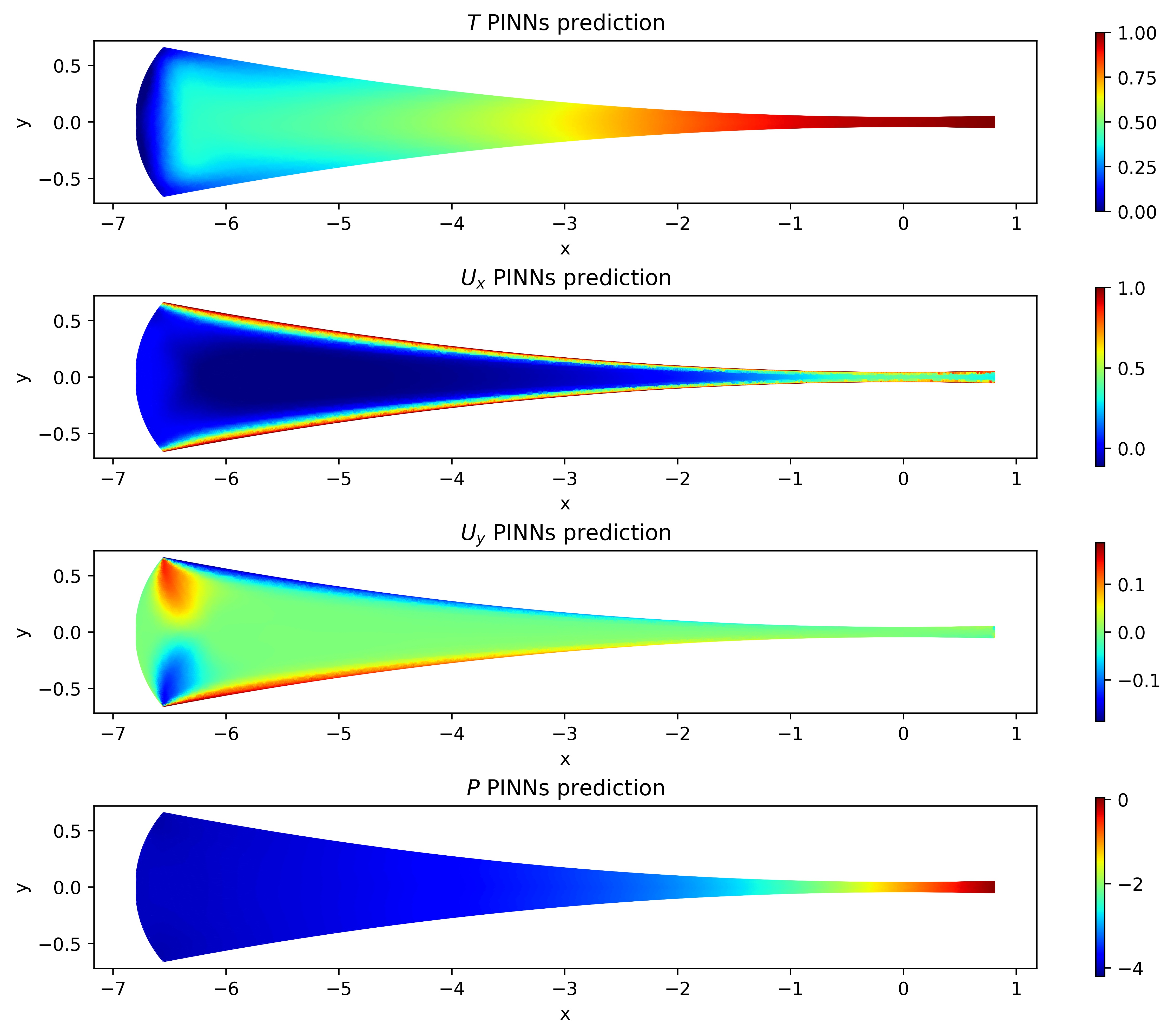} }}%
    \subfloat[PINNs prediction: case 2]{{\includegraphics[width=4cm]{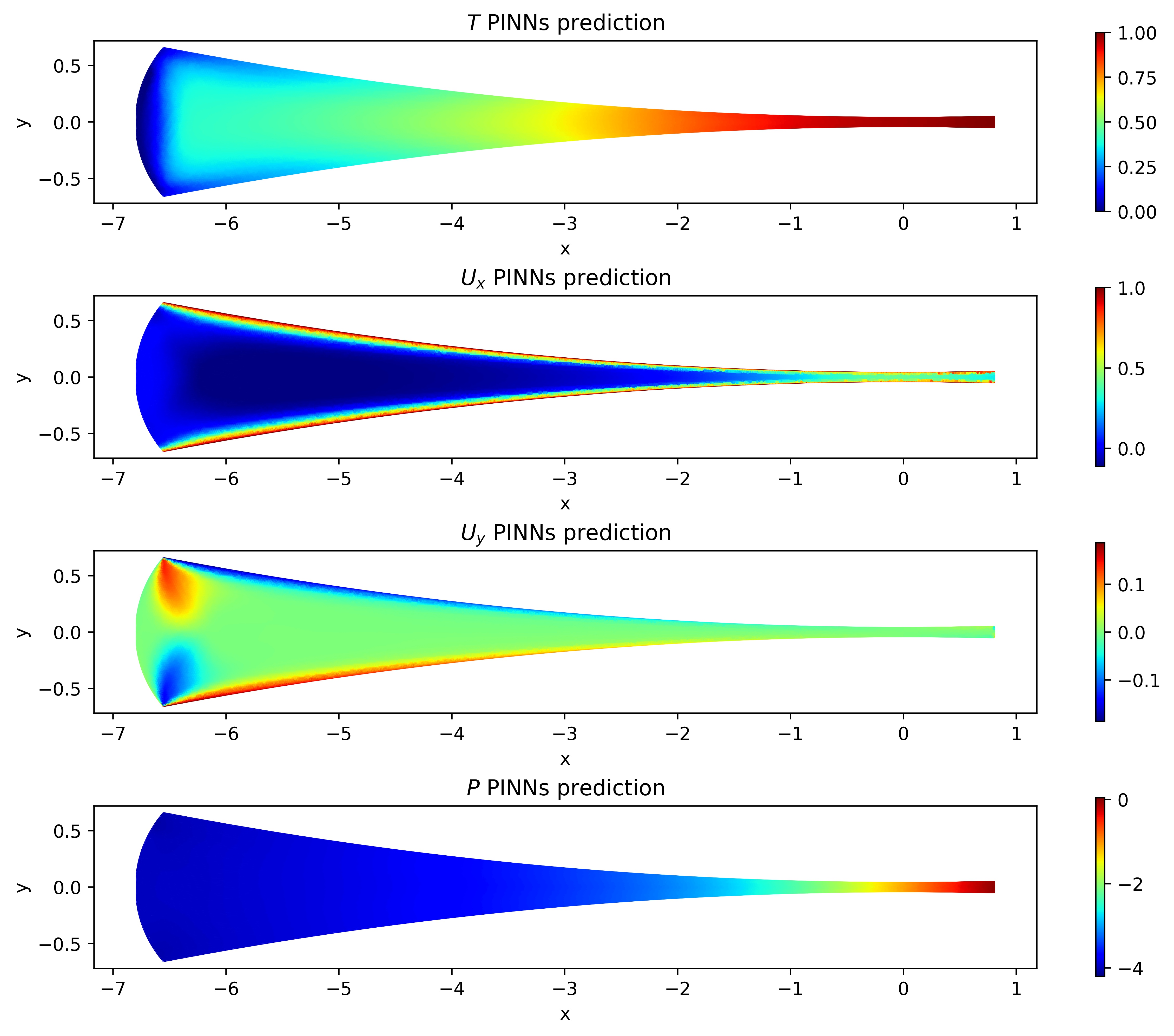} }}%
    \subfloat[PINNs prediction: case 3]{{\includegraphics[width=4cm]{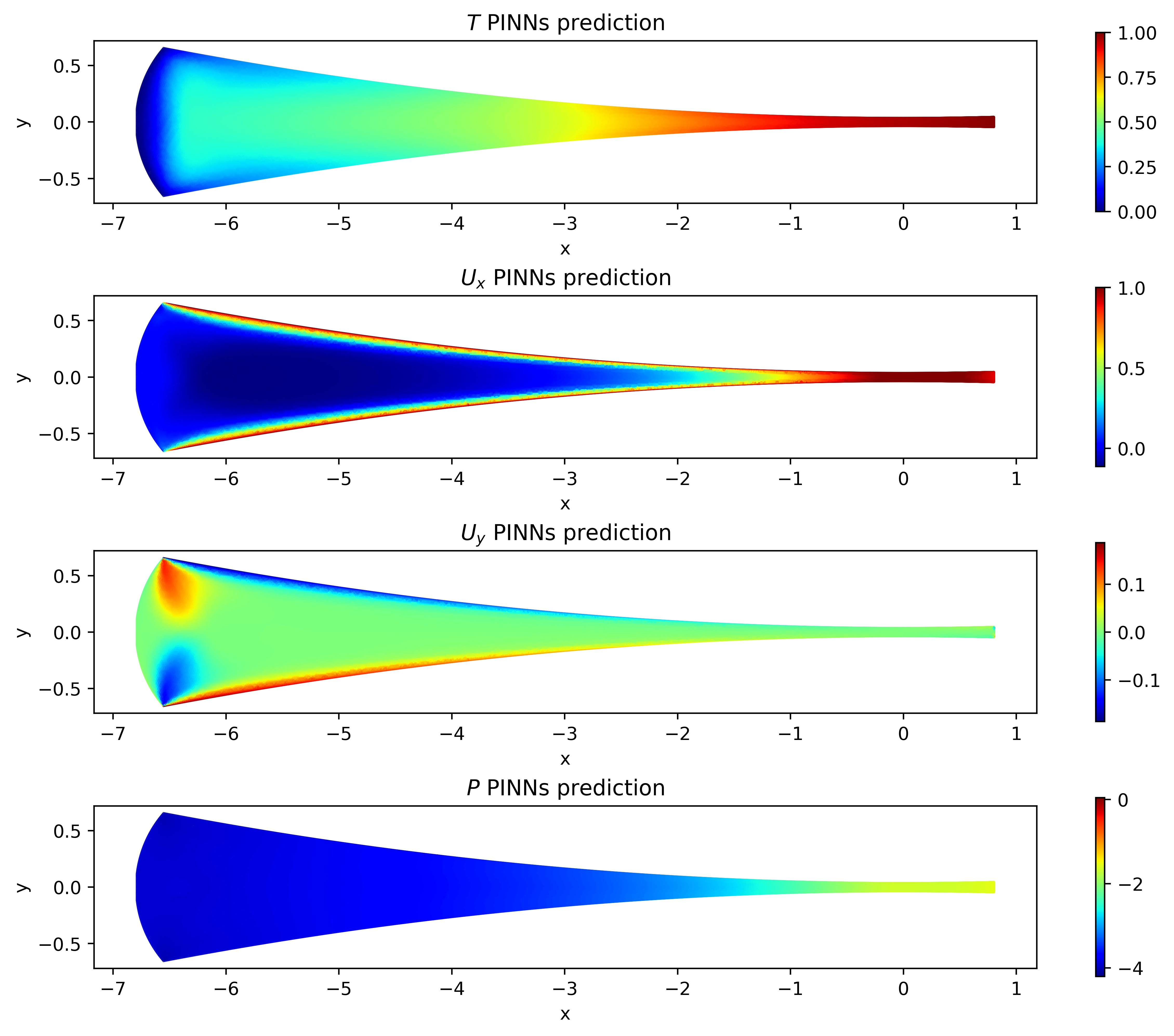} }}%
    \caption{\textit{Reference solution and PINNs prediction when using in additional the boundary condition for the velocity and the temperature as learning data.}}%
    \label{infer_sol_bc}%
\end{figure}

Figure \ref{loss_compare_bc} shows the detail for each loss term in the cost function and Figure \ref{infer_sol_bc} presents the visual performance of PINNs for all cases. We see that PINNs are capable to predict accurately the physical fields even in the case where there are only sensors at the input and output lines of the calender (case 2). However, when there are only sensors at the input zone (case 3), PINNs are able to fit the boundary condition for the temperature and the velocity, but still fail to approximate accurately the solution at the output of the calender.

\begin{figure}[H]
    \centering
    \subfloat[Case 1]{{\includegraphics[width=5cm]{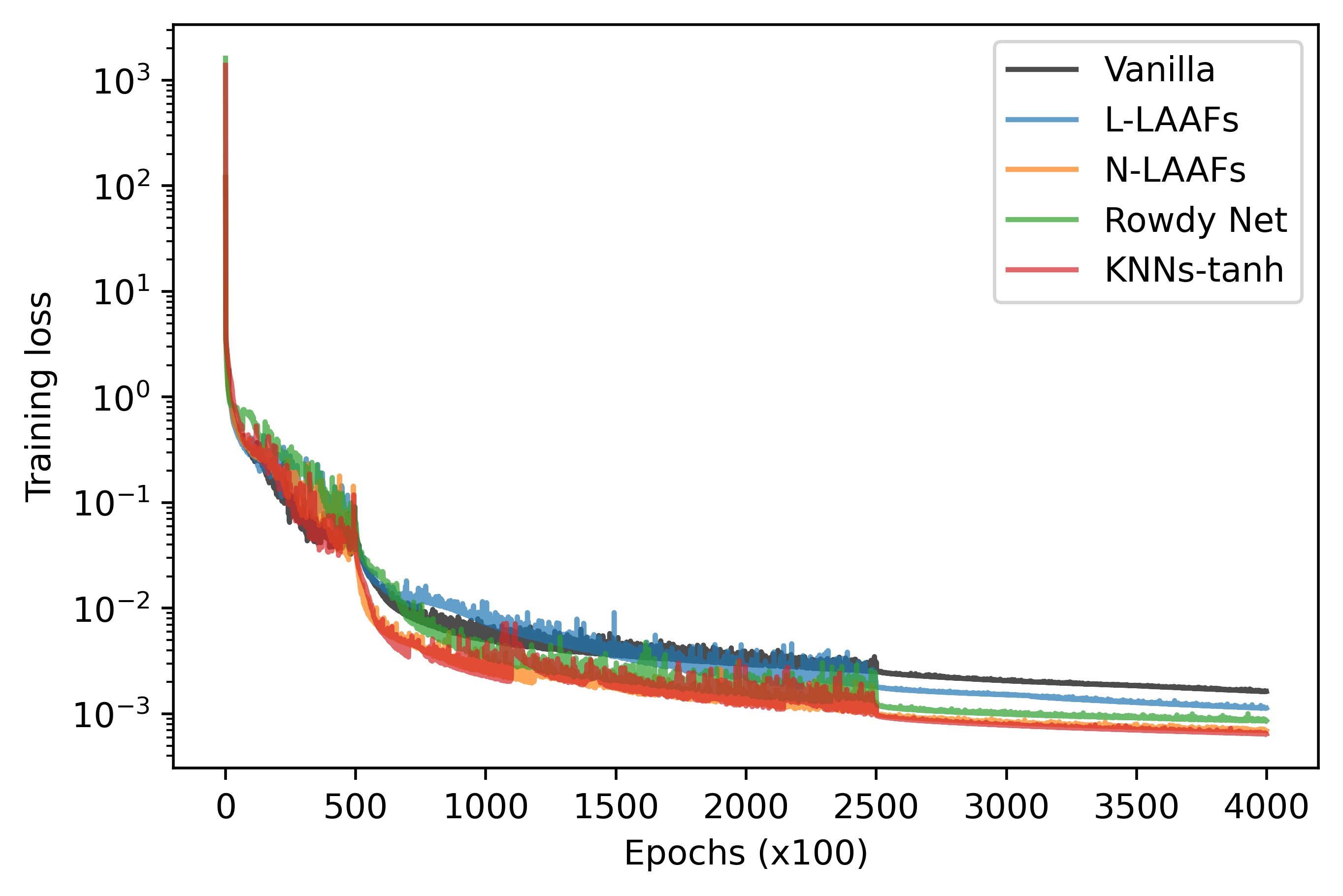} }}%%
    \subfloat[Case 2]{{\includegraphics[width=5cm]{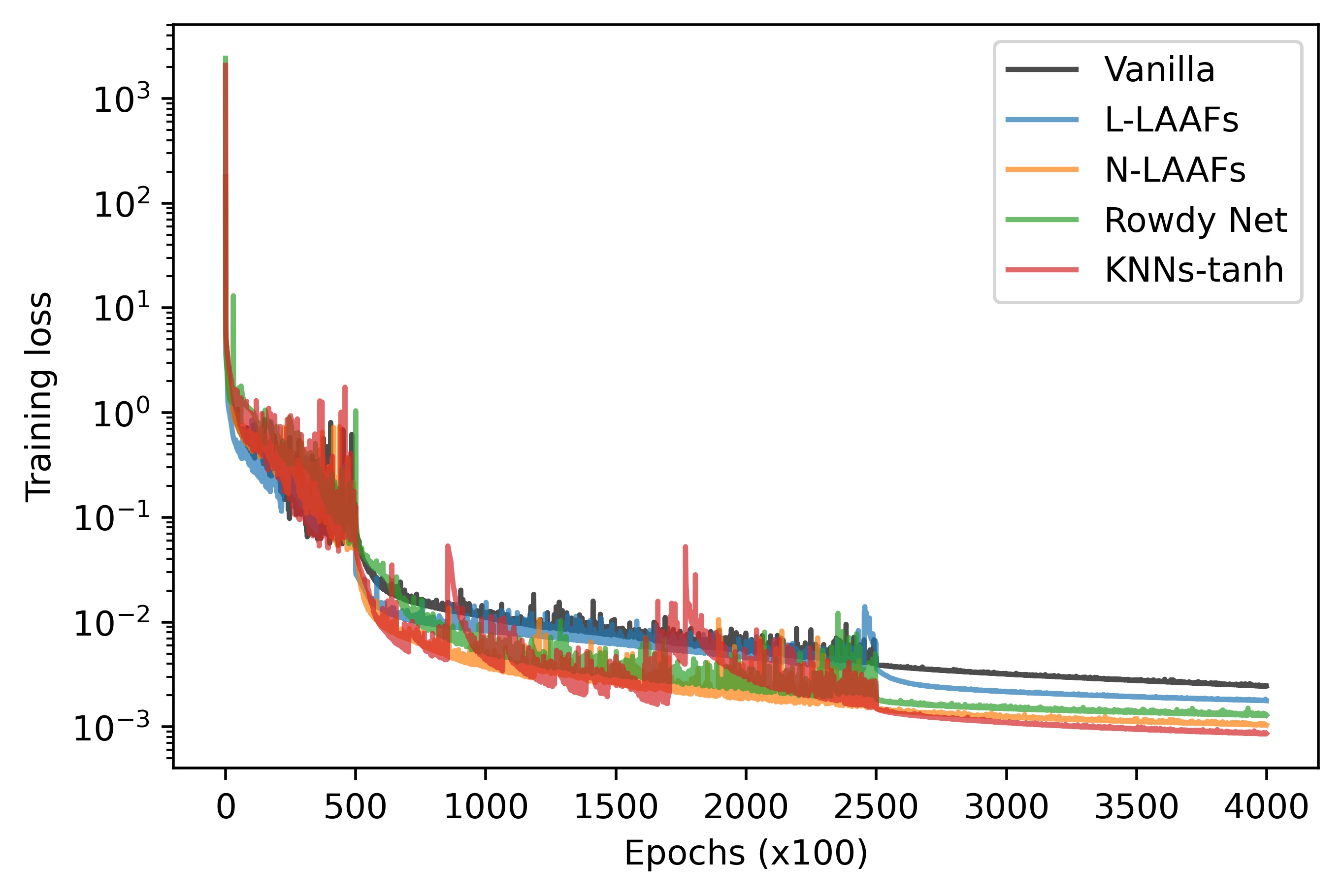} }}%
    \subfloat[Case 3]{{\includegraphics[width=5cm]{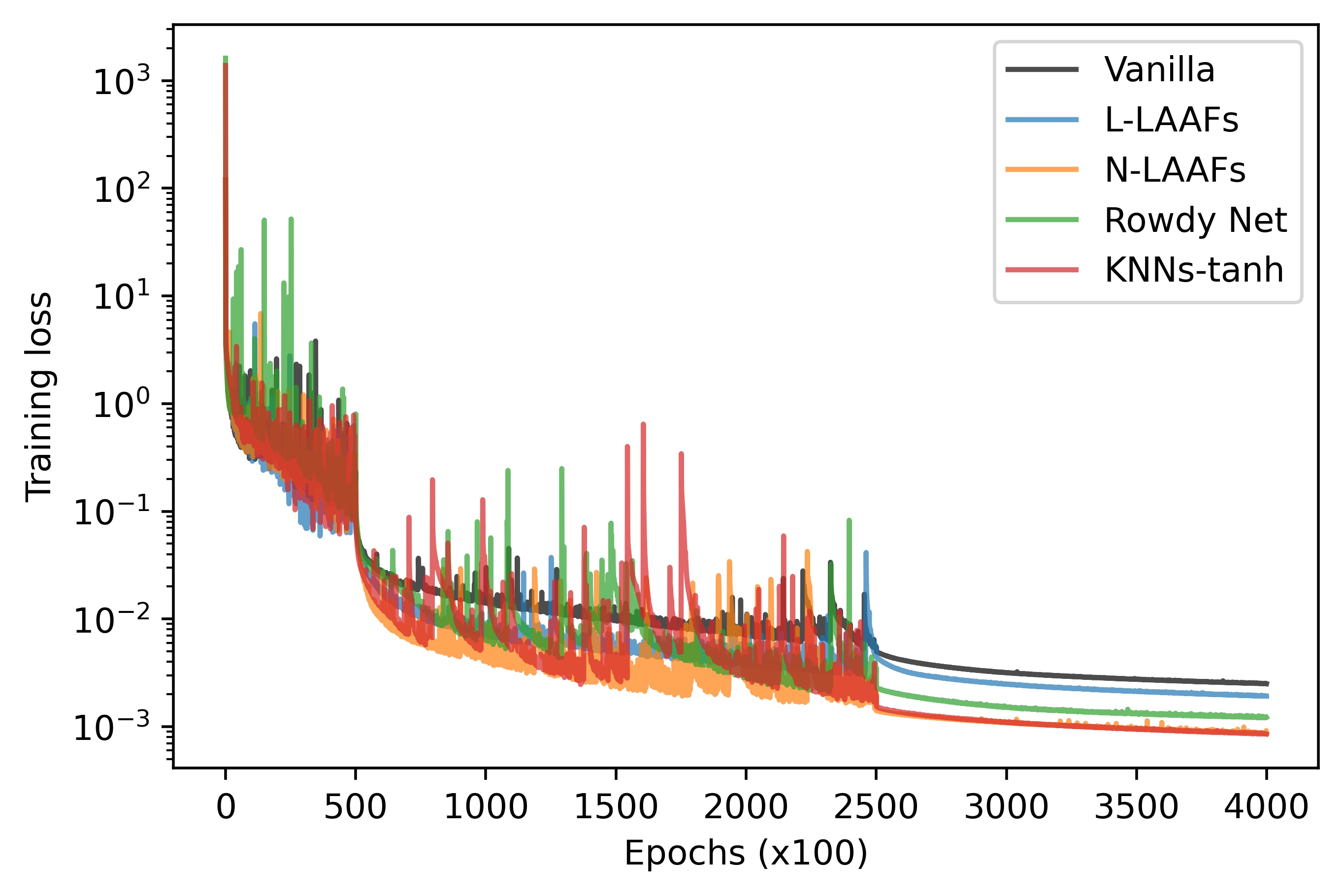} }}%
    \caption{\textit{Cost function during training process of PINNs with different approaches.}}%
    \label{loss_compare}%
\end{figure}

Next, we compare vanilla PINNs with PINNs using locally adaptive activation function (L-LAAFs and N-LAAFs) and PINNs with deep Kronecker networks (Rowdy Net and KNNs-tanh) to verify whether these methods help to improve the performance of PINNs. For L-LAAFs and N-LAAFs, we fix the scaling factor $n=1$ and initialize the adaptive parameters $a^k=1$ or $a^k_i=1$. For Rowdy Net and KNNs-tanh, we fix $K=2$ and $\alpha_l^k=\omega_l^k=1$ and the scaling factor $n=1$ (for Rowdy Net). Figure \ref{loss_compare} shows the cost function during the training process of these methods in different scenarios of sensor location. Table \ref{infer_lnlaaf} illustrates the performance of each model in case 2 of sensor location in terms of relative $\mathcal{L}^2$ errors with the reference solution. We see that when using locally adaptive activation functions or deep Kronecker networks, PINNs provide better results compared to vanilla PINNs. This performance is as expected since these models offer higher degrees of freedom in the optimization problem than vanilla PINNs. Thus the neural networks have additional flexibility in training, which leads to better performance than vanilla PINNs. When comparing LAAFs and deep Kronecker networks, we see that the model with KNNs-tanh gives the best accuracy for all physical fields, and then come N-LAAFs, Rowdy Net and L-LAAFs. However, during the training process, we observe that the models with deep Kronecker networks, especially KNNs-tanh, are very sensitive during the training process with high learning rate. These models only become more stable when the learning rate decreases. We note that we can further increase the accuracy of Rowdy Net and KNNs-tanh by increasing the number of terms $K$, which however leads to a high computational cost. In this work, we only consider $K=2$, which already yields satisfying performance. Besides that, in Table \ref{infer_lnlaaf}, we see that the error for $T$ is larger than the other fields. We can reduce this error by increasing the number of training epochs (see Table \ref{infer_err_noise} for more details).

\begin{table}[H]
\centering
\begin{tabular}{ |c|c|c|c|c| } 
\hline
 & $\epsilon_T$ & $\epsilon_{u_x}$ & $\epsilon_{u_y}$ & $\epsilon_P$\\
\hline
Vanilla & 11.1 $\pm$ 0.14 & 7.57 $\pm$ 0.12 & 5.73 $\pm$ 0.04 & 1.79 $\pm$ 0.00\\
L-LAAFs & 10.8 $\pm$ 0.16 & 7.50 $\pm$ 0.13 & 5.69 $\pm$ 0.05 & 1.61 $\pm$ 0.00\\
N-LAAFs & 10.2 $\pm$ 0.14 & 7.00 $\pm$ 0.09 & 5.19 $\pm$ 0.02 & 1.10 $\pm$ 0.00 \\
Rowdy Net & 10.3 $\pm$ 0.13 & 7.24 $\pm$ 0.09 & 5.33 $\pm$ 0.03 & 1.32 $\pm$ 0.00 \\
\textbf{KNNs-tanh} & \textbf{10.1 $\pm$ 0.16} & \textbf{ 6.89 $\pm$ 0.10} & \textbf{5.05 $\pm$ 0.04} & \textbf{0.96 $\pm$ 0.00} \\
\hline
\end{tabular}
\caption{\textit{Relative $\mathcal{L}^2$ errors between reference solution and PINNs prediction with different methods in case 2.}}
\label{infer_lnlaaf}
\end{table}

Next, we study two different cases for the distribution of collocation points: in the first case, these points are generated randomly inside the domain while in the second case, they are generated randomly on the finite element mesh. An example of the visualization of these collocation points can be seen in Figure \ref{calender_geo_colloc}. Table \ref{infer_err_3case} summarizes the performance of PINNs using KNNs-tanh approach in terms of relative $\mathcal{L}^2$ errors compared with the reference solution. These errors are evaluated on the finite element mesh. It can be observed that, using the collocation points randomly taken from the finite element mesh allows PINNs to give a much better precision than using the points randomly taken inside the domain. We also notice that, since the errors are evaluated on the finite element mesh which is much finer at the output, the error of the prediction at the output gives a more important weight than the one at the input in the final accuracy. This remark can be seen clearly in case 3 of the placement of the sensors where we obtain a very high relative error for $u_x$ and $p$, even though, PINNs only fail to predict accurately the solution at the output of the calender (which is reasonable as we only have supervised points at the input).
\begin{table}[H]
\centering
\begin{tabular}{ |c|c|c|c|c|c|c|c|c| } 
\cline{2-9}
\multicolumn{1}{c|}{\multirow{2}{*}{}} & \multicolumn{2}{c|}{$\epsilon_T$} & \multicolumn{2}{c|}{$\epsilon_{u_x}$} & \multicolumn{2}{c|}{$\epsilon_{u_y}$} & \multicolumn{2}{c|}{$\epsilon_p$} \\
\cline{2-9}
\multicolumn{1}{c|}{} & (1) & (2) & (1) & (2) & (1) & (2) & (1) & (2) \\
\hline
Case 1 & 14.3 & \textbf{7.86} & 17.8 & \textbf{2.56} & 5.59 & \textbf{5.41} & 18.9 & \textbf{0.29}\\
Case 2 & 13.9 & \textbf{10.1} & 49.9 & \textbf{6.90} & 10.3 & \textbf{5.04} & 0.95 & \textbf{0.39}\\
Case 3 & 14.8 & \textbf{5.83} & 125 & 107 & 5.76 & \textbf{5.44} & 55.1 &47.5 \\
\hline
\end{tabular}
\caption{\textit{Relative $\mathcal{L}^2$ errors between the reference solution and the predictions by PINNs in different scenarios of sensors' placement.} The errors are evaluated on the finite element mesh (FE). We compare two cases of the collocation points' distribution: (1) these points are taken randomly on the random mesh. (2) these points are taken randomly on the finite element mesh. The results highlighted in bold are the ones that provide satisfying predictions visually and numerically.}
\label{infer_err_3case}
\end{table}

\subsubsection{Impact of noisy measurements:} 
\begin{figure}[H]
    \centering
    \subfloat[$T$]{{\includegraphics[width=3.9cm]{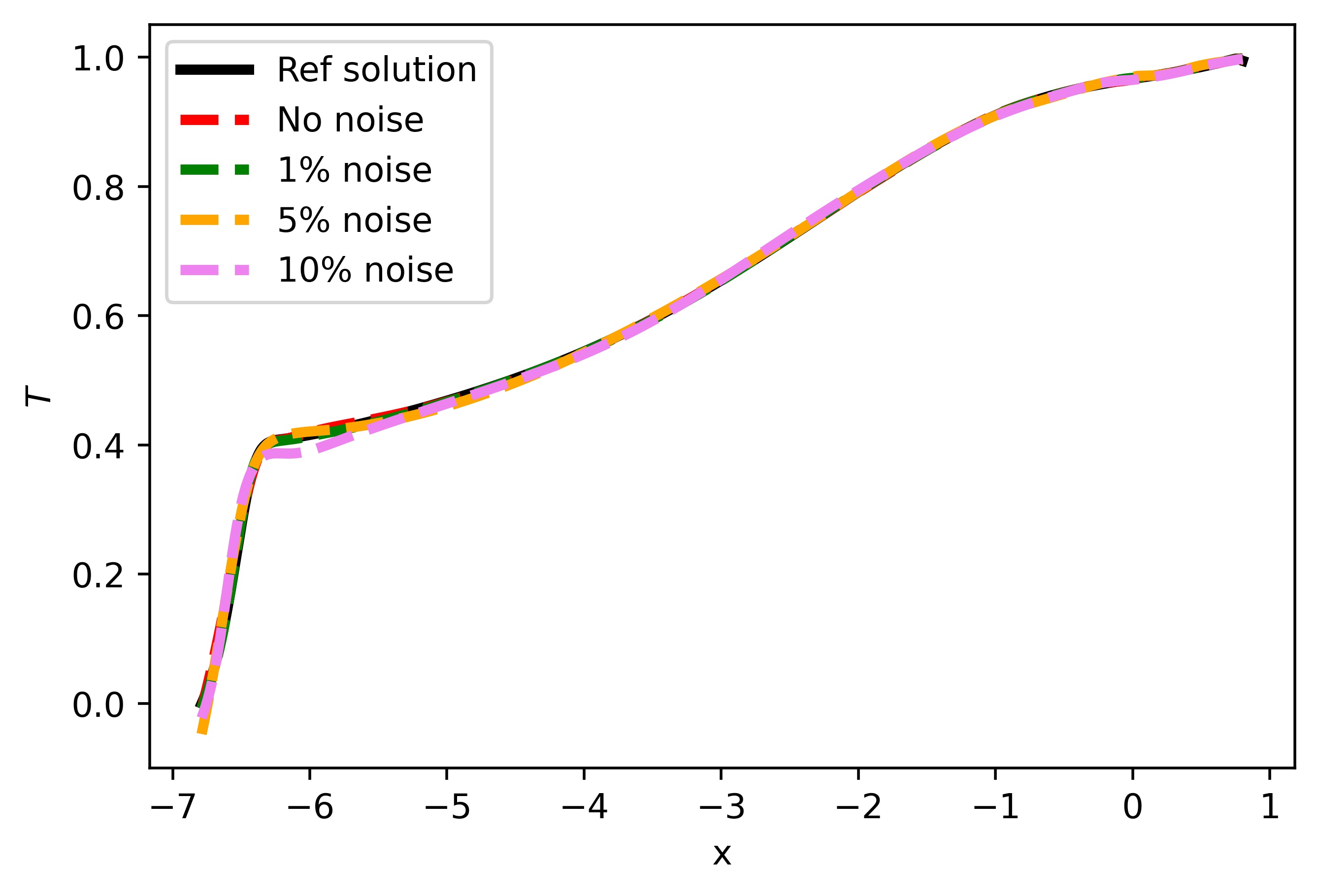} }}%%
    \subfloat[$u_x$]{{\includegraphics[width=4cm]{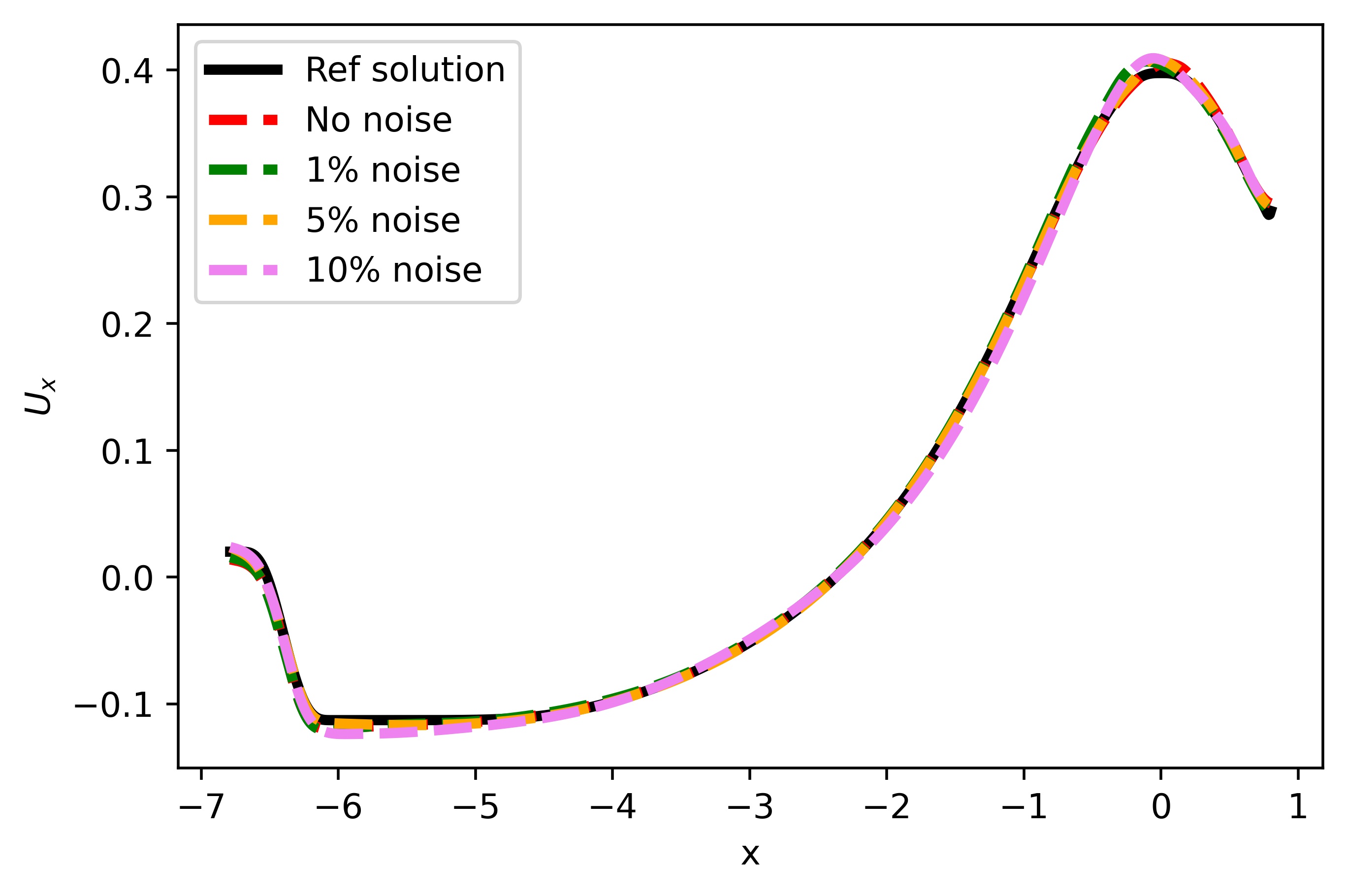} }}%
    \subfloat[$u_y$]{{\includegraphics[width=4.1cm]{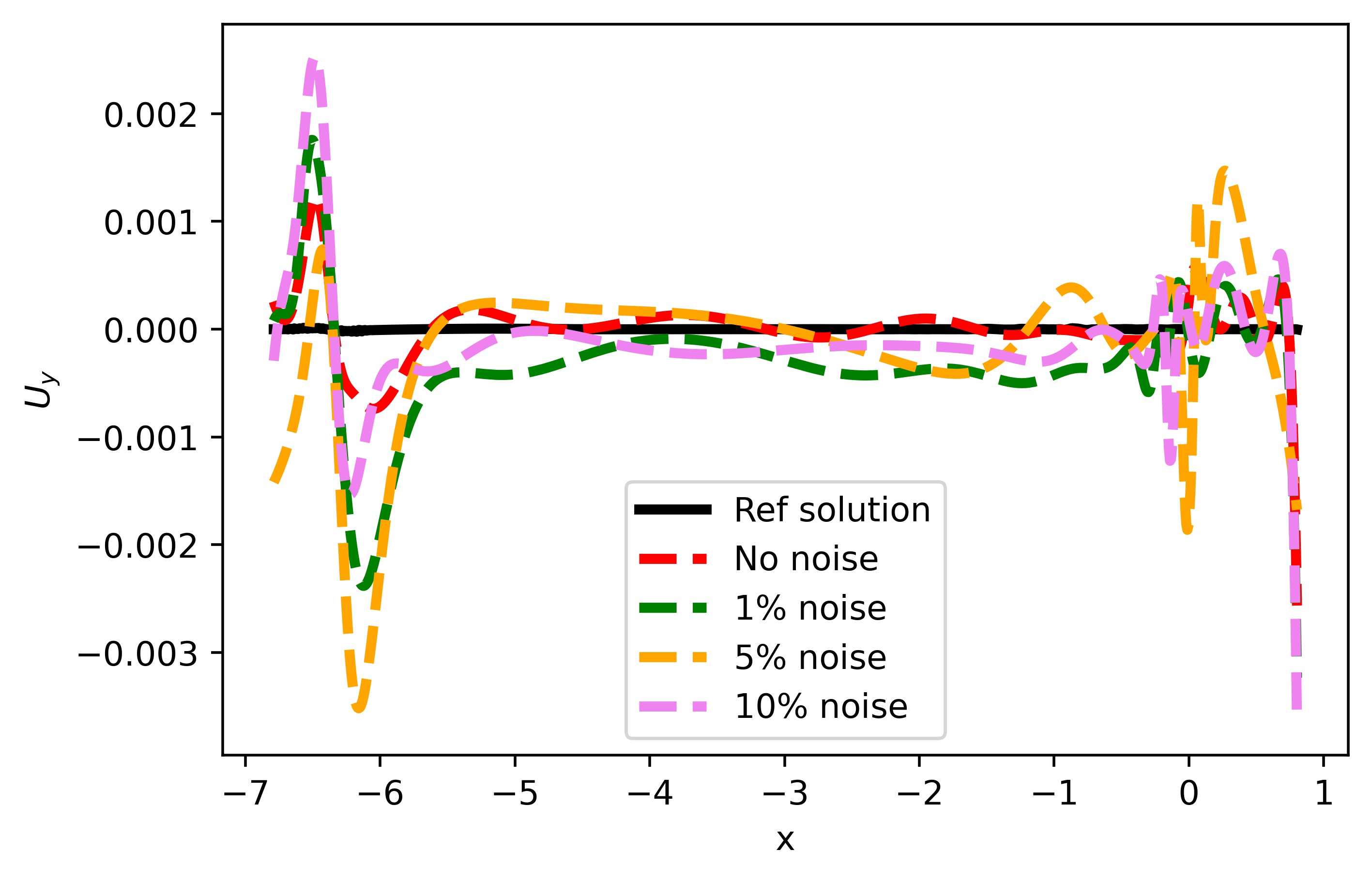} }}%
    \subfloat[$p$]{{\includegraphics[width=4cm]{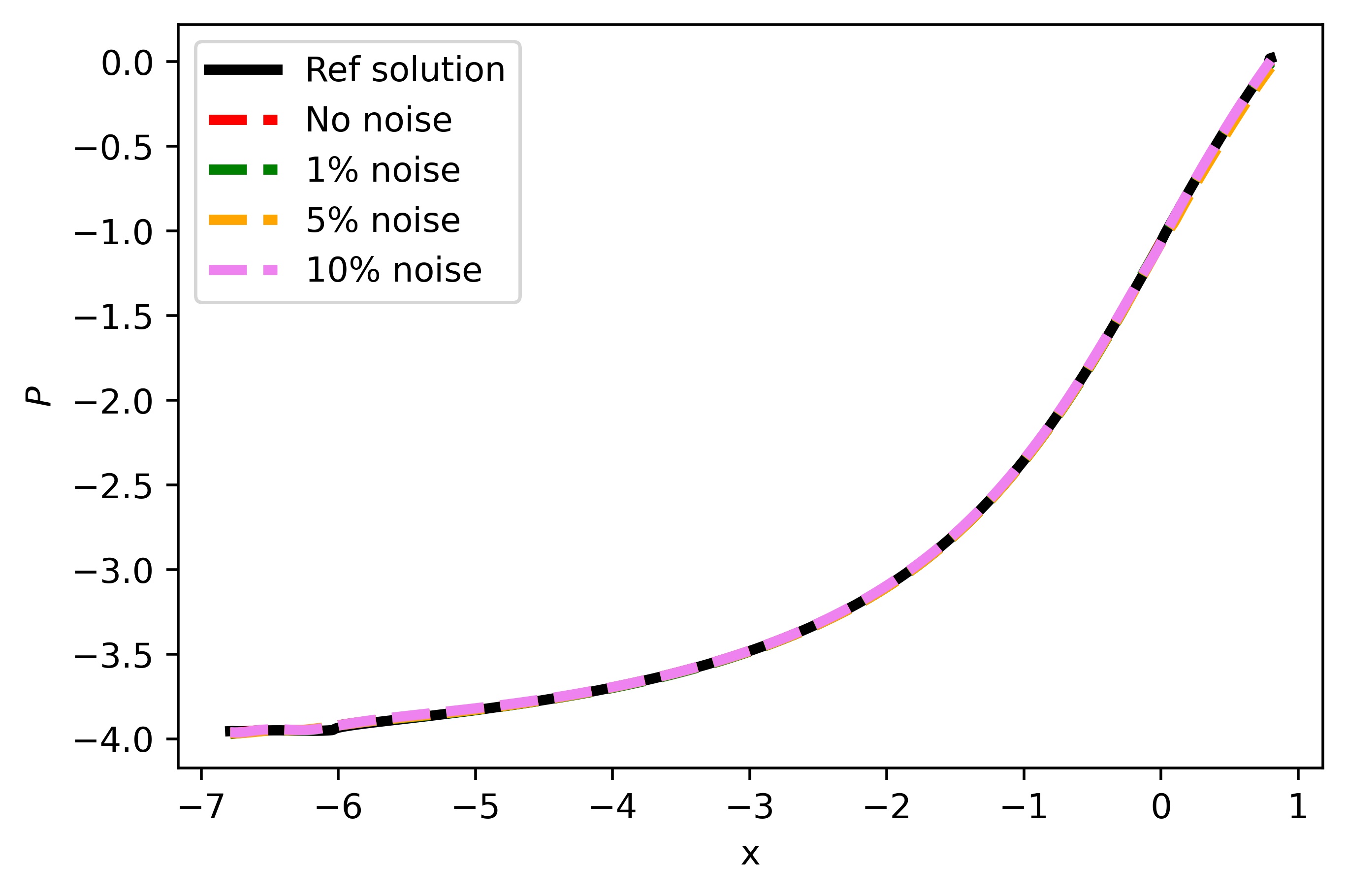} }}%
    \quad
    \subfloat[$T$]{{\includegraphics[width=4cm]{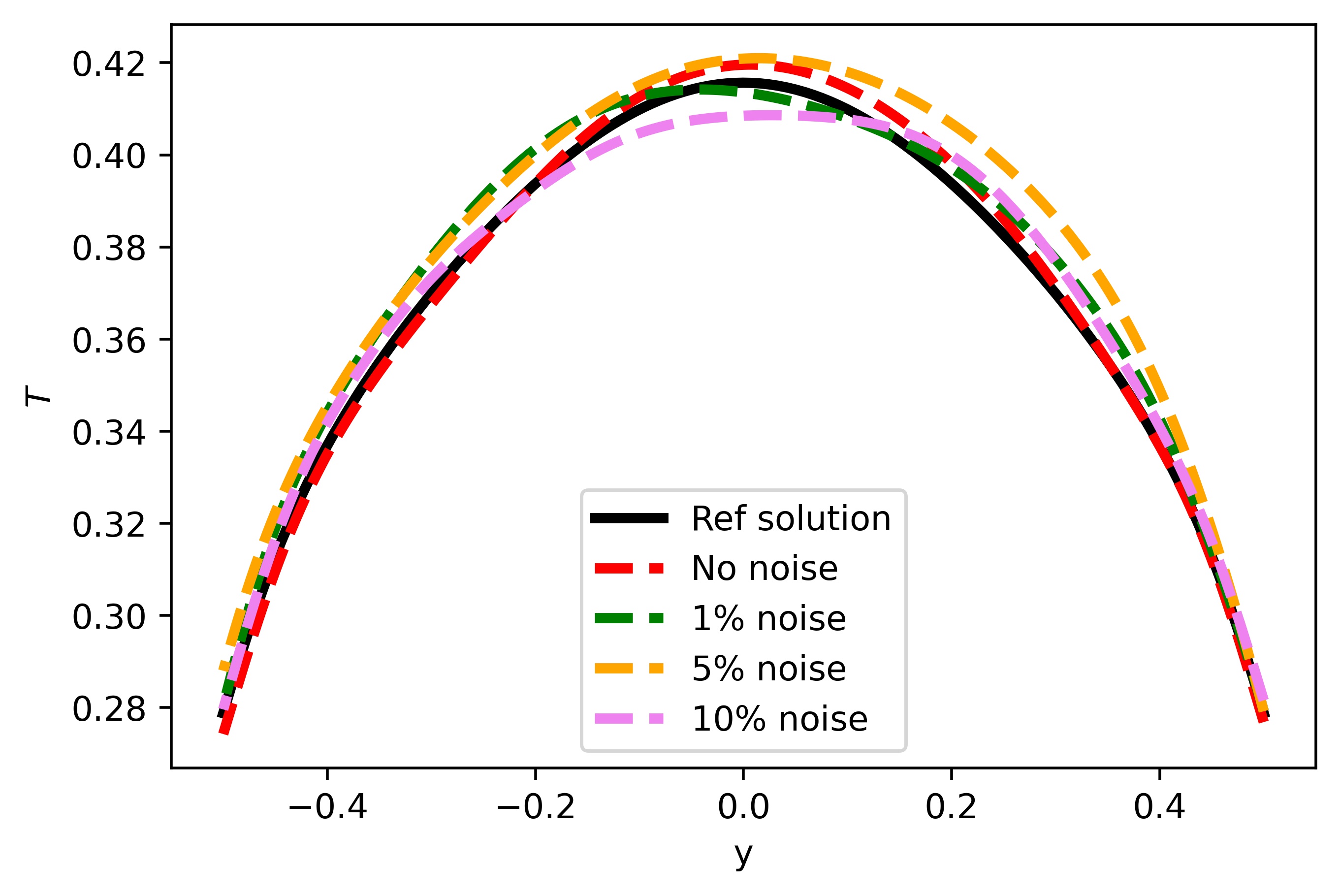} }}%%
    \subfloat[$u_x$]{{\includegraphics[width=4cm]{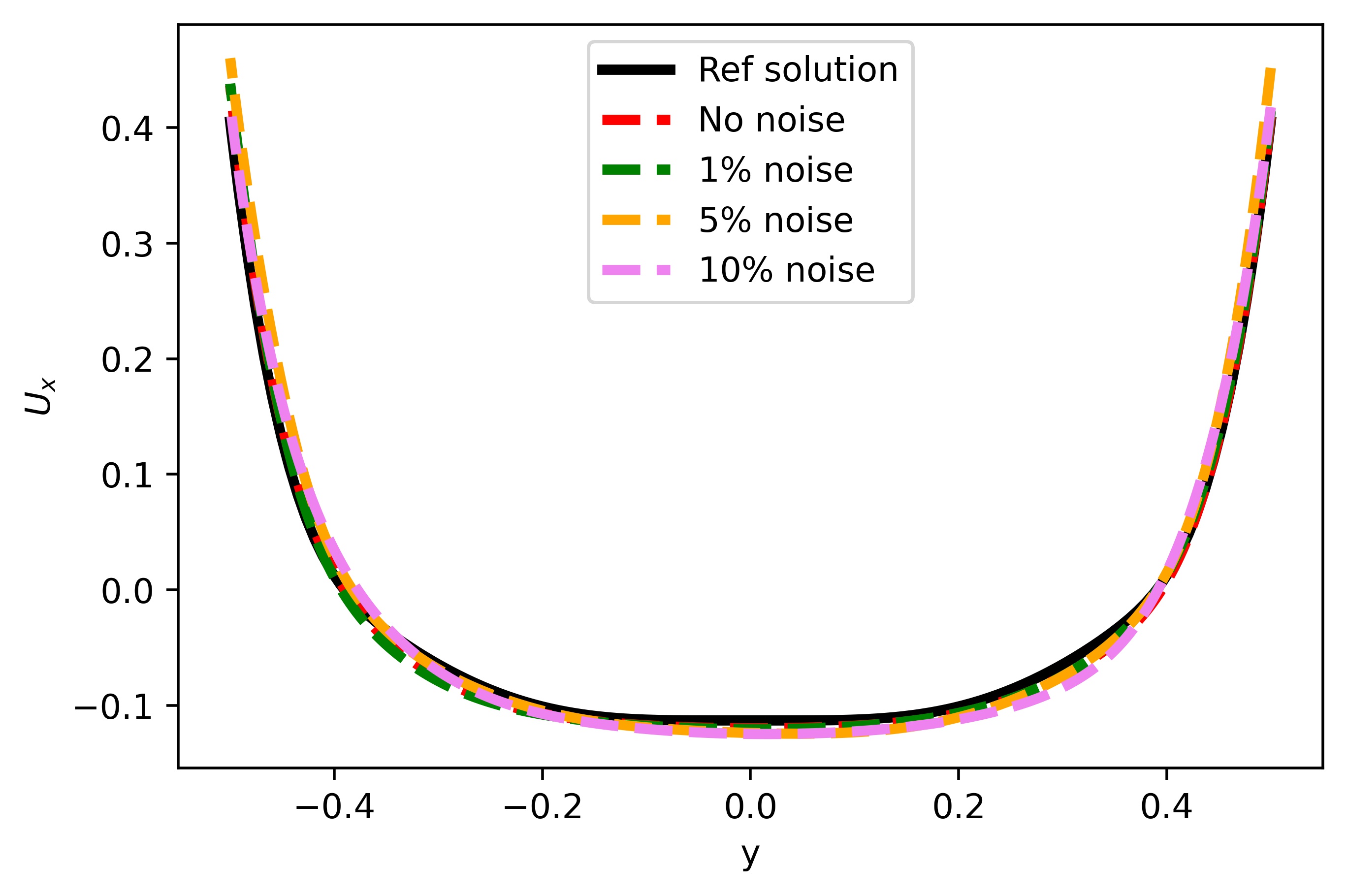} }}%
    \subfloat[$u_y$]{{\includegraphics[width=4.1cm]{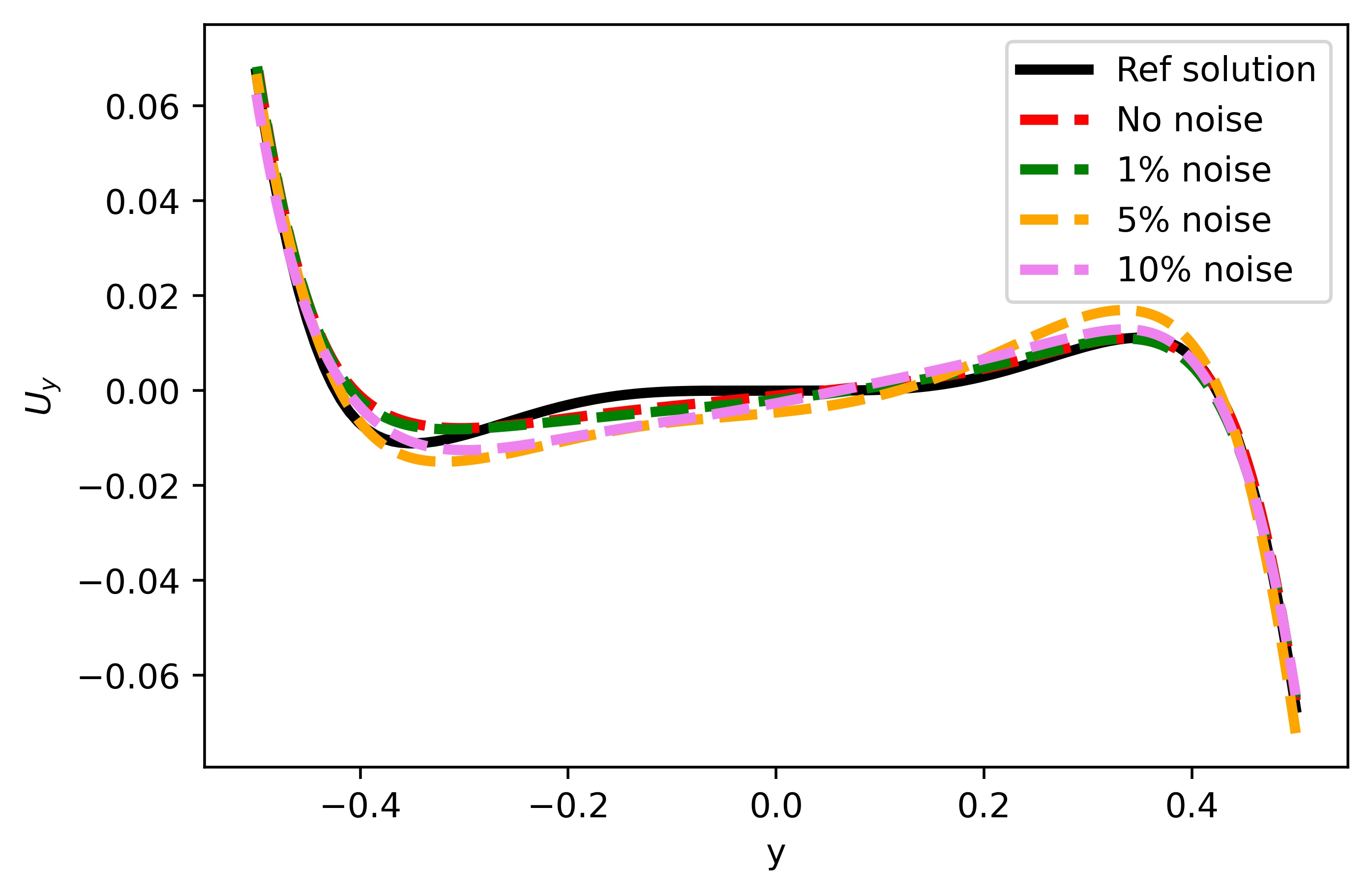} }}%
    \subfloat[$p$]{{\includegraphics[width=4.1cm]{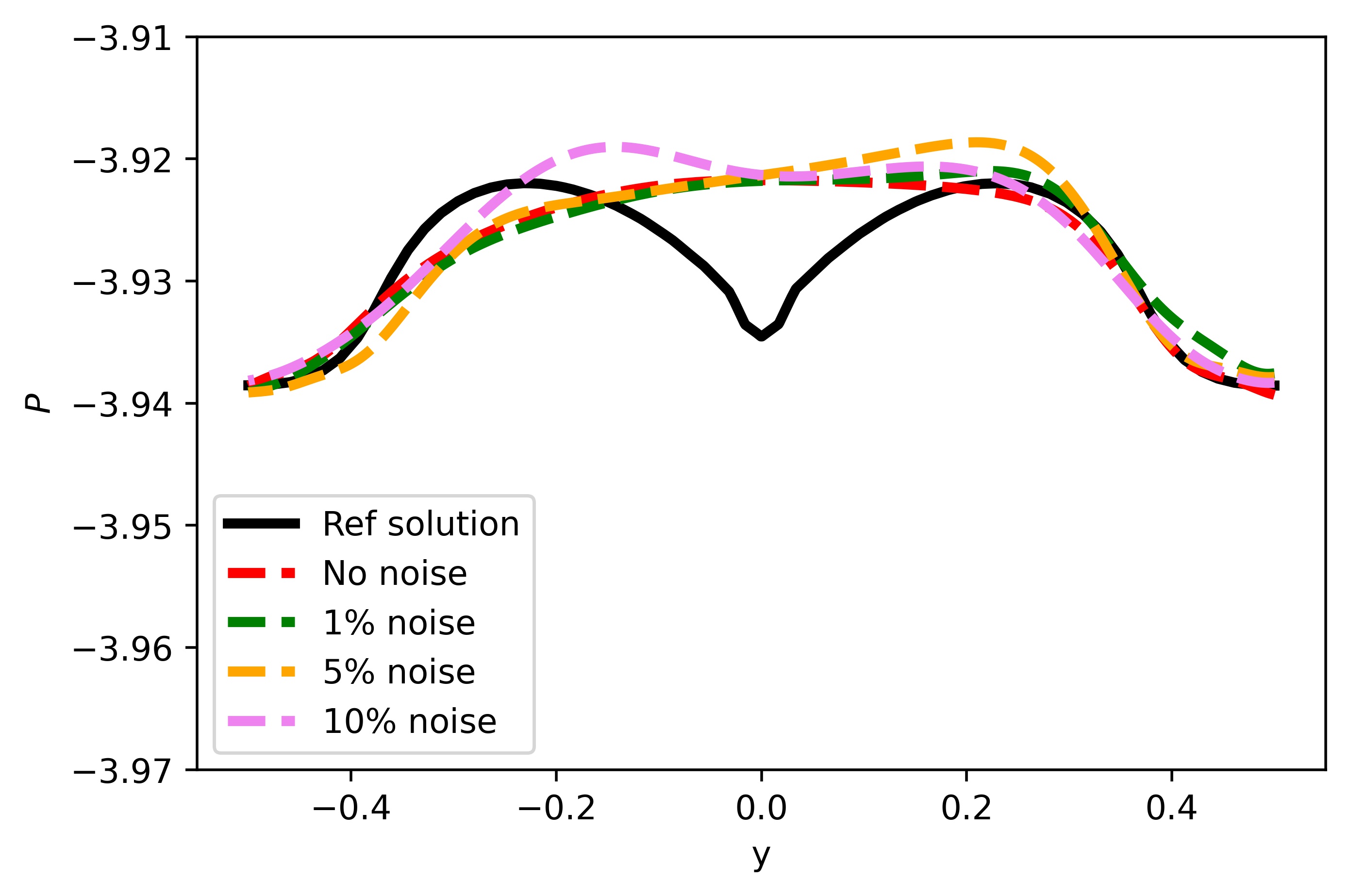} }}%
    \caption{\textit{Reference solution and PINNs prediction for the physical fields (a-d) at the line $y=0$ and (e-h) at the line $x=-6.5$.}}%
    \label{infer_noise}%
\end{figure}

In realistic circumstances, the measurements captured from the sensors are always corrupted with some noise. We investigate here the impact of noisy measurements on the quality of PINNs prediction. To this end, we only consider the case where there are two lines of sensors that are put at the input and output of the calender (case 2). We generate the noise as uncorrelated Gaussian noise for the supervised data. We vary the level of noise to verify the effectiveness of PINNs when dealing with noisy measurements. PINNs configuration using KNNs-tanh is the same as the cases before except that now we train with more epochs. More precisely, we use Adam optimizer with 50,000 epochs with the learning rate $lr=10^{-3},$ 200,000 epochs with the learning rate $lr=10^{-4}$ and 300,000 epochs with $lr=10^{-5}$. Table \ref{infer_err_noise} illustrates in more detail the quality of PINNs prediction in terms of  relative $\mathcal{L}^2$ errors with the reference solution. These errors are evaluated on the finite element mesh. It is clear that the accuracy of PINNs prediction remains robust even when the supervised data are corrupted with $10\%$ noise. Figure \ref{infer_noise} shows the visualization of the reference solution and PINNs prediction at the horizontal line $y=0$ and at the vertical line $x=-6.5$. We can see that at the line $y=0$, the visual prediction for $u_y$ does not fit perfectly the reference solution. However, when we visualize the results at the line $x=-6.5$, which locates very close to the input of the calender, the predictions for $u_y$ fit accurately the reference solution with comparable errors to other fields. On this line, we see that the visual predictions for $T$ and $p$ are not accurate. Even though these errors are very small compared to the scale of $T$ and $p$, this behavior is somehow expected since we do not have many collocation points at the input of the calender. As a result, the predictions for all fields in this area are not as accurate as the ones at the output.

\begin{table}[H]
\centering
\begin{tabular}{|c|c|c|c|c|} 
\hline
& $\epsilon_T$  & $\epsilon_{u_x}$  & $\epsilon_{u_y}$ & $\epsilon_p$ \\
\hline
0\% noise & 3.24 & 4.63 & 4.87 & 0.83 \\
1\% noise & 4.16 & 5.73 & 6.16 & 1.07 \\
5\% noise & 4.55 & 6.94 & 6.57 & 1.64 \\
10\% noise & 5.60 & 7.21 & 7.16 & 1.69 \\
\hline
\end{tabular}
\caption{\textit{Relative $\mathcal{L}^2$ errors between the reference solution and the predictions by PINNs with different level of noise.}}
\label{infer_err_noise}
\end{table}

\subsection{Identifying unknown physical parameters}

In the industrial context, it is often the case that some physical parameters are not defined but we observe some measurements of the solution. These problems are referred to as inverse problems, which are difficult to be solved in practice using traditional methods. In this section, we aim to demonstrate the capability of PINNs of identifying an unknown physical parameter (namely the thermal conductivity of the rubber $\lambda$) from some observed measurements. To this end, we suppose that we do not know the value of the thermal conductivity $\lambda$ in the temperature equation (Eq. (\ref{enerconv_ori})) but we dispose of some measurements for the temperature and the velocity (which may be noisy) from the sensors. We note that the conductivity is essential to define the value of the Péclet number and Brinkman number in the dimensionless PDE. Hence, we use PINNs to define these two unknown parameters in the dimensionless PDE and then deduce the value of the conductivity. We note that, in this case, the only considered PDE is the equation of conservation of energy (Eq. (\ref{enerconv})) as the conductivity only appears in this equation. We study the impact of the sensors' location and also the impact of noisy measurements on PINNs identification capability. 

In identification problems, the unknown parameters (Péclet number and Brinkman number in our case) are considered as network parameters that are learned during the training process. As for the case before, we use PINNs with the spatial coordinates $x,y$ as inputs and $T, p, u_x, u_y$ as outputs and minimize the following cost function:
\begin{align*}
    L &= L_{data} + L_{pde}\\
    &=\dfrac{\omega_T}{N_T}\sum_{i=1}^{N_T}(\hat{T}^i - T^{i*})^2 +
    \dfrac{1}{N_U}\sum_{i=1}^{N_U}\Big(\omega_{u_x}(\hat{u}_x^i - u_x^{i*})^2+\omega_{u_y}(\hat{u}_y^i - u_y^{i*})^2\Big) +
    \dfrac{1}{N_f}\sum_{i=1}^{N_f}\omega_4e_4^2
\end{align*}
where $e_4$ is the dimensionless PDE residual corresponding to Eq. (\ref{enerconv}). The weight coefficients are fixed as follows: $\omega_T=\dfrac{N_T}{\sum_{i=1}^{N_T}( T^{i*})^2}$, $\omega_{u_x}=\dfrac{N_U}{\sum_{i=1}^{N_U}( u_x^{i*})^2}$ and $\omega_{u_y}=\dfrac{N_U}{\sum_{i=1}^{N_U}( u_y^{i*})^2}$ and $\omega_4=1$. We note that here, no information on the boundary is given.

\begin{figure}[H]
    \centering
    \subfloat[Case 1]{{\includegraphics[width=5cm]{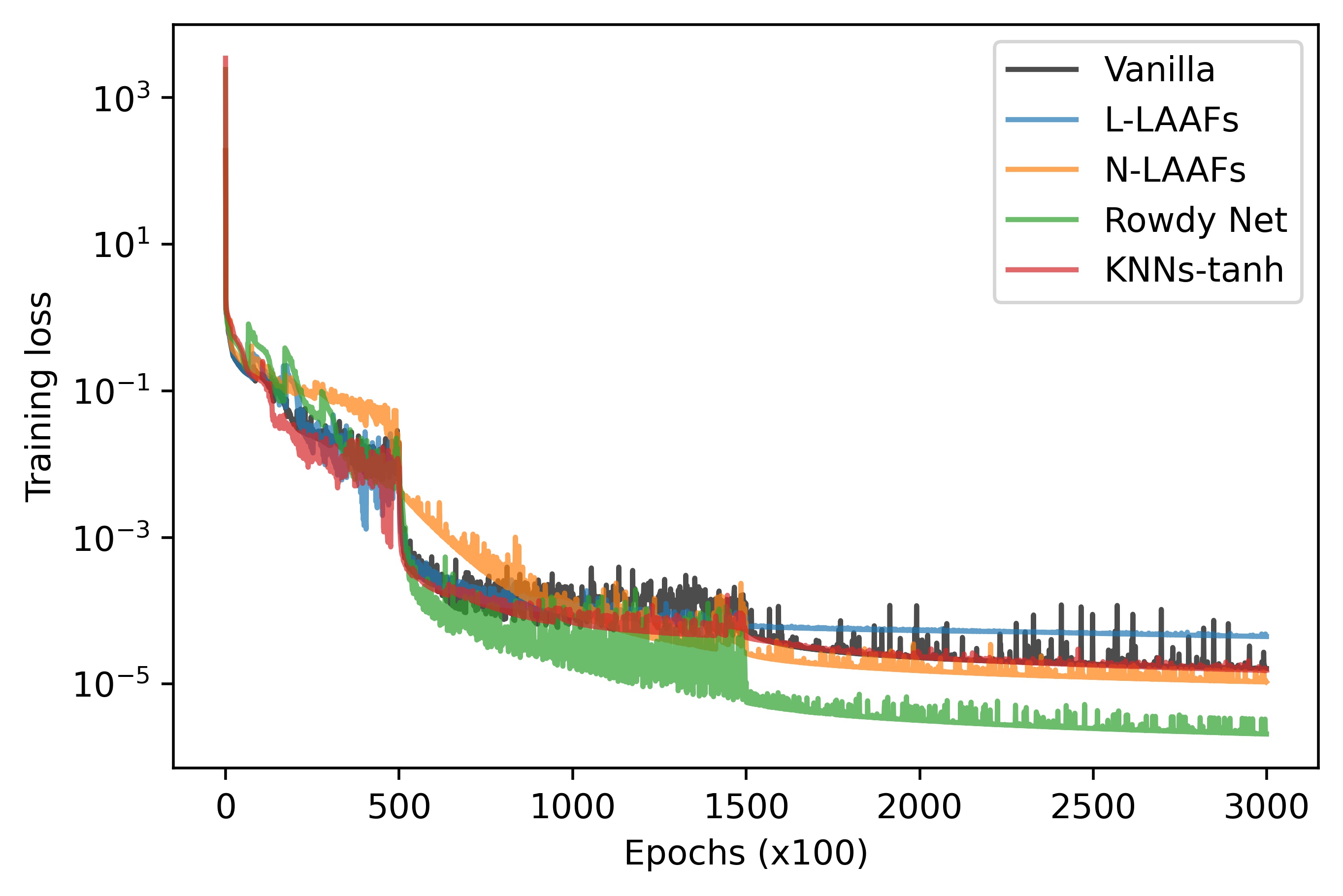} }}%%
    \subfloat[Case 2]{{\includegraphics[width=5cm]{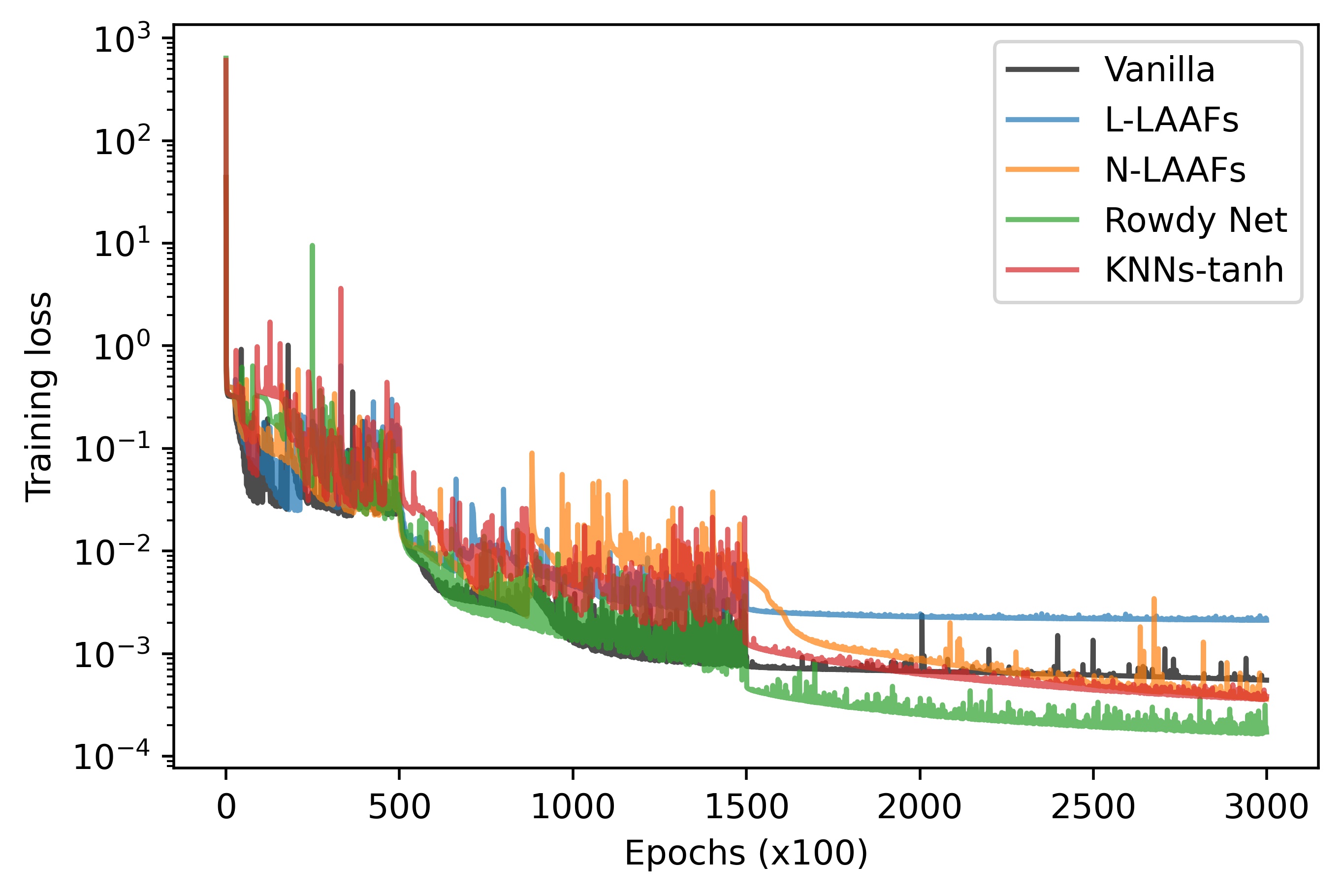} }}%
    \subfloat[Case 3]{{\includegraphics[width=5cm]{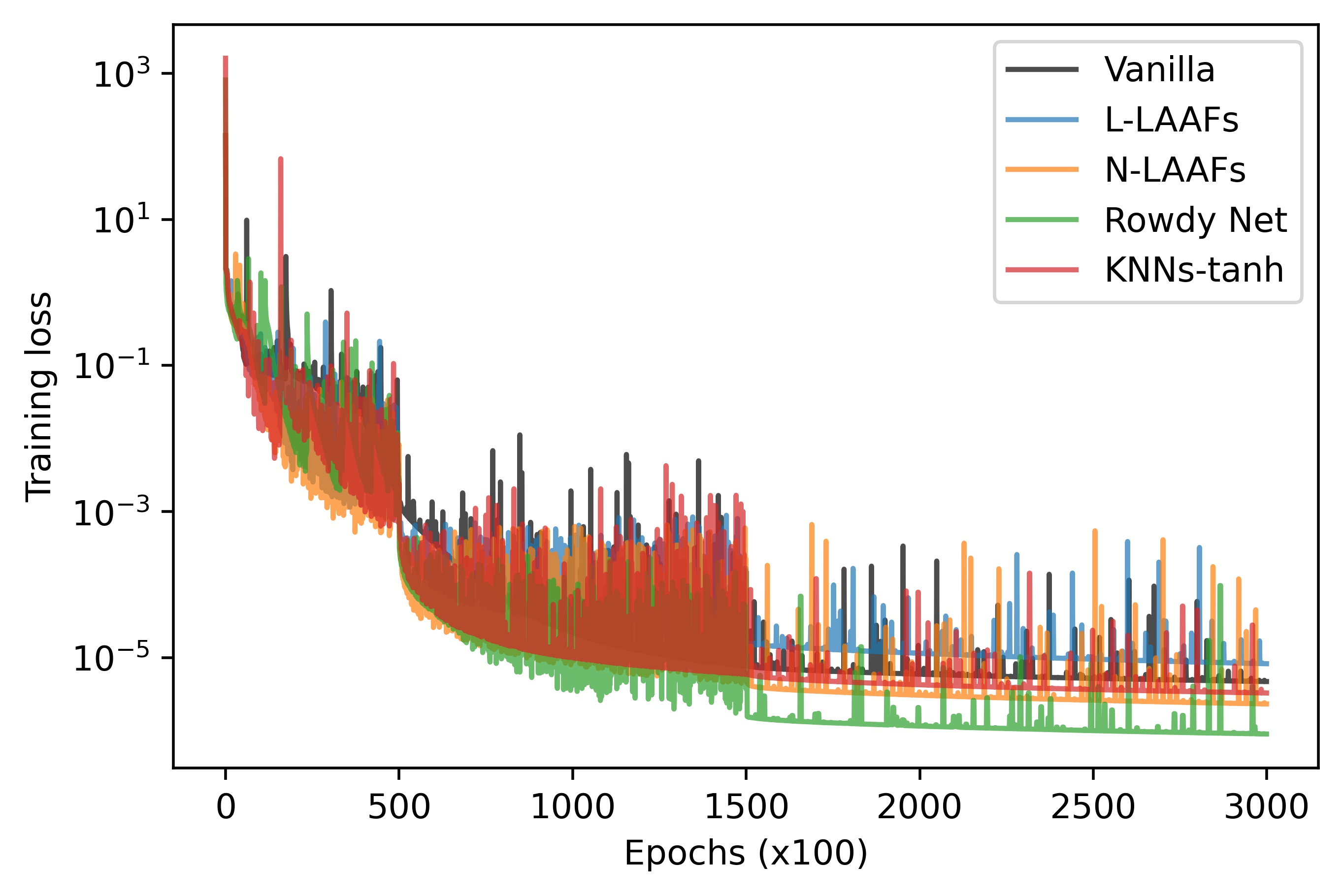} }}%
    \caption{\textit{Cost function during training process with different approaches in inverse problem.}}%
    \label{loss_inv_compare}%
\end{figure}

We first employ and compare vanilla PINNs with PINNs using locally adaptive activation function and deep Kronecker networks. For the training of PINNs, the number of measurements captured from the sensors for the temperature and the velocity is $N_T=N_U=500$. The number of collocation points is $N_f=10,000$ and these points are randomly taken from the finite element mesh. To minimize the cost function, we use Adam optimizer with 50,000 epochs with the learning rate $lr=10^{-3},$ 100,000 epochs with the learning rate $lr=10^{-4}$ and 150,000 epochs with $lr=10^{-5}$. For L-LAAFs and N-LAAFs, we fix the scaling factor $n=1$ and initialize the adaptive parameters $a^k=1$ or $a^k_i=1$. For Rowdy Net and KNNs-tanh, we fix $K=2$ and $\alpha_l^k=\omega_l^k=1$ and the scaling factor $n=1$ (for Rowdy Net). Figure \ref{loss_inv_compare} shows the cost function during the training process for different cases of sensor location. In all cases, we see that PINNs with Rowdy Net offer significantly faster convergence and better training loss than other models. In addition, we note that during the training process, except L-LAAFs, other models are very sensitive with high learning rates, especially the models with Rowdy Net and KNNs-tanh, and then become more stable with smaller learning rate. Table \ref{inv_compare_err} illustrates in more detail the performance of each approach in terms of relative errors in case 1 of sensor location. Again, we see clearly that the models with KNNs-tanh and N-LAAFs perform slightly better than the vanilla PINNs while the model with Rowdy Net outperforms the others. We can increase the performance of Rowdy Net and KNNs-tanh by increasing the number of terms $K$. However, we observe that with $K=2$, these models already offer satisfying results. 

\begin{table}[H]
\centering
\begin{tabular}{ |c|c|c|c| } 
\hline
 & $\epsilon_{1/Pe}$ & $\epsilon_{Br/Pe}$ & $\epsilon_{\lambda}$\\
\hline
Vanilla & 22.5 $\pm$ 0.51 & 11.2 $\pm$ 0.41 & 0.85 $\pm$ 0.02 \\
L-LAAFs & 41.6 $\pm$ 0.30 & 23.3 $\pm$ 0.33 & 1.49 $\pm$ 0.01 \\
N-LAAFs & 16.0 $\pm$ 0.42 & 7.45 $\pm$ 0.04 & 0.61 $\pm$ 0.03  \\
\textbf{Rowdy Net} & \textbf{2.18 $\pm$ 0.04} & \textbf{0.09 $\pm$ 0.02} & \textbf{0.04 $\pm$ 0.00}  \\
KNNs-tanh & 19.3 $\pm$ 0.32 & 8.16 $\pm$ 0.05 & 0.74 $\pm$ 0.03\\
\hline
\end{tabular}
\caption{\textit{Relative error (in \textpertenthousand) between reference values and PINNs prediction with different approaches in case 1.}}
\label{inv_compare_err}
\end{table}

Next, we investigate the impact of the location of sensors on the performance of PINNs. To this end, we use the same configuration as before and PINNs with Rowdy Net, which is proven to be more efficient than other models as shown in Figure \ref{loss_inv_compare}, to identify the unknown parameters in all cases.

\begin{figure}[H]
    \centering
    \subfloat[$1/Pe$]{{\includegraphics[width=5.5cm]{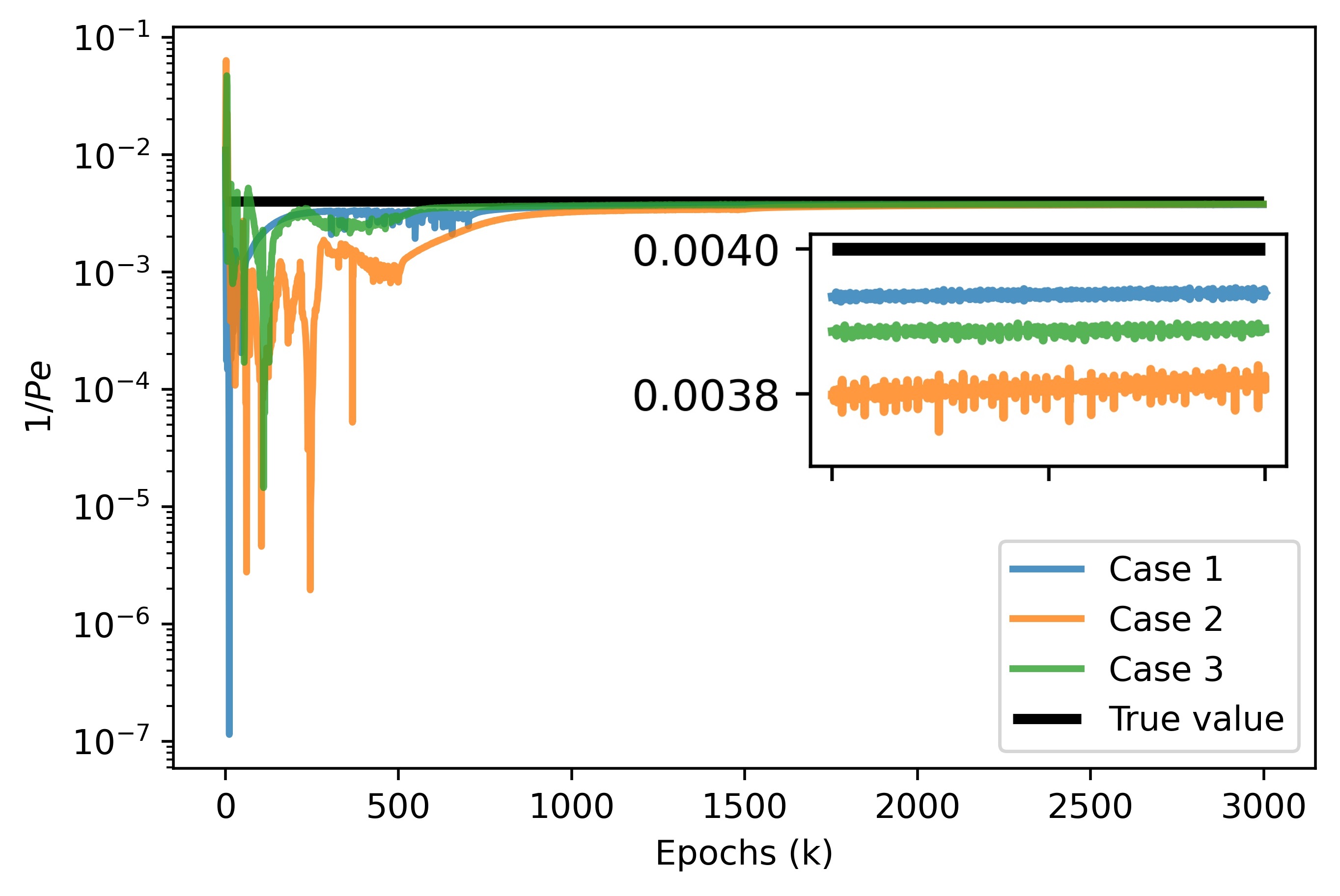} }}%%
    \subfloat[$Br/Pe$]{{\includegraphics[width=5.5cm]{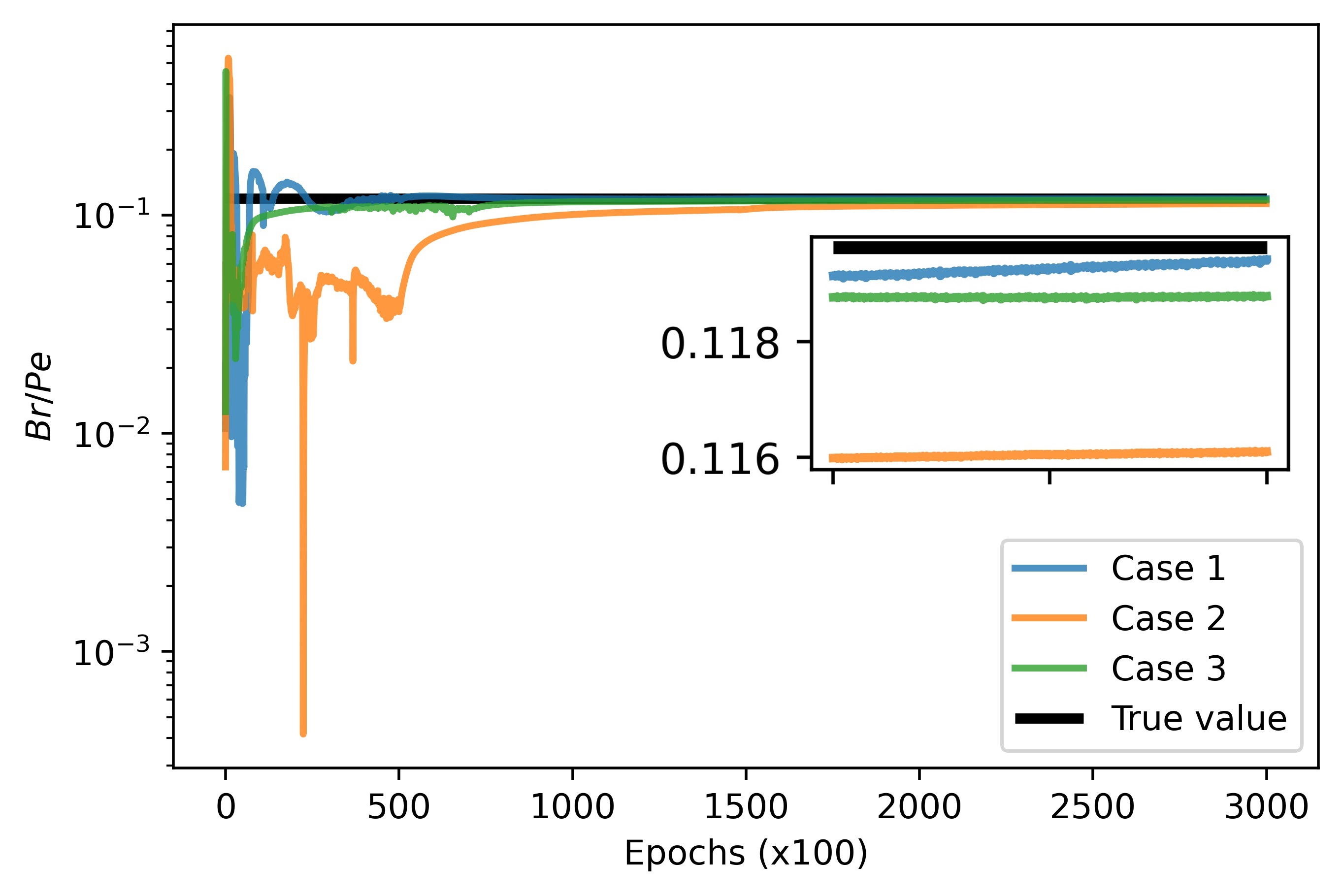} }}%
    \subfloat[$\lambda$]{{\includegraphics[width=5.5cm]{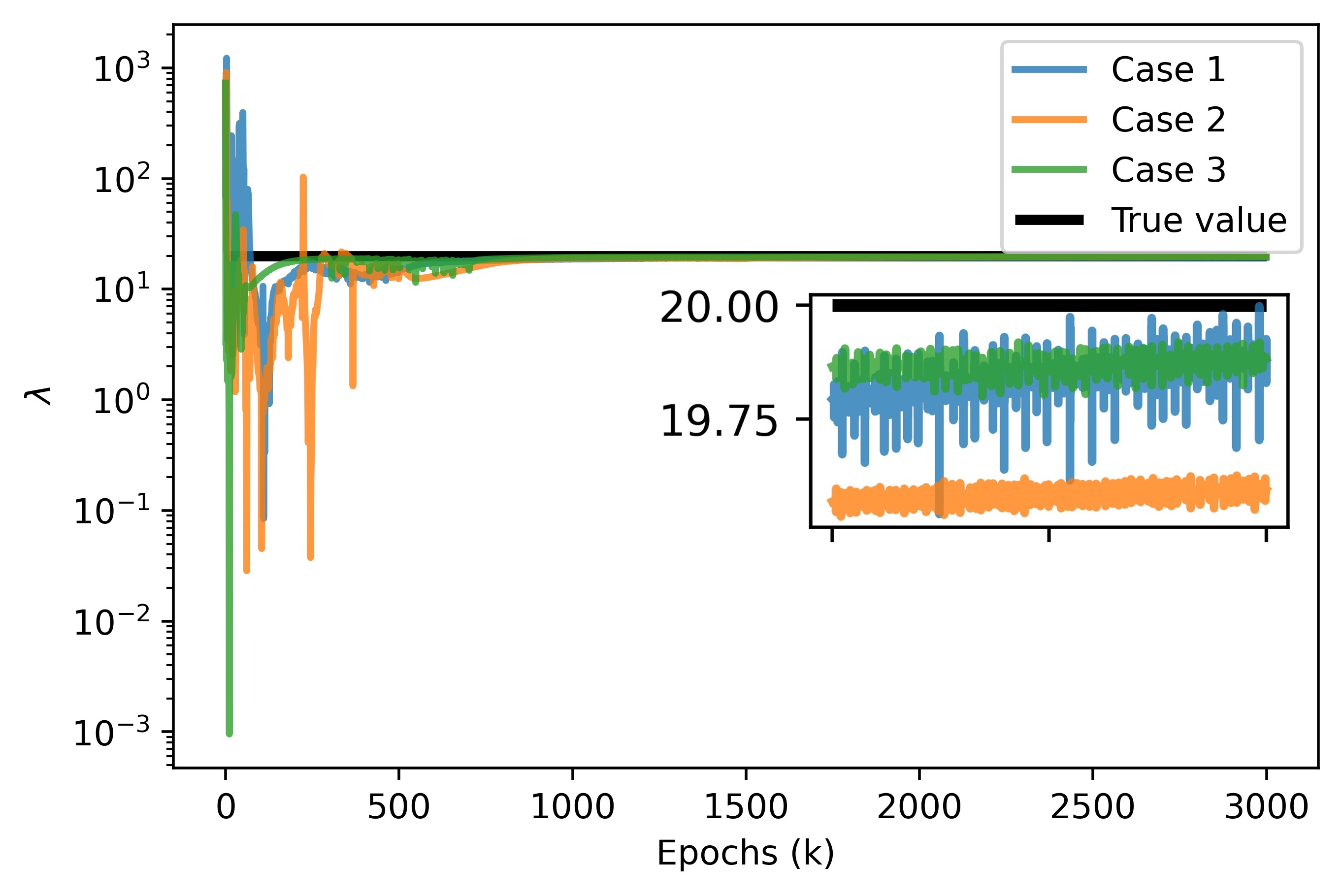} }}%
    \caption{\textit{Prediction for unknown parameters during the training process while considering different cases of the location of sensors.} The small plots show the results at the last 100,000 epochs.}%
    \label{iden_inv}%
\end{figure}
\begin{table}[H]
\centering
\begin{tabular}{ |c|c|c|c| } 
\hline
 & $1/Pe$ & $Br/Pe$ & $\lambda$ \\
\hline
High-fidelity model & 4.00e-03 & 1.20e-01 & 2.00e+01 \\
\hline
Case 1 & 3.94e-03 & 1.20e-01 & 1.99e+01 \\
Case 2 & 3.81e-03 & 1.16e-01 & 1.96e+01 \\
Case 3 & 3.89e-03 & 1.19e-01 & 1.98e+01 \\
\hline
\end{tabular}
\caption{\textit{Prediction for unknown parameters at the end of PINNs training process.}}
\label{iden_error}
\end{table}
Figure \ref{iden_inv} shows the prediction values for $1/Pe, Br/Pe$ and $\lambda$ during the process. We note that the value of $\lambda$ is inferred from the values of $1/Pe$ and $Br/Pe$. In all cases, the predictions of PINNs for PDEs parameters converge to the true values. Table \ref{iden_error} presents the final prediction for these parameters at the end of the training process. We see that the case 1 (Figure \ref{calender_geo_sensor0}) and case 3 of the location of sensors (Figure \ref{calender_geo_sensor6}) provide a slightly better performance than the case 2 (Figure \ref{calender_geo_sensor1}). However, it is remarkable that in case 2, with only two lines of sensors at the input and output of the calender, PINNs are still capable of approximating the unknown parameters with high accuracy.

Next, we investigate the impact of noisy measurements on the performance of PINNs. We generate the noise as uncorrelated Gaussian noise. We vary the number of supervised points and the level of noise to verify the effectiveness of PINNs when dealing with noisy measurements. We use the same configuration as the case before for PINNs with Rowdy Net, however, we increase the number of epochs for the training process. More precisely, we use Adam optimizer with 50,000 epochs with the learning rate $lr=10^{-3},$ 100,000 epochs with the learning rate $lr=10^{-4}$ and 200,000 epochs with $lr=10^{-5}$.

\begin{figure}[H]
    \centering
    \subfloat[Case 1]{{\includegraphics[width=5cm]{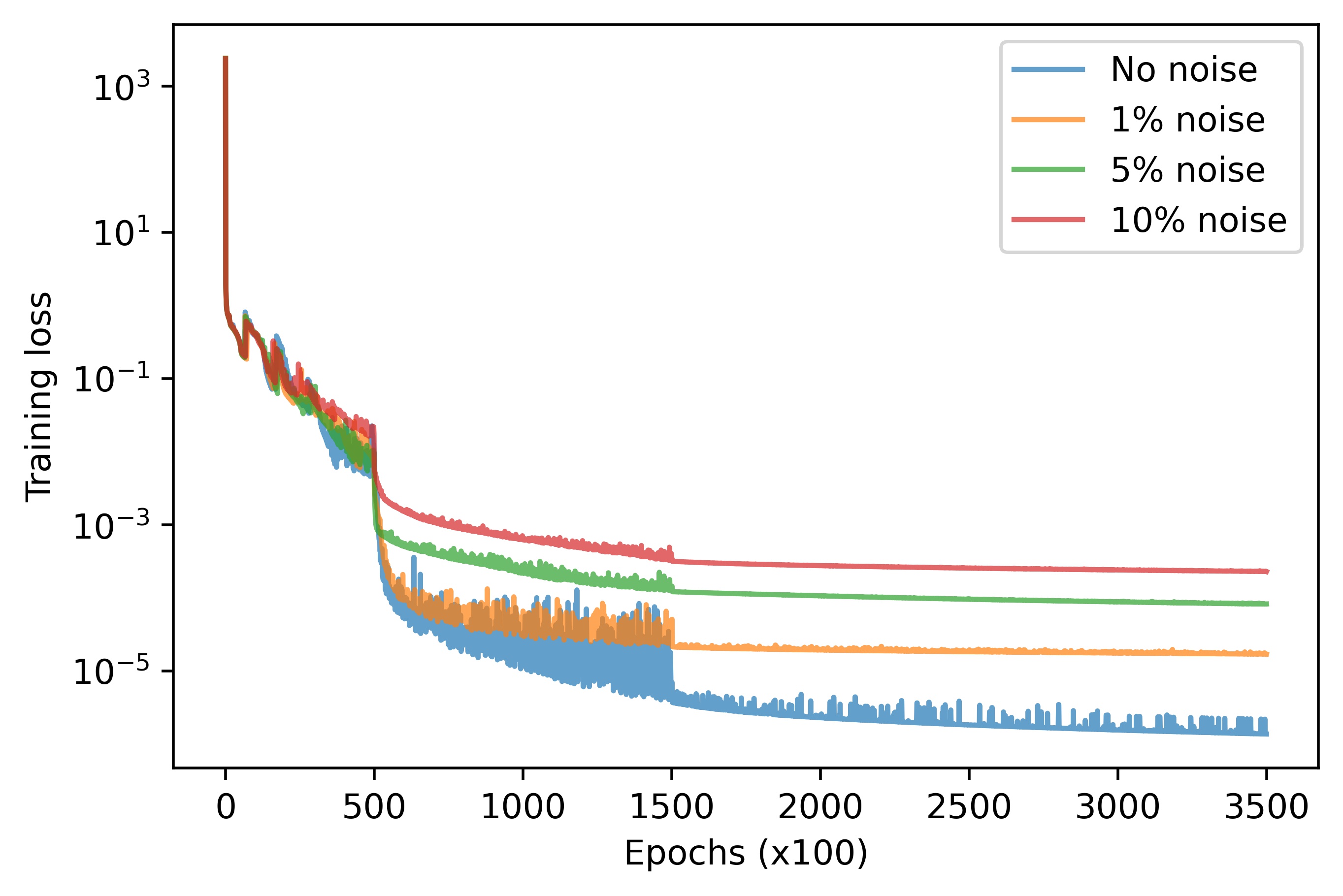} }}%%
    \subfloat[Case 2]{{\includegraphics[width=5cm]{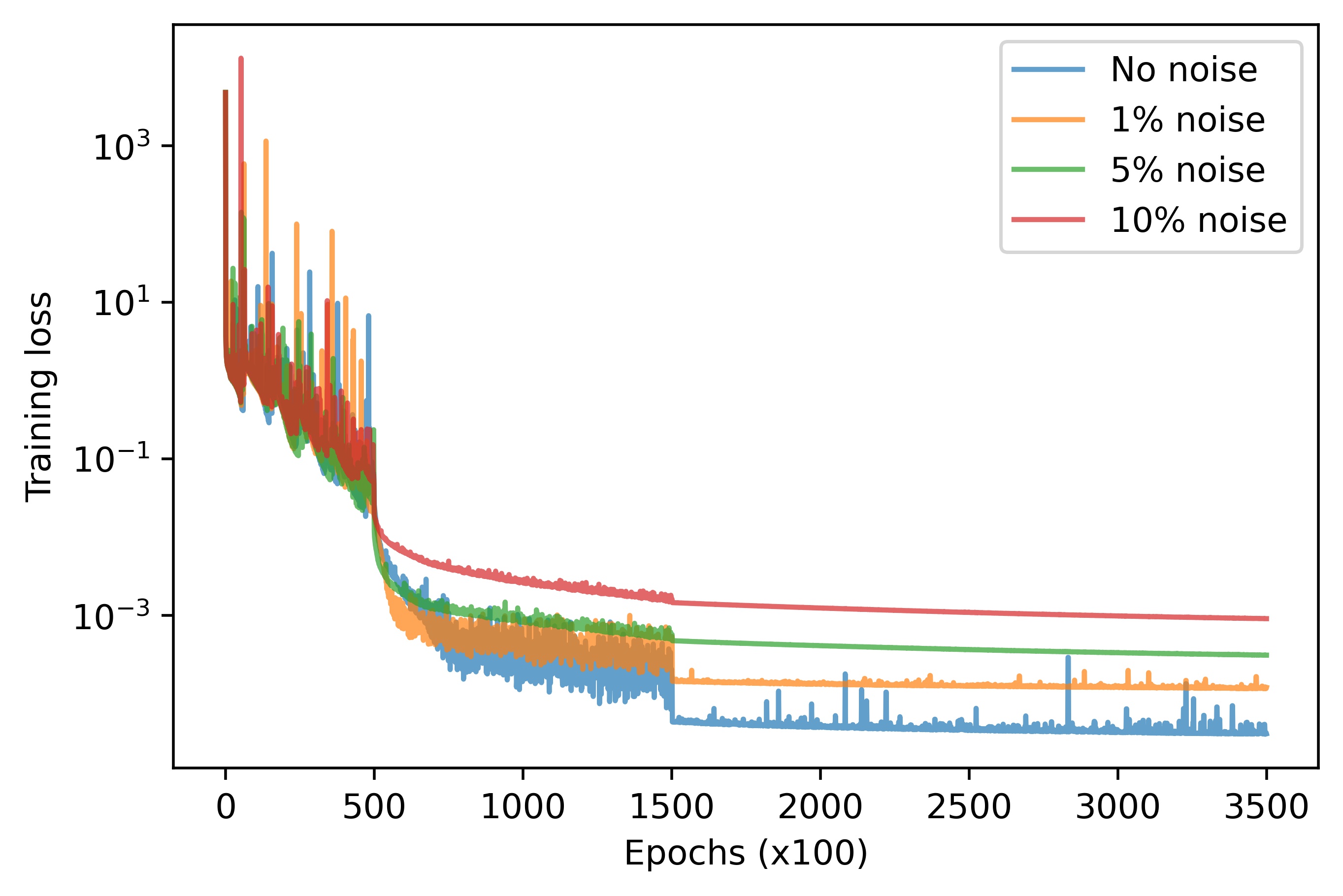} }}%
    \subfloat[Case 3]{{\includegraphics[width=5cm]{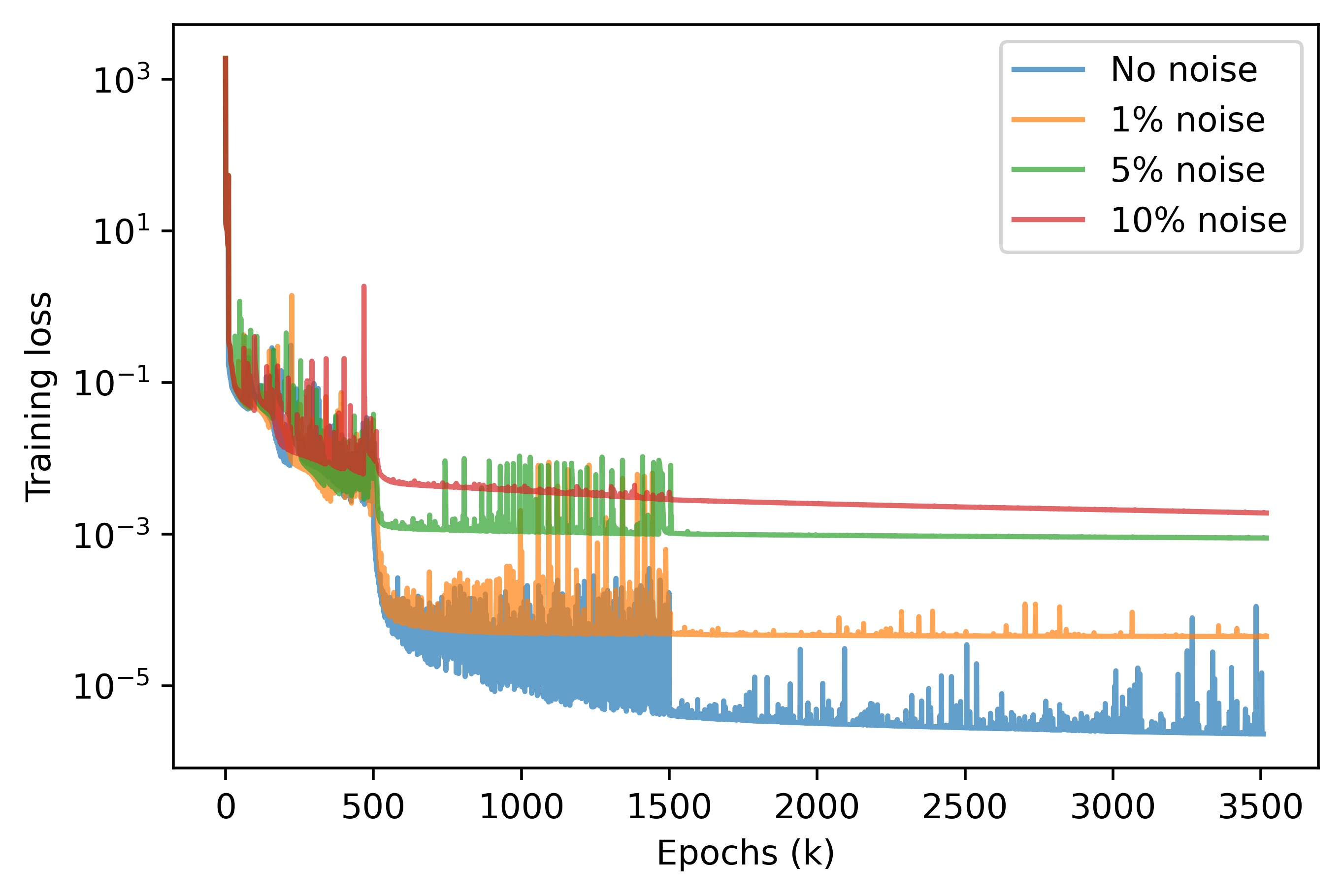} }}%
    \caption{\textit{Cost function during training process with noisy measurements.}}%
    \label{loss_inv}%
\end{figure}

\begin{table}[H]
\centering
\begin{tabular}{ |c|c|c|c| } 
\hline
 & $1/Pe$ & $Br/Pe$ & $\lambda$ \\
\hline
High-fidelity model & 4.00e-03 & 1.20e-01 & 2.00e+01 \\
\hline
$0\%$ noise & 3.94e-03 & 1.19e-01 & 1.98e+01 \\
$1\%$ noise & 3.89e-03 & 1.18e-01 & 1.97e+01 \\
$5\%$ noise & 3.81e-03 & 1.16e-01 & 1.96e+01 \\
$10\%$ noise & 3.77e-03 & 1.16e-01 & 1.94e+01 \\
\hline
\end{tabular}
\caption{\textit{Prediction for unknown parameters at the end of PINNs training process while considering different levels of noisy measurements in supervised data in case 3.}}
\label{iden_error_noise}
\end{table}

Figure \ref{loss_inv} illustrates the cost function during the training process for different level of noisy measurements. We see that the more the level of noise increases, the more the model has difficulties in minimizing the cost function. This is somehow expected since with the effect of noise, the PDEs constraints are less respected by the supervised measurements. Table \ref{iden_error_noise} summarizes the performance of PINNs for case 3 in terms of prediction values compared to reference values. We see that, even though the accuracy decreases when the level of noise increases, PINNs are still able to approximate the unknown parameters with high accuracy even when the training data are corrupted with $10\%$ noise. 

\section{Conclusion}

In this study, we demonstrated the interest of using Physics-Informed Neural Networks (PINNs) for the modeling of the rubber calendering process. As far as our knowledge, this is the first time PINNs are applied to solve a non-Newtonian fluid thermo-mechanical problem. PINNs have shown a great capability to infer all the physical fields without using any supervised data for the velocity and the pressure fields but taking some knowledge on the boundary for the velocity and the temperature. We investigated the methods using locally adaptive activation functions (L-LAAFs and N-LAAFs) and deep Kronecker networks (Rowdy Net and KNNs-tanh). The results showed that these methods help to improve significantly the accuracy of PINNs predictions. We demonstrated an important impact of the placement of the sensors (or the position of supervised data) on the prediction quality of PINNs. We investigated the effect of the distribution of collocation points on the prediction quality. We concluded that training PINNs with the collocation points that were taken randomly on a finite element mesh (which provides an \textit{a priori} knowledge on high gradient location) provides much better models than training with the collocation points that were taken randomly inside the domain. We also tested the robustness of PINNs when dealing with noisy measurements. The results demonstrated that when training with sufficient epochs, PINNs are capable of predicting the reference solution with the same accuracy as when using high-fidelity data. Concerning the identification problem, we showed that with any case of the location of sensors, PINNs are able to approximate the unknown thermal conductivity. We also demonstrated that even when the supervised data are corrupted with noise, the quality of PINNs prediction still remains accurate as expected.

Through our work, it can be seen that there are many challenges and opportunities for further research. We see that, in the cost function, we used pre-specified weight coefficients to achieve a good performance in PINNs. However, these pre-specified coefficients were obtained through trial and error method, and they may give different performances if there are any small changes in the geometry or the supervised measurements. Thus there is a need for an adaptive method that is able to provide the best values for these coefficients after the training process \citep{mcclenny2020self, wang2020understanding, wang2020and}. Besides that, we observed that in some cases, PINNs are able to give a satisfying performance qualitatively, but however still provide a very high relative error compared to the reference solution. This suggests to examine the accuracy of PINNs on more appropriate quantities which present some expected physical properties. We can think of methods of model reduction such as Proper Orthogonal Decomposition (POD) \citep{berkooz1993proper} or Dynamic Mode Decomposition (DMD) \citep{schmid2010dynamic} to extract the most important modes (in terms of energy or dynamic), and then calculate the error of PINNs on these modes. Another perspective would be to combine PINNs and reduced-order methods to improve learning efficiency. We aim to purchase these lines of research in our future work.

\section*{Acknowledgements}
T.N.K. Nguyen is funded by Michelin and CEA through the Industrial Data Analytics and Machine Learning chair of ENS Paris-Saclay. Part of this work has been funded by Region Ile-de-France. Part of the computations has been executed on Atos Edge computer. In addition, the authors would like to thank Christophe Millet for fruitful discussions and the anonymous reviewers whose comments and suggestions helped improve and clarify significantly this paper.

\section*{Conflict of interest}
The authors declare no conflict of interest.

\bibliographystyle{plainnat} %unsrtnat
\bibliography{main} 

\appendix
\section{Validation of PINNs with analytical solution}\label{sec_analytic_pipe}
\begin{figure}[H]
\centering
\includegraphics[width=4cm]{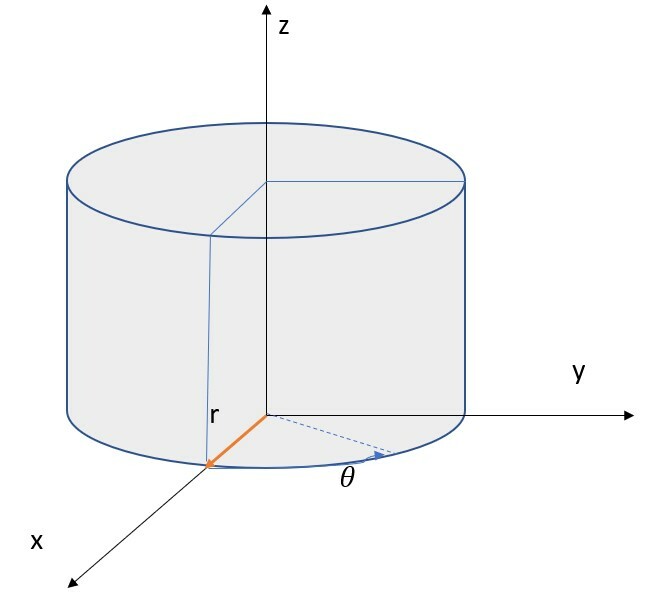}
\caption{Sketch of the pipe configuration}
\label{cylinder}
\end{figure}

In this section, we provide the comparison of PINNs and the analytical solution in a pipe (Figure \ref{cylinder}), using the same Stokes equation but without thermal coupling. The considered PDE is written as follows:
\begin{align*}
    &\nabla.(2\eta(\bm{\vec{u}})\bar{\bar{\epsilon}}(\bm{\vec{u}})) - \vec{\nabla}p =\vec{0} \\
    &\nabla.\bm{\vec{u}} =0  \\
\end{align*}
where the genrealized strain rate $\gamma (\bm{\vec{u}})=\sqrt{2\sum_{i,j}\bar{\bar{\epsilon}}_{i,j}^2}$ and the dynamic viscosity: $\eta(\bm{\vec{u}})=K|\gamma (\bm{\vec{u}})|^{n-1}$. We set the boundary condtion as follows: the velocity of the fluid at the wall is supposed to be zero (no-slip condition), and the input flow rate $Q$ is fixed.

To calculate the analytical solution, we use the cylindrical coordinates $(r, \theta, z)$. We suppose that the flow is axisymmetric ($\dfrac{\partial}{\partial \theta}=0$), and we search for a solution $\bm{\vec{u}}=(u_r, u_{\theta}, u_z)$ where $u_r=0, u_{\theta}=0$. 
By re-writting the PDE in cylindrical coordinates, the equation of conservation of mass gives us:
\begin{align*}
    \nabla.\bm{\vec{u}} =0 \Leftrightarrow \dfrac{\partial u_z}{\partial z} = 0
\end{align*}
And the equation of conservation of energy gives us:
\begin{align*}
    &\nabla.(2\eta(\bm{\vec{u}})\bar{\bar{\epsilon}}(\bm{\vec{u}})) - \vec{\nabla}p =\vec{0} \\
    \Leftrightarrow & \begin{cases}
      \dfrac{\partial p}{\partial r} = 0\\
      \dfrac{\partial p}{\partial \theta}=0\\
      \dfrac{1}{r}\dfrac{\partial }{\partial r}(r\eta(\bm{\vec{u}})\dfrac{\partial u_z}{\partial r}) - \dfrac{\partial p}{\partial z} = 0
    \end{cases} 
\end{align*}
In the last obtained equation, by taking the integral three times and using the boundary condition $u_z=0$ at $r=R$, we get:
\begin{align*}
    u_z = (-\dfrac{\Delta P}{LK} \dfrac{1}{2})^{1/n}\dfrac{n}{n+1}(R^{\frac{n+1}{n}}-r^{\frac{n+1}{n}}) 
\end{align*}
To calculate $u_z$ explicitly, we use the analytical formulation of the pressure difference along the pipe length:
\begin{align*}
\Delta P =-
2(\dfrac{Q(\frac{1}{n}+3)}{\pi R^3})^nL\dfrac{K}{R}
\end{align*}
where $L$ is the length of the pipe and $Q$ the flow rate.

We now use PINNs to solve this problem by assuming that the boundary condition is unknown but we dispose of the solution at some points in the pipe. The goal is to infer the pressure and the velocity at all points in the domain. Due to the symmetry of the problem, we only consider a quarter of the pipe. 
\begin{figure}[H]
    \centering
    \subfloat[Analytical solution]{{\includegraphics[width=15.5cm]{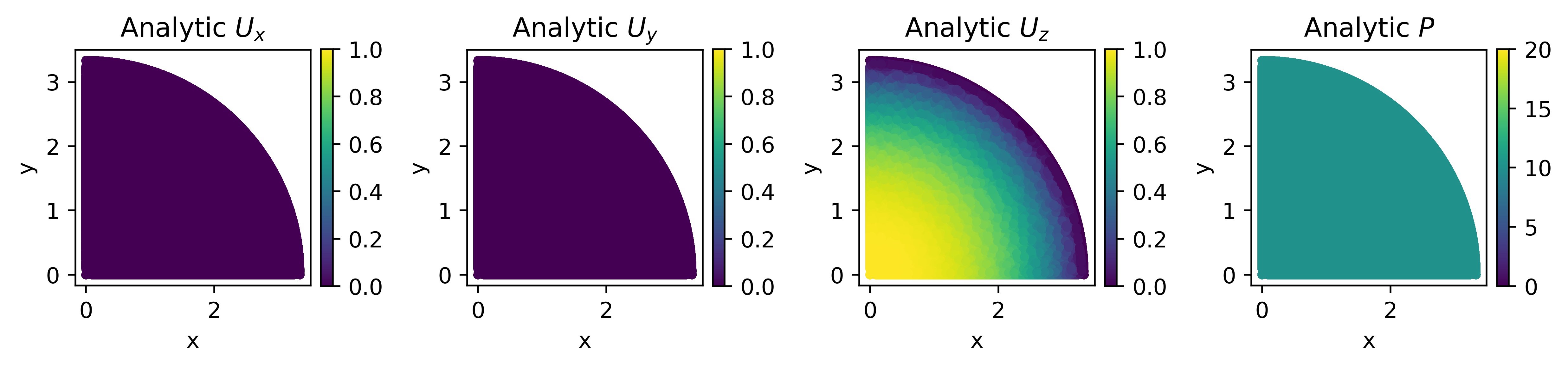} }}%%
    \quad
    \subfloat[PINNs prediction]{{\includegraphics[width=15.5cm]{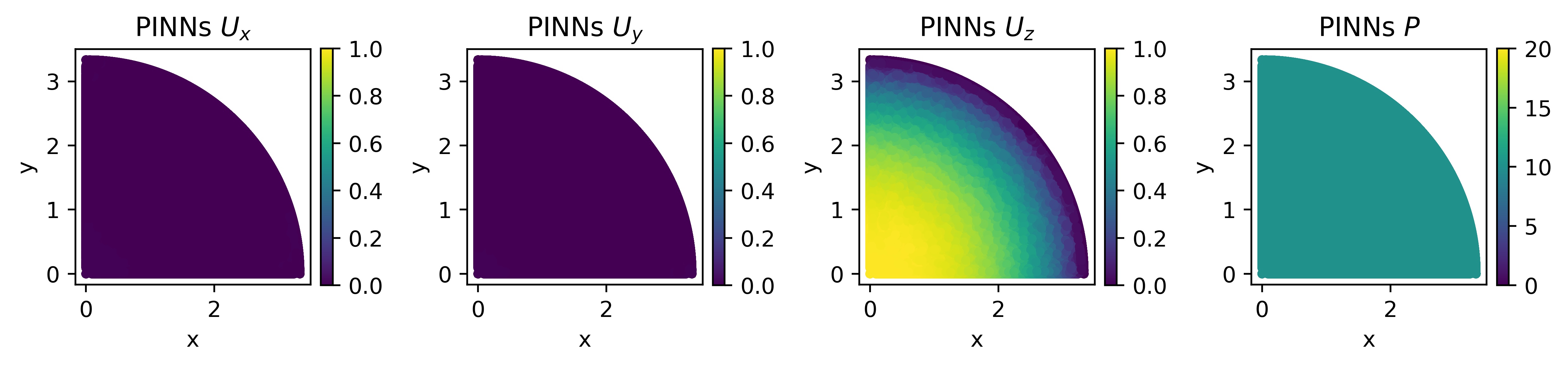} }}%
    \quad
    \subfloat[Absolute errors between analytical solution and PINNs prediction]{{\includegraphics[width=15.5cm]{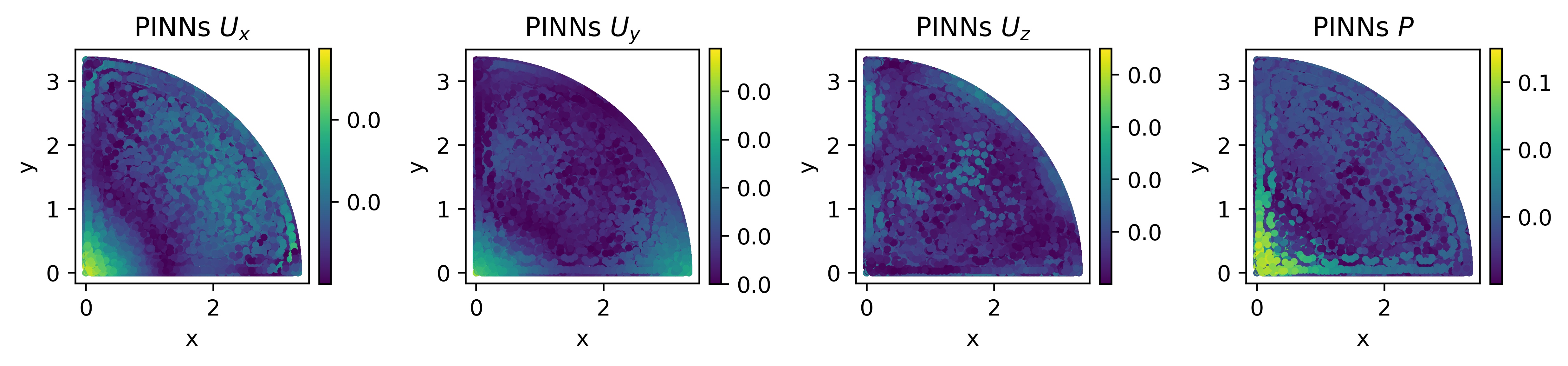} }}%
    \caption{\textit{Visualization of analytical solution and PINNs prediction at the plan $z=cst$:} from left to right: the velocity components $(u_x,u_y,u_z)$ and the pressure $p$.}%
    \label{pipe_visu}%
\end{figure}
We fix the number of supervised points for the velocity as $N_U=5,000$ and the number of collocation points is $N_f=20,000$. We suppose that both the supervised points and collocation points are randomly distributed in the pipe. We use a feedforward network of 5 layers and 100 neurons in each layer. To minimize the cost function, we use Adam optimizer with 20,000 epochs with the learning rate $lr=10^{-3}$. The cost function is defined similarly to the cost function in section \ref{sec_calender_infer}. Figure \ref{pipe_visu} presents the comparison of the analytical solution and PINNs prediction at the plan $z=cst$. The relative $\mathcal{L}^2$ errors at all points in the pipe for $u_z$ and $P$ are 2.59e-02 and 1.87e-02, respectively.

\end{document}